%% file: main.tex
\documentclass[journal]{IEEEtran}
\usepackage{amsmath,amsfonts}
\usepackage{algorithmic}
\usepackage{algorithm}
\usepackage{array}
\usepackage[caption=false,font=normalsize,labelfont=sf,textfont=sf]{subfig}
\usepackage{textcomp}
\usepackage{stfloats}
\usepackage{url}
\usepackage{verbatim}
\usepackage{graphicx}
\usepackage{cite}
\usepackage{preamble}

\newtheorem{example}{Example}
\newtheorem{theorem}{Theorem}
\newtheorem{proposition}{Proposition}
\newtheorem{corollary}{Corollary}

\input{content/acronyms}

\allowdisplaybreaks

\begin{document}
\bstctlcite{BSTcontrol}

\title{Time Series Compression using Quaternion Valued Neural Networks and Quaternion Backpropagation}

\author{
    Johannes Pöppelbaum and Andreas Schwung
    \thanks{Johannes Pöppelbaum and Andreas Schwung are with South Westphalia University of Applied Sciences}
}

\markboth{\textit{Preprint}}{}%


\maketitle

\begin{abstract}
\input{content/abstract}
\end{abstract}


\input{content/01_introduction}
\input{content/02_related_work}

\input{content/03_fundamentals}

\input{content/04_backpropagation}
\input{content/04-2_autoGrad}
\input{content/methodology}

\input{content/experiments_ts_compression}

\input{content/06_conclusion}




\bibliographystyle{IEEEtran}

\bibliography{bibtex/bib/IEEEabrv, backprop, references, bstControl, additional_references}

\newpage
\appendices
\input{content/appendix}

\vfill

\end{document}

%% file: content/acronyms.tex
\usepackage[nolist]{acronym}

\begin{acronym}
    \acro{IoT}{Internet-of-Things}
    \acro{TE}{Tennessee Eastman}
    \acro{MLP}{multilayer perceptron}
    \acro{RNN}{recurrent neural network}
    \acro{CNN}{convolutional neural network}
    \acro{NN}{neural network}
    \acro{QNN}{quaternion valued neural network}
    \acro{PAA}{piecewise aggregate approximation}
    \acro{VAE}{variational autoencoder}
    \acro{GAN}{generative adversarial networks}
\end{acronym}

%% file: content/abstract.tex
We propose a novel quaternionic time-series compression methodology where we divide a long time-series into segments of data, extract the min, max, mean and standard deviation of these chunks as representative features and encapsulate them in a quaternion, yielding a quaternion valued time-series. This time-series is processed using quaternion valued neural network layers, where we aim to preserve the relation between these features through the usage of the Hamilton product.
To train this quaternion neural network, we derive quaternion backpropagation employing the GHR calculus, which is required for a valid product and chain rule in quaternion space. Furthermore, we investigate the connection between the derived update rules and automatic differentiation.
We apply our proposed compression method on the Tennessee Eastman Dataset, where we perform fault classification using the compressed data in two settings: a fully supervised one and in a semi supervised, contrastive learning setting. Both times, we were able to outperform real valued counterparts as well as two baseline models: one with the uncompressed time-series as the input and the other with a regular downsampling using the mean. Further, we could improve the classification benchmark set by SimCLR-TS from $81.43\%$ to $83.90\%$.

%% file: content/01_introduction.tex
\section{Introduction}

\IEEEPARstart{N}{owadays}, an extensive amount of time-series data is widely available, be it from industrial applications, \ac{IoT} devices or wearables. Due to high sampling frequencies of sensors or long recording times, long sequences of data are the consequence. Often, the aim is to process them using machine learning models. However, this gets difficult as recurrent architectures might struggle with longer sequences as it is difficult to squeeze all seen information into a fixed length representation \cite{cho_properties_2014, bahdanau_neural_2014}. 

Also, with more data to process, training times of models increase drastically. Likewise, storage requirements grow and potentially transmission times and energy consumption increases which is especially relevant for \ac{IoT} devices.
The authors of \cite{tucker_model_2021} show that a higher sampling rate as such did not increase the models accuracy while requiring significantly more training time when using a 1d-cnn based model architecture. Similarly, in \cite{nakagome_empirical_2020} downsampling yielded performance improvements in a decoding task for multiple network architecture types.

Naturally, the question whether we can reduce the number of datapoints required for the machine learning task by applying a cheap to compute data compression arises. One common used approach is \ac{PAA} \cite{keogh_dimensionality_2001} which divides a time series in $N$ segments and takes the mean of the $n$ datapoints included in $N$ to form the compressed time series. It is popular for being very intuitive and easy and cheap to compute. However, this does not account for data deviations, as a sine wave with a high amplitude may later appear similar to a constant value. To overcome this downside and enrich the downsampled or compressed time-series with additional descriptive properties, we propose a novel, quaternion based time-compression method. In particular, we propose to divide a longer time-series into smaller chunks from which we extract four representative features, namely the min, max, mean and standard deviation. As this four properties are strongly related, we encapsulate them in the mathematical object of a quaternion.
By utilizing the Hamilton product in quaternion valued one dimensional convolution layer, we aim to preserve their relationship during the feature extraction. While similar approaches have been proposed in the image domain \cite{zhu_quaternion_2018, parcollet_quaternion_2019} where the r-, g- and b-channel of an image have been encoded in a quaternion, our approach is novel to the time-series domain, especially in combination with the quaternionic compression methodology.
It can serve as an easy to implement improvement to already existing, \ac{PAA} based implementations and pipelines without major refactoring, leveraging the strengths of \ac{QNN}, namely relationship preservation and sparsity in comparison to a real valued equivalent.

Following the big success and broad range of applications of real-valued neural networks, in recent years along complex valued models \cite{hirose_complex-valued_2013, trabelsi_deep_2017, benvenuto_complex_1992, ishizuka_modeling_2018, popa_complex-valued_2017, hayakawa_applying_2018} also quaternion valued models \cite{zhu_quaternion_2018, vedaldi_3d-rotation-equivariant_2020, parcollet_quaternion_2016, parcollet_speech_2018, gaudet_deep_2018, onyekpe_quaternion_2021} appealed the interest of researchers and gain more  popularity over time. Applications are e.g. image tasks, 3D point cloud processing, speech / language tasks or sensor fusion. 
Quaternion valued MLPs using elementwise operating activation functions are further proven to be universal interpolators in $\mathbb{H}$ \cite{arena_multilayer_1997}.
To train these quaternion valued neural networks by means of gradient descent, \cite{nitta_quaternary_1995, parcollet_quaternion_2018, matsui_quaternion_2004} introduced quaternion backpropagation variants, using partial derivatives with respect to the four quaternion components. 
However, they do not use the GHR calculus \cite{xu_enabling_2015}, which is required for the validity of the product and chain rule in quaternion space (compare \ref{subsubsec:simple_partial_derivatives}).

In this paper, after introducing the required quaternion algebra, we propose a novel quaternion backpropagation, based on the GHR calculus introduced in \cite{xu_enabling_2015}, considerably extending \cite{xu_optimization_2016}. We provide implementation ready formulas for all parts of the computational chain including intermediate results.
Furthermore, we propose a novel quaternion based time-series compression and test it using the \ac{TE} dataset.

The main contributions of our work are the following:
\begin{itemize}
    \item We propose a novel time-series compression method yielding a shorter, quaternion valued time-series. This compressed time-series is subsequently used for fault classification using a quaternion valued neural network model.
    \item We derive quaternion backpropagation using the GHR calculus which defines proper product and chain rules. We employ detailed calculations for each intermediate result, giving proper insights on the underlying quaternion mathematics.
    \item We investigate the relation of the derived quaternion backpropagation and automatic differentiation, revealing insights on scenarios where it can be used to train \ac{QNN}.
    \item We experimentally prove the effectiveness of the proposed compression and provide detailed comparisons with real-valued counterparts as well as two baseline models in fully supervised learning setups. We further use a state of the art self-supervised learning setup to demonstrate that our approach outperforms previously set classification benchmarks.
\end{itemize}

The paper is structured as follows: Section \ref{sec:related work} presents the related work and Section \ref{sec:fundamentals} the required fundamentals. Section \ref{sec:quaternion_backpropagation} continues with the derived quaternion valued backpropagation algorithm, Section \ref{sec:autoGrad} with the relation of quaternion backpropagation and automatic differentiation and Section \ref{sec:methodology} with the proposed time-series compression methodology. Section \ref{sec:experiments_ts} provides the experimental evaluation. Finally, the paper is concluded by Section \ref{sec:conclusion}.

%% file: content/02_related_work.tex
\section{Related work}
\label{sec:related work}

The related work can mainly be divided into the following categories: time-series compression, backpropagation techniques, especially in the complex or hypercomplex domain and finally applications of quaternion valued neural networks which are described in the following.

\paragraph*{Time-series Compression}
In the field of time-series data compression, it has to be separated between lossless and lossy compression techniques. The former is characterized by the fact that they allow for an error free reconstruction. Examples are SPRINTZ\cite{blalock_sprintz_2018}, DACTS\cite{gomez-brandon_lossless_2021}, RAKE\cite{campobello_rake_2017}, RLBE\cite{spiegel_comparative_2018} or DRH\cite{mogahed_development_2018}.
In contrast, the latter does not enable a perfect reconstruction, but instead  allows for a higher compression rate as lossless techniques are limited in that regard. An often used lossy approach is to approximate the time-series using piece-wise linears \cite{luo_piecewise_2015,yongwei_ding_novel_2010,kitsios_sim-piece_2023} or polynominals \cite{eichinger_time-series_2015,lemire_better_2007,ukil_iot_2015}. Further approaches are e.g. SZ\cite{di_fast_2016} or LFZip\cite{chandak_lfzip_2020}. 
However, the compression approach is fundamentally different as it is not aiming for extracting representative features but instead uses approximations of the original data in a function representation.
Furthermore, the focus here is usually on subsequent reconstruction rather than a fault classification using the compressed data.

The authors of \cite{azar_robust_2020} investigate the effect on the classification performance for univariate and multivariate time-series data when applying compressed sensing, discrete wavelet transform (DWT), the SZ algorithm from \cite{di_fast_2016} and a combination of DWT, lifting scheme and SZ as compression methods. 
In \cite{concha_compressive_2010}, the authors could show that by applying compressive sensing, the classification performance using a hidden Markov model for human activity recognition on video data could often be increased.
Similarly, the authors of \cite{azar_energy_2019} were able to show that the use of a lossy compression does not reduce the prediction quality when using medical data to predict human stress levels using a \ac{MLP} architecture. 
We differ from these works as we propose a quaternion based compression, intended for time-series data and to be processed with the Hamilton product within quaternion valued \ac{NN}, for the first time. Furthermore, contrary to
\cite{azar_robust_2020} and \cite{azar_energy_2019} we directly feed the compressed time-series data to the fault classification model instead of performing a reconstruction first and waive on utilizing the frequency domain as in the former.

Finally, \cite{karamitopoulos_dispersion-based_2009} showed that PAA based approaches can benefit from additionally using the standard deviation rather than just the mean. Likewise, in \cite{lkhagva_new_2006} additionally to the mean, the minimum and maximum values are used. Our proposed methodology can be seen as the fusion of the two approaches within the quaternion space.

\paragraph*{Backpropagation}
The breakthrough of training neural networks using backpropagation started with the well known paper \cite{rumelhart_learning_1986} and was extended and modified in a number of ways in the following. As neural networks emerged from the real numbers to the complex models, equivalent complex variants were introduced in \cite{leung_complex_1991, nitta_extension_1997, la_corte_newtons_2014}. A detailed description of the usage of the CR / Wirtinger Calculus, which comes in handy for complex backpropagation, can be found in \cite{kreutz-delgado_complex_2009}. None of them target quaternion valued models though. Related are also gradient based optimization techniques in the quaternion domain as described in \cite{mandic_quaternion_2011, xu_enabling_2015}.
However, they differ from our work as they don't target multi-layered, perceptron based architectures.

Although, in contrast, these multi-layered, perceptron based architectures are targeted in \cite{nitta_quaternary_1995, parcollet_quaternion_2018, matsui_quaternion_2004}, the respective used approaches suffer from the fact that the chain rule does not hold there. We overcome this by employing the GHR-Calculus and derive the quaternion backpropagation based on this, as initially proposed by \cite{xu_optimization_2016}, and significantly extend it.

\paragraph*{Application of Quaternion Neural Networks}
With regards to quaternion neural network applications, \cite{zhu_quaternion_2018} composes fully quaternion valued models by combining quaternion convolution and quaternion fully-connected layers to use them on  classification and denoising tasks of color images. 
Similarly, \cite{grassucci_quaternion-valued_2021} uses quaternion valued \acp{VAE} for image reconstruction and generation tasks, outperforming real valued \acp{VAE} while simultaneously reducing the number of parameters. The same observation was made by \cite{razavi-far_quaternion_2022} when using quaternion \ac{GAN} for image generation.
The authors of \cite{vedaldi_3d-rotation-equivariant_2020} achieve rotational invariance for point-cloud inputs by utilizing quaternion valued inputs and features, the used weights and biases however are real valued matrices. 
In the context of speech/language understanding, \cite{parcollet_quaternion_2016} employs a quaternion \ac{MLP} approach, where the quaternion model could outperform real valued counterparts while requiring fewer epochs to be trained.
Similarly, \cite{parcollet_speech_2018} utilizes quaternion \acp{CNN} and quaternion \acp{RNN} for this task. Again, these models were able to outperform their real-valued counterparts while requiring approximately four times fewer trainable parameters.
In a related task, \cite{muppidi_speech_2021} employs quaternion \ac{CNN} for emotion recognition in speech.
In \cite{gaudet_deep_2018}, a novel quaternion weight initialization strategies as well as quaternion batch-normalization is proposed to perform image classification and segmentation with quaternion valued models. 
A further application of recurrent quaternion models is \cite{onyekpe_quaternion_2021} where quaternion \acp{RNN}, specifically Quaternion Gated Recurrent Units are applied on sensor fusion for navigation and human activity recognition.
Finally, \cite{qin_fast_2022} proposes quaternion product units and uses them to human action and hand action recognition. Furthermore, they utilize them for point cloud classification.
All these mentioned approaches differ from our work as none is concerned with time series compression of industrial data in the quaternion domain, in particular with a fault classification downstream task where the compressed data serves as the models input.

%% file: content/03_fundamentals.tex
\section{Fundamentals}
\label{sec:fundamentals}

This section introduces the required fundamentals, starting with the mathematics of quaternions, followed by quaternion derivatives and the regular backpropagation algorithm for real valued neural networks. 

\subsection{Quaternions}

\subsubsection{Quaternion Algebra}

Quaternions were discovered by Hamilton in 1843 \cite{hamilton_ii_1844} as a method to extend the complex numbers to the three dimensional space. A quaternion $\quaternion{q}, \quaternion{q} \in \mathbb{H}$ consists of three parts, one real and three imaginary

\begin{equation}
    \quaternion{q} = \quaternionComponents{q} = \quatCompR{q} + \mathbf{q}
\end{equation}

where $\quatCompR{q}, \quatCompI{q}, \quatCompJ{q}, \quatCompK{q} \in \mathbb{R}$. Often, $\quatCompR{q}$ is referred to as the real part and $\mathbf{q}$ as the vector part.
The imaginary components $\imagI, \imagJ, \imagK$ have the properties

\begin{equation}
\begin{aligned}
    \imagI^2 = \imagJ^2 = \imagK^2 = \imagI\imagJ\imagK = -1 \\
    \imagI\imagJ = +\imagK,~~ \imagJ\imagK = +\imagI,~~ \imagK\imagI = +\imagJ \\
    \imagJ\imagI = -\imagK,~~ \imagK\imagJ = -\imagI,~~ \imagI\imagK = -\imagJ .
\end{aligned}
\end{equation}
Similar to complex numbers, quaternions have a conjugate:
\begin{equation}
    \quatConj{q} = \quaternionConjComponents{q} = \quatCompR{q} - \mathbf{q}
\end{equation}
A quaternion $\quaternion{q}$ fulfilling $\left\lVert \quaternion{q} \right\rVert = \sqrt{\quaternion{q}\quatConj{q}} = \sqrt{\quatCompR{q}^2 + \quatCompI{q}^2 + \quatCompJ{q}^2 + \quatCompK{q}^2} = 1$ is called a unit quaternion and a quaternion $\quaternion{q}$ with $\quatCompR{q} = 0$ is called a pure quaternion.

The addition of two quaternions $\quaternion{x}, \quaternion{y}$ is defined as
\begin{equation}
    \quaternion{x} + \quaternion{y} = (\quatCompR{x} + \quatCompR{y}) + (\quatCompI{x} + \quatCompI{y})i + (\quatCompJ{x} + \quatCompJ{y})j + (\quatCompK{x} + \quatCompK{y})k
\end{equation}
and multiplication as 
\begin{equation}
\begin{aligned}
    \quaternion{x} \otimes \quaternion{y} &= \quatCompR{x}\quatCompR{y} - \mathbf{x} \cdot \mathbf{y} +  x_0\mathbf{y}+y_0\mathbf{x} + \mathbf{x} \times \mathbf{y} \\
	&= ( \quatCompR{x} \quatCompR{y} - \quatCompI{x} \quatCompI{y} - \quatCompJ{x} \quatCompJ{y} - \quatCompK{x} \quatCompK{y} )   \\
	&+ ( \quatCompR{x} \quatCompI{y} + \quatCompI{x} \quatCompR{y} + \quatCompJ{x} \quatCompK{y} - \quatCompK{x} \quatCompJ{y} ) \imagI \\
	&+ ( \quatCompR{x} \quatCompJ{y} - \quatCompI{x} \quatCompK{y} + \quatCompJ{x} \quatCompR{y} + \quatCompK{x} \quatCompI{y} ) \imagJ \\
	&+ ( \quatCompR{x} \quatCompK{y} + \quatCompI{x} \quatCompJ{y} - \quatCompJ{x} \quatCompI{y} + \quatCompK{x} \quatCompR{y} ) \imagK .
\end{aligned}
\label{equ:hamilton_product}
\end{equation}

We further define the quaternion Hadamard-product of two quaternions $\quaternion{x}, \quaternion{y}$ as 

\begin{equation}
    \quaternion{x} \circ \quaternion{y} = \quatCompR{x}\quatCompR{y} + \quatCompI{x}\quatCompI{y} \imagI + \quatCompJ{x}\quatCompJ{y} \imagJ + \quatCompK{x} \quatCompK{y} \imagK .
\end{equation}

\subsubsection{Quaternion Involutions}

Of particular importance for the quaternion derivatives are the quaternion self-inverse mappings or involutions, defined as \cite{ell_quaternion_2007} 

\begin{equation}
    \phi(\eta) = \quaternion{q}^\eta = \eta \quaternion{q}\eta ^{-1} = \eta \quaternion{q}\eta^* = -\eta \quaternion{q}\eta
\label{equ:quaternionInvolutionDefinition}
\end{equation}

where $\quaternion{\eta}$ is a pure unit quaternion. Using Equation \eqref{equ:quaternionInvolutionDefinition}, we can create the involutions as

\begin{equation}
\begin{aligned}
    \quaternion{q} &= q_0 + q_1i + q_2j + q_3k \\
    \quatInvI{q} &= -i\quaternion{q}i = q_0 + q_1i - q_2j - q_3k \\
    \quatInvJ{q} &= -j\quaternion{q}j = q_0 - q_1i + q_2j - q_3k \\
    \quatInvK{q} &= -k\quaternion{q}k = q_0 - q_1i - q_2j + q_3k .\\
\end{aligned}
\label{equ:quaternionInvolution1}
\end{equation}
The corresponding conjugate involutions are
\begin{equation}
\begin{aligned}
    \quatConjInvI{q} &= q_0 - q_1i + q_2j + q_3k \\
    \quatConjInvJ{q} &= q_0 + q_1i - q_2j + q_3k \\
    \quatConjInvK{q} &= q_0 + q_1i + q_2j - q_3k .\\
\end{aligned}
\end{equation}

With their help, various relations can be created which come in handy in the following quaternion valued derivation calculations, as they often help in simplifying the calculations and to avoid elaborate term sorting \cite{sudbery_quaternionic_1979, xu_enabling_2015}:  

\begin{equation}
\begin{aligned}
    \quatCompR{q} &= \frac{1}{4} \left( \quaternion{q} \!+\! \quatInvI{q} \!+\! \quatInvJ{q} \!+\! \quatInvK{q} \right), 
    &\quatCompI{q} = -\frac{i}{4} \left( \quaternion{q} \!+\! \quatInvI{q} \!-\! \quatInvJ{q} \!-\! \quatInvK{q} \right) \\
    \quatCompJ{q} &= -\frac{j}{4} \left( \quaternion{q} \!-\! \quatInvI{q} \!+\! \quatInvJ{q} \!-\! \quatInvK{q} \right), 
    &\quatCompK{q} = -\frac{k}{4} \left( \quaternion{q} \!-\! \quatInvI{q} \!-\! \quatInvJ{q} \!+\! \quatInvK{q} \right) \\
\end{aligned}
\label{equ:quaternionInvolution2}
\end{equation}

\begin{equation}
\begin{aligned}
    \quatCompR{q} &= \frac{1}{4} \left( \quatConj{q} \!+\! \quatConjInvI{q} \!+\! \quatConjInvJ{q} \!+\! \quatConjInvK{q} \right) , 
    &\quatCompI{q} = \frac{i}{4} \left( \quatConj{q} \!+\! \quatConjInvI{q} \!-\! \quatConjInvJ{q} \!-\! \quatConjInvK{q} \right) \\
    \quatCompJ{q} &= \frac{j}{4} \left( \quatConj{q} \!-\! \quatConjInvI{q} \!+\! \quatConjInvJ{q} \!-\! \quatConjInvK{q} \right) ,
    &\quatCompK{q} = \frac{k}{4} \left( \quatConj{q} \!-\! \quatConjInvI{q} \!-\! \quatConjInvJ{q} \!+\! \quatConjInvK{q} \right) \\
\end{aligned}
\label{equ:quaternionInvolution3}
\end{equation}

\begin{equation}
\begin{aligned}
    \quatConj{q}     &= \frac{1}{2} \left( -\quaternion{q} \!+\! \quatInvI{q} \!+\! \quatInvJ{q} \!+\! \quatInvK{q} \right) ,
    &\quatConjInvI{q} = \frac{1}{2} \left( \quaternion{q} \!-\! \quatInvI{q} \!+\! \quatInvJ{q} \!+\! \quatInvK{q} \right) \\
    \quatConjInvJ{q} &= \frac{1}{2} \left( \quaternion{q} \!+\! \quatInvI{q} \!-\! \quatInvJ{q} \!+\! \quatInvK{q} \right) ,
    &\quatConjInvK{q} = \frac{1}{2} \left( \quaternion{q} \!+\! \quatInvI{q} \!+\! \quatInvJ{q} \!-\! \quatInvK{q} \right) \\
\end{aligned}
\label{equ:quaternionInvolution4}
\end{equation}

\begin{equation}
\begin{aligned}
    \quaternion{q} &= \frac{1}{2} \left( -\quatConj{q} \!+\! \quatConjInvI{q} \!+\! \quatConjInvJ{q} \!+\! \quatConjInvK{q} \right) ,
    \mkern-14mu&\quatInvI{q}   = \frac{1}{2} \left( \quatConj{q} \!-\! \quatConjInvI{q} \!+\! \quatConjInvJ{q} \!+\! \quatConjInvK{q} \right) \\
    \quatInvJ{q}   &= \frac{1}{2} \left( \quatConj{q} \!+\! \quatConjInvI{q} \!-\! \quatConjInvJ{q} \!+\! \quatConjInvK{q} \right) ,
    \mkern-14mu&\quatInvK{q}   = \frac{1}{2} \left( \quatConj{q} \!+\! \quatConjInvI{q} \!+\! \quatConjInvJ{q} \!-\! \quatConjInvK{q} \right) \\
\end{aligned}
\label{equ:quaternionInvolution5}
\end{equation}

\subsection{Quaternion Derivatives}

\subsubsection{Partial derivatives}
\label{subsubsec:simple_partial_derivatives}

For a quaternion valued function $f(\quaternion{q}), \quaternion{q} = \quaternionComponents{q}$, \cite{nitta_quaternary_1995} and \cite{parcollet_quaternion_2018} calculate the derivatives as follows:

\begin{equation}
    \frac{\partial f}{\partial \quaternion{q}} 
= \left(\frac{\partial f}{\partial \quatCompR{q}} + \frac{\partial f}{\partial \quatCompI{q}}\imagI + \frac{\partial f}{\partial \quatCompJ{q}}\imagJ + \frac{\partial f}{\partial \quatCompK{q}}\imagK \right)
\label{equ:naive_quaternion_derivation}
\end{equation}
However, as we will show now, neither the product rule nor the chain rule hold for this approach.

\begin{proposition}
\label{prop:product_rule}
    For the derivative of a quaternion valued function $f(\quaternion{q}), \quaternion{q} \in \mathbb{H}$ following \eqref{equ:naive_quaternion_derivation}, the product rule does not hold.
\end{proposition}
\begin{proof}
    Consider $f(\quaternion{q}) = \quaternion{q}\quatConj{q} = \quatCompR{q}^2 + \quatCompI{q}^2 + \quatCompJ{q}^2 + \quatCompK{q}^2$ as the function of choice. The direct derivation following \eqref{equ:naive_quaternion_derivation} yields
    \begin{equation}
    \begin{split}
        \frac{\partial f}{\partial \quaternion{q}} 
        &= \left(\frac{\partial f}{\partial \quatCompR{q}} + \frac{\partial f}{\partial \quatCompI{q}}\imagI + \frac{\partial f}{\partial \quatCompJ{q}}\imagJ + \frac{\partial f}{\partial \quatCompK{q}}\imagK \right) \\
        &= (2\quatCompR{q} + 2\quatCompI{q}\imagI + 2\quatCompJ{q}\imagJ +2\quatCompK{q}\imagK ) 
        = 2\quaternion{q}
    \end{split}
    \end{equation}
    Using the product rule, we can calculate the same derivation using

    \begin{equation}
    \begin{aligned}
        \frac{\partial}{\partial \quaternion{q}} \left( \quaternion{q}\quatConj{q} \right)
        &= \quaternion{q} \frac{\partial q^*}{\partial \quaternion{q}} + \frac{\partial \quaternion{q}}{\partial \quaternion{q}}\quatConj{q} \\
        &= \quaternion{q}  \frac{\partial}{\partial \quaternion{q}} \left( \quatCompR{q} - \quatCompI{q} \imagI - \quatCompJ{q} \imagJ - \quatCompK{q} \imagK \right) \\
        &+ \frac{\partial}{\partial \quaternion{q}} \left( \quatCompR{q} + \quatCompI{q} \imagI + \quatCompJ{q} \imagJ + \quatCompK{q} \imagK \right) \quatConj{q} \\
        &= \quaternion{q} \left( 1 - \imagI\imagI - \imagJ\imagJ - \imagK\imagK \right) 
        + \left( 1 + \imagI\imagI + \imagJ\imagJ + \imagK\imagK \right) \quatConj{q} \\
        &= 4\quaternion{q} - 2 \quatConj{q} \neq 2\quaternion{q} .
    \end{aligned}
    \end{equation}
    
\end{proof}

\begin{proposition}
\label{prop:chain_rule}
    For the derivative of a quaternion valued function $f(\quaternion{z}(\quaternion{x}, \quaternion{y}));~ \quaternion{x}, \quaternion{y}, \quaternion{z} \in \mathbb{H}$ following \eqref{equ:naive_quaternion_derivation}, the chain rule does not hold.
\end{proposition}
\begin{proof}
    Consider $f(\quaternion{q}) = \quaternion{z} \quatConj{z}; ~\quaternion{z}=\quaternion{x}\quaternion{y}$ as the function of choice.
    
    We start by calculating the derivative without the chain rule:

    \begin{equation}
    \begin{aligned}
        \frac{\partial f}{\partial \quaternion{x}} &= \frac{\partial}{\partial \quaternion{x}} \Bigl[ (\quaternion{x}\quaternion{y})\conj{(\quaternion{x}\quaternion{y})} \Bigr] \\
        &= \frac{\partial}{\partial x} \Bigl[ \left(\quatCompR{x} + \quatCompI{x} i + \quatCompJ{x} j + \quatCompK{x} k\right) \left(\quatCompR{y} + \quatCompI{y} i + \quatCompJ{y} j + \quatCompK{y} k\right) \\
        &~\left(\quatCompR{y} - \quatCompI{y} i - \quatCompJ{y} j - \quatCompK{y} k\right) \left(\quatCompR{x} - \quatCompI{x} i - \quatCompJ{x} j - \quatCompK{x} k\right) \Bigr] \\
        &= \frac{\partial}{\partial x} \Bigl[ \quatCompR{x}^{2} \quatCompR{y}^{2} + \quatCompR{x}^{2} \quatCompI{y}^{2} 
        + \quatCompR{x}^{2} \quatCompJ{y}^{2} + \quatCompR{x}^{2} \quatCompK{y}^{2} \\
        &+ \quatCompI{x}^{2} \quatCompR{y}^{2} + \quatCompI{x}^{2} \quatCompI{y}^{2} 
        + \quatCompI{x}^{2} \quatCompJ{y}^{2} + \quatCompI{x}^{2} \quatCompK{y}^{2} 
        + \quatCompJ{x}^{2} \quatCompR{y}^{2} + \quatCompJ{x}^{2} \quatCompI{y}^{2}\\ 
        &+ \quatCompJ{x}^{2} \quatCompJ{y}^{2} + \quatCompJ{x}^{2} \quatCompK{y}^{2}  
        + \quatCompK{x}^{2} \quatCompR{y}^{2} + \quatCompK{x}^{2} \quatCompI{y}^{2} 
        + \quatCompK{x}^{2} \quatCompJ{y}^{2} + \quatCompK{x}^{2} \quatCompK{y}^{2} \Bigr] \\
        &= 2 \quatCompR{x} \quatCompR{y}^{2} + 2 \quatCompR{x} \quatCompI{y}^{2} + 2 \quatCompR{x} \quatCompJ{y}^{2} + 2 \quatCompR{x} \quatCompK{y}^{2} \\
        &+ \left(2 \quatCompI{x} \quatCompR{y}^{2} + 2 \quatCompI{x} \quatCompI{y}^{2} + 2 \quatCompI{x} \quatCompJ{y}^{2} + 2 \quatCompI{x} \quatCompK{y}^{2}\right) \imagI \\
        &+ \left(2 \quatCompJ{x} \quatCompR{y}^{2} + 2 \quatCompJ{x} \quatCompI{y}^{2} + 2 \quatCompJ{x} \quatCompJ{y}^{2} + 2 \quatCompJ{x} \quatCompK{y}^{2}\right) \imagJ \\
        &+ \left(2 \quatCompK{x} \quatCompR{y}^{2} + 2 \quatCompK{x} \quatCompI{y}^{2} + 2 \quatCompK{x} \quatCompJ{y}^{2} + 2 \quatCompK{x} \quatCompK{y}^{2}\right) \imagK
    \end{aligned}
    \end{equation}
    Using the chain rule $\frac{\partial f}{\partial \quaternion{x}} = \frac{\partial f}{\partial \quaternion{z}} \frac{\partial \quaternion{z}}{\partial \quaternion{x}}$, the outer derivative yields
    \begin{equation}
    \begin{aligned}
        \frac{\partial f}{\partial \quaternion{z}} &= \frac{\partial}{\partial \quaternion{z}} \Bigl(   \quaternion{z}\quatConj{z} \Bigr)
        = \frac{\partial}{\partial z} \Bigl( \quatCompR{z}^2 + \quatCompI{z}^2 + \quatCompJ{z}^2 + \quatCompK{z}^2 \Bigr) \\
        &= 2(\quatCompR{z} + \quatCompI{z} \imagI + \quatCompJ{z} \imagJ + \quatCompK{z} \imagK) = 2\quaternion{z}
    \end{aligned}
    \end{equation}
    and the inner derivative is
    \begin{equation}
    \begin{aligned}
        \frac{\partial \quaternion{z}}{\partial \quaternion{x}} &= \frac{\partial}{\partial \quaternion{x}} \Bigl( \quaternion{x}\quaternion{y} \Bigr) 
        = \frac{\partial}{\partial \quaternion{x}} \Bigl( \quatCompR{x}
        \quatCompR{y} + \quatCompR{x} \quatCompI{y}\imagI + \quatCompR{x} \quatCompJ{y}\imagJ + \quatCompR{x} \quatCompK{y}\imagK \\
        &+ \quatCompI{x} \quatCompR{y}\imagI- \quatCompI{x} \quatCompI{y} + \quatCompI{x} \quatCompJ{y}\imagK- \quatCompI{x} \quatCompK{y}\imagJ 
        + \quatCompJ{x} \quatCompR{y}\imagJ - \quatCompJ{x} \quatCompI{y}\imagK \\
        &- \quatCompJ{x} \quatCompJ{y} + \quatCompJ{x} \quatCompK{y}\imagI 
        + \quatCompK{x} \quatCompR{y}\imagK + \quatCompK{x} \quatCompI{y}\imagJ- \quatCompK{x} \quatCompJ{y}\imagI- \quatCompK{x} \quatCompK{y} \Bigr) \\
        &= 2(-\quatCompR{y} + \quatCompI{y}\imagI + \quatCompJ{y}\imagJ + \quatCompK{y})\imagK 
        = -2 \quatConj{y} .
    \end{aligned}
    \end{equation}
    Combining inner and outer derivative yields 
    \begin{equation}
    \begin{aligned}    
        \frac{\partial f}{\partial \quaternion{x}} &= \frac{\partial f}{\partial \quaternion{z}} \frac{\partial \quaternion{z}}{\partial \quaternion{x}} 
        = 2\quaternion{z} (-2\quatConj{y}) 
        = -4(\quaternion{x}\quaternion{y}\quatConj{y}) \\
        &= - 4 \quatCompR{x} \quatCompR{y}^{2} - 4 \quatCompR{x} \quatCompI{y}^{2} - 4 \quatCompR{x} \quatCompJ{y}^{2} - 4 \quatCompR{x} \quatCompK{y}^{2} \\
        &+\left(- 4 \quatCompI{x} \quatCompR{y}^{2} - 4 \quatCompI{x} \quatCompI{y}^{2}- 4 \quatCompI{x} \quatCompJ{y}^{2}- 4 \quatCompI{x} \quatCompK{y}^{2} \right)\imagI \\
        &+\left(- 4 \quatCompJ{x} \quatCompR{y}^{2} - 4 \quatCompJ{x} \quatCompI{y}^{2}- 4 \quatCompJ{x} \quatCompJ{y}^{2} - 4 \quatCompJ{x} \quatCompK{y}^{2} \right)\imagJ \\
        &+\left(- 4 \quatCompK{x} \quatCompR{y}^{2} - 4 \quatCompK{x} \quatCompI{y}^{2} - 4 \quatCompK{x} \quatCompJ{y}^{2} - 4 \quatCompK{x} \quatCompK{y}^{2} \right)\imagK
    \end{aligned}
    \end{equation}

    Obviously the results don't match up and hence chain rule does not hold for this way of calculating quaternion derivatives.
\end{proof}

\subsubsection{GHR-Calculus} 

Initial improvements on Quaternion Derivatives yielded the HR-Calculus, however it still lacks the validity of the product- and chain-rule (Compare Appendix \ref{subsec:HrCalculus}).
Hence, \cite{xu_enabling_2015} extends it to the GHR-Calculus, enabling the definition of a novel quaternion product- and chain-rule.  The derivative and conjugate derivative are defined as

\begin{equation}
\begin{aligned}
    \frac{\partial f}{\partial \quaternion{q}^\quaternion{\mu}} = \frac{1}{4}\left(
    \frac{\partial f}{\partial \quatCompR{q}} - 
    \frac{\partial f}{\partial \quatCompI{q}}\imagI^\quaternion{\mu} - 
    \frac{\partial f}{\partial \quatCompJ{q}}\imagJ^\quaternion{\mu}- 
    \frac{\partial f}{\partial \quatCompK{q}}\imagK^\quaternion{\mu}
    \right)
\end{aligned}
\label{equ:ghr_calculus}
\end{equation}

\begin{equation}
\begin{aligned}
    \frac{\partial f}{\partial \quaternion{q}^{\mu*} } = \frac{1}{4}\left(
    \frac{\partial f}{\partial \quatCompR{q}} + 
    \frac{\partial f}{\partial \quatCompI{q}}\imagI^\quaternion{\mu} + 
    \frac{\partial f}{\partial \quatCompJ{q}}\imagJ^\quaternion{\mu} + 
    \frac{\partial f}{\partial \quatCompK{q}}\imagK^\quaternion{\mu}
    \right)
\end{aligned}
\label{equ:ghr_calculus_conjugate}
\end{equation}
whereby $\quaternion{\mu} \neq 0, ~ \quaternion{\mu} \in \mathbb{H}$.

\paragraph{The Conjugate Rule}

For a real valued function $f$, \cite{xu_enabling_2015} define the conjugate rule as
\begin{equation}
    \left(\frac{\partial f}{\partial \quaternion{q}^{\mu}} \right)^*
    =
    \frac{\partial f}{\partial \quaternion{q}^{\mu *}}, ~
    \left(\frac{\partial f}{\partial \quaternion{q}^{\mu *}} \right)^*
    =
    \frac{\partial f}{\partial \quaternion{q}^{\mu}}
\label{equ:conjugate_rule}
\end{equation}

\paragraph{The Product Rule} 

Furthermore, they define the quaternion product rule as 
 
\begin{equation}
    \frac{\partial(fg)}{\partial \quaternion{q}^\quaternion{\mu}} 
    = f \frac{\partial(g)}{\partial \quaternion{q}^\quaternion{\mu}} + \frac{\partial(f)}{\partial \quaternion{q}^{\quaternion{g}\quaternion{\mu}}} g,~
    \frac{\partial(fg)}{\partial \quaternion{q}^{\quaternion{\mu} *}} 
    = f \frac{\partial(g)}{\partial \quaternion{q}^{\quaternion{\mu} *}} + \frac{\partial(f)}{\partial \quaternion{q}^{\quaternion{g}\quaternion{\mu} *}} g .
\end{equation}

\paragraph{The Chain Rule}

Finally, \cite{xu_enabling_2015} also defines a quaterniary chain rule as 

\begin{equation}
    \frac{\partial f(g(q))}{\partial q^\mu} = 
    \frac{\partial f}{\partial g^{\nu}} \frac{\partial g^{\nu}}{\partial q^\mu} + 
    \frac{\partial f}{\partial g^{\nu i}} \frac{\partial g^{\nu i}}{\partial q^\mu} + 
    \frac{\partial f}{\partial g^{\nu j}} \frac{\partial g^{\nu j}}{\partial q^\mu} + 
    \frac{\partial f}{\partial g^{\nu k}} \frac{\partial g^{\nu k}}{\partial q^\mu}
\label{equ:ghr_chain_rule}
\end{equation}
\begin{equation}
    \frac{\partial f(g(q))}{\partial q^{\mu*}} = 
    \frac{\partial f}{\partial g^{\nu}} \frac{\partial g^{\nu}}{\partial q^{\mu*}} + 
    \frac{\partial f}{\partial g^{\nu i}} \frac{\partial g^{\nu i}}{\partial q^{\mu*}} + 
    \frac{\partial f}{\partial g^{\nu j}} \frac{\partial g^{\nu j}}{\partial q^{\mu*}} + 
    \frac{\partial f}{\partial g^{\nu k}} \frac{\partial g^{\nu k}}{\partial q^{\mu*}}
\end{equation}
with $\quaternion{\mu}, \quaternion{\nu} \in \mathbb{H},~ \quaternion{\mu} \quaternion{\nu}  \neq 0$.

Note that unless otherwise stated, in the following we always use $\quaternion{\mu} = \quaternion{\nu} = 1 + 0\imagI + 0\imagJ + 0\imagK$ as this simplifies the notation throughout the calculations.
For further understanding, detailed example calculations can be found in the Appendix \ref{subsec:ExaplesGHR}.

%% file: content/04_backpropagation.tex
\section{Quaternion Backpropagation}
\label{sec:quaternion_backpropagation}

In this section, we derive the backpropagation for \acp{QNN} based on the GHR-Calculus.
We start by defining the forward phase and the quaterniary loss function, then the derivation of the actual backpropagation follows. Specifically, we start with the derivatives of the final layer of a \ac{QNN} and work our way backwards through the architecture and continue with deriving the derivatives for an arbitrary hidden layer, following the usual backpropagation order of sequence.

\subsection{Forward phase and Loss function}

We can formulate the forward phase of a regular feed-forward \ac{QNN} layer $(l)$ with $n$ inputs and $m$ outputs as follows:

\begin{equation}
\begin{aligned}
    \quaternion{a}_i^{(l)} &= \sigma(\quaternion{z}_i^{(l)}),~~ 
    \quaternion{z}_i^{(l)} 
    = \sum_{j=1}^{n} \quaternion{w}_{i,j}^{(l)} \quaternion{a}_j^{(l-1)} + \quaternion{b}_i^{(l)}
\end{aligned}
\end{equation}
where $i \in \{1, \dots, m\}$, $j \in \{1, \dots, n\}$ and $\quaternion{w}, \quaternion{y}, \quaternion{b}, \quaternion{a}, \quaternion{z} \in \mathbb{H}$. 
The corresponding matrix-vector formulation is

\begin{equation}
\begin{aligned}
    \quatVec{a}^{(l)} &= \sigma(\quatVec{z}^{(l)}), ~~ 
    \quatVec{z}^{(l)} 
    = \quatVec{W}^{(l)} \quatVec{a}^{(l-1)} + \quatVec{b}^{(l)}
\end{aligned}
\label{equ:layerMatrixVector}
\end{equation}
where $\quatVec{W} \in \mathbb{H}^{m \times n}$ and $\quatVec{b} \in \mathbb{H}^{m}$ .The final output $\quatVec{y} \in \mathbb{H}^{m}$ of the last layer $L$ and hence the overall model is $\quatVec{a}^{(L)}$.

We define the loss function $\loss(\cdot)$ between the desired output $\quaternion{d}_i$ and the actual output $\quaternion{y}_i$ as
\begin{equation}
    \loss = \quatVec{e}^{*^T} \quatVec{e} 
\end{equation}
whereby $\quaternion{e}_i \in \{1, ..., m \}$ and $\quaternion{e}_i = \quaternion{d}_i - \quaternion{y}_i$. By doing so, we obtain the real valued loss 
\begin{equation}
\begin{aligned}
    \loss &= \quatConj{e}_1 \quaternion{e}_1 + \quatConj{e}_2 \quaternion{e}_2 + \dots + \quatConj{e}_m \quaternion{e}_m  \\
    &= ({\quatCompR{e}}_1^2 + {\quatCompI{e}}_1^2 + {\quatCompJ{e}}_1^2 + {\quatCompK{e}}_1^2)
    + ({\quatCompR{e}}_2^2 + {\quatCompI{e}}_2^2 + {\quatCompJ{e}}_2^2 + {\quatCompK{e}}_2^2) \\
    &+ \dots
    + ({\quatCompR{e}}_m^2 + {\quatCompI{e}}_m^2 + {\quatCompJ{e}}_m^2 + {\quatCompK{e}}_m^2) \\
    &= ({\quatCompR{d}}_1 - {\quatCompR{y}}_1)^2 + ({\quatCompI{d}}_1 - {\quatCompI{y}}_1)^2 \\
    &+ ({\quatCompJ{d}}_1 - {\quatCompJ{y}}_1)^2 + ({\quatCompK{d}}_1 - {\quatCompK{y}}_1)^2 \\
    &+ ({\quatCompR{d}}_2 - {\quatCompR{y}}_2)^2 + ({\quatCompI{d}}_2 - {\quatCompI{y}}_2)^2 \\ 
    &+ ({\quatCompJ{d}}_2 - {\quatCompJ{y}}_2)^2 + ({\quatCompK{d}}_2 - {\quatCompK{y}}_2)^2 \\
    &+ \dots \\
    &+ ({\quatCompR{d}}_m - {\quatCompR{y}}_m)^2 + ({\quatCompI{d}}_m - {\quatCompI{y}}_m)^2 \\ 
    &+ ({\quatCompJ{d}}_m - {\quatCompJ{y}}_m)^2 + ({\quatCompK{d}}_m - {\quatCompK{y}}_m)^2 .
\end{aligned}
\label{equ:loss}
\end{equation}

\subsection{Final layer}
\label{subsec:final_layer}

We start deriving the quaternion backpropagation algorithm by first considering the output of the final layer of a QNN without the usage of an activation function. If an activation function shall be used, one can use intermediate results and apply the strategy from the following Subsection \ref{subsec:hidden_layer}. Recalling Equation \eqref{equ:layerMatrixVector}, the output $\quatVec{y}$ of a \ac{QNN} is calculated as

\begin{equation}
    \quatVec{y} = \quatVec{W}^{(L)} \quatVec{a}^{(L-1)} + \quatVec{b}^{(L)}
\end{equation}
where $\quatVec{a}^{(L-1)}$ is the output of the previous layer. One output quaternion $\quaternion{y}_i$ can be obtained using
\begin{equation}
    \quaternion{y}_i = \sum_{j=1}^{n} \quaternion{w}_{i,j}^{(L)} \quaternion{a}_j^{(L-1)} + \quaternion{b}_i^{(L)} .
\end{equation}

For deriving $\loss$ with respect to the weights $\quatVec{W}^{(L)}$, biases $\quatVec{b}^{(L)}$ and activation outputs of the previous layer $\quatVec{a}^{(L-1)}$, we utilize the chain rule where we first derive $\frac{\partial \loss}{\partial \quatVec{y}}$ and then $\frac{\partial \quatVec{y}}{\partial \conj{\quatVec{W}}}$, $\frac{\partial \quatVec{y}}{\partial \conj{\quatVec{b}}}$ and $\frac{\partial \quatVec{y}}{\partial \quatVec{a}}$.

Note that we use the conjugate $\conj{\quatVec{W}}$ and $\conj{\quatVec{b}}$ as this is the direction of the steepest descent \cite{mandic_quaternion_2011, xu_enabling_2015}. For better readability, we waive on the subscripts $\square_{i,j}$ indicating the matrix/vector elements as well as the superscript $\square^{(L)}$ throughout the following calculations.

\subsubsection{Derivative with respect to the weights}
\label{subsubsec:derivativeWeightsFinalLayer}

\begin{theorem}
    The derivative of $\loss$ with respect to the weights $\quaternion{w}^{(L)}_{i, j}$ of a \ac{QNN} is 
    \begin{equation}
        \frac{\partial \loss(\quaternion{y}(\quaternion{w}, \quaternion{b})}{\partial \quatConj{w}}
        = -\frac{1}{2}\quaternion{e} \quatConj{a}= -\frac{1}{2}(\quaternion{d}-\quaternion{y}) \quatConj{a}.
    \end{equation}
\label{theorem:weight_final_layer}
\end{theorem}

\begin{proof}
The derivative with respect to the weights $\quaternion{w}^{(L)}_{i, j}$ following the GHR-Calculus is calculated as follows:

\begin{equation}
    \frac{\partial \loss(\quaternion{y}(\quaternion{w}, \quaternion{b})}{\partial \quatConj{w}} = 
    \quaternionDerivative{\loss}{\quaternion{y}}{\quatConj{w}}.
\end{equation}
We start by calculating the respective left partial derivatives

\begin{equation}
\begin{aligned}
    \frac{\partial \loss}{\partial \quaternion{y}} \!
    &= \! \frac{\partial }{\partial \quaternion{y}} \left[ (\quatCompR{d} \!-\! \quatCompR{y})^2 + (\quatCompI{d} \!-\! \quatCompI{y})^2 + (\quatCompJ{d} \!-\! \quatCompJ{y})^2 + (\quatCompK{d} \!-\! \quatCompK{y})^2 \right] \\ 
    &= \! \frac{1}{4} \! \left[-2(\quatCompR{d} \!-\! \quatCompR{y}) \!+\! 2(\quatCompI{d} \!-\! \quatCompI{y})\imagI \!+\! 2(\quatCompJ{d} \!-\! \quatCompJ{y})\imagJ \!+\! 2(\quatCompK{d} \!-\! \quatCompK{y})\imagK \right] \\
    &= \! -\frac{1}{2}\left[(\quatCompR{d} \!-\! \quatCompR{y}) \!-\! (\quatCompI{d} \!-\! \quatCompI{y})\imagI \!-\! (\quatCompJ{d} \!-\! \quatCompJ{y})\imagJ \!-\! (\quatCompK{d} \!-\! \quatCompK{y})\imagK \right] \\
    &= \! -\frac{1}{2} (\quaternion{d} - \quaternion{y})^* = -\frac{1}{2} \quatConj{e}
\end{aligned}
\label{equ:leftDerivFinalLayerReal}
\end{equation}
For the derivatives with respect to the involutions $\quatInvI{y}$, $\quatInvJ{y}$ and $\quatInvK{y}$ we just state the final results here, the detailed calculations can be found in the Appendix \ref{subsec:AppFinalLayerLeftDerivatives}.

\begin{equation}
\begin{aligned}
    \frac{\partial \loss}{\partial \quatInvI{y}} \!
    &= \! -\frac{1}{2} \quatConjInvI{(d - y)} = -\frac{1}{2} \quatConjInvI{e}
\end{aligned}
\label{equ:leftDerivFinalLayerI}
\end{equation}

\begin{equation}
\begin{aligned}
    \frac{\partial \loss}{\partial \quatInvJ{y}} \!
    &= \! -\frac{1}{2} \quatConjInvJ{(d - y)} = -\frac{1}{2} \quatConjInvJ{e}
\end{aligned}
\label{equ:leftDerivFinalLayerJ}
\end{equation}

\begin{equation}
\begin{aligned}
    \frac{\partial \loss}{\partial \quatInvK{y} } \mkern-3.5mu
    &= \mkern-3.5mu -\frac{1}{2} \quatConjInvK{(d - y)} = -\frac{1}{2} \quatConjInvK{e}
\end{aligned}
\label{equ:leftDerivFinalLayerK}
\end{equation}
Now we calculate the right partial derivatives of $\quaternion{y}, \quatInvI{y}, \quatInvJ{y}~ \text{and}~\quatInvK{y}$ with respect to $\quatConj{w}$:

\begin{equation}
\begin{aligned}
    \frac{\partial \quaternion{y}}{\partial \quatConj{w}} 
    &= \frac{\partial (\quaternion{w}\quaternion{a} + \quaternion{b})}{\partial \quatConj{w}}
    = \frac{\partial (\quaternion{w}\quaternion{a})}{\partial \quatConj{w}} \\
    &= \frac{\partial}{\partial \quatConj{w}} [
    (\quatCompR{a} \quatCompR{w} - \quatCompI{a} \quatCompI{w} - \quatCompJ{a} \quatCompJ{w} - \quatCompK{a} \quatCompK{w}) \\
    &+ (\quatCompR{a} \quatCompI{w} + \quatCompI{a} \quatCompR{w} - \quatCompJ{a} \quatCompK{w} + \quatCompK{a} \quatCompJ{w} ) \imagI \\
    &+ (\quatCompR{a} \quatCompJ{w} + \quatCompI{a} \quatCompK{w} + \quatCompJ{a} \quatCompR{w} - \quatCompK{a} \quatCompI{w} ) \imagJ \\
    &+ (\quatCompR{a} \quatCompK{w} - \quatCompI{a} \quatCompJ{w} + \quatCompJ{a} \quatCompI{w} + \quatCompK{a} \quatCompR{w} ) \imagK ] \\
    &= \frac{1}{4} [ \quatCompR{a} + \quatCompI{a} \imagI + \quatCompJ{a} \imagJ + \quatCompK{a} k + (\quatCompR{a} \imagI - \quatCompI{a} + \quatCompJ{a} \imagK - \quatCompK{a} \imagJ) \imagI \\
    &+ (\quatCompR{a} \imagJ - \quatCompI{a} \imagK - \quatCompJ{a} + \quatCompK{a} i) \imagJ + (\quatCompR{a} \imagK + \quatCompI{a} \imagJ - \quatCompJ{a} \imagI - \quatCompK{a}) \imagK ] \\
    &= \frac{1}{2} \left[ -\quatCompR{a} + \quatCompI{a} + \quatCompJ{a} + \quatCompK{a} \right] 
    = - \frac{1}{2} \quatConj{a}
\end{aligned}
\label{equ:RightDerivFinalLayerReal}
\end{equation}
For the involutions, again we just state the final results, the detailed calculations can be found in the Appendix \ref{subsec:AppFinalLayerWeights}.
\begin{equation}
    \frac{\partial \quatInvI{y}}{\partial \quatConj{w}} =
    \frac{\partial \quatInvJ{y}}{\partial \quatConj{w}} =
    \frac{\partial \quatInvK{y}}{\partial \quatConj{w}} = 
    \frac{1}{2} \quatConj{a}
\label{equ:RightDerivFinalLayerImag}
\end{equation}
Combining both parts to form the overall derivative yields

\begin{equation}
\begin{aligned}
    \frac{\partial \loss(\quaternion{y}(\quaternion{w}, \quaternion{b})}{\partial \quatConj{w}} 
    &= \frac{-1}{2}\quatConj{e}\frac{-1}{2}\quatConj{a}
    + \frac{-1}{2}\quatConjInvI{e}\frac{1}{2}\quatConj{a} \\
    &+ \frac{-1}{2}\quatConjInvJ{e}\frac{1}{2}\quatConj{a}
    + \frac{-1}{2}\quatConjInvK{e}\frac{1}{2}\quatConj{a} \\
    &= \frac{1}{4} \left[ \quatConj{e} - \quatConjInvI{e} - \quatConjInvJ{e} - \quatConjInvK{e} \right] \quatConj{a} 
    \stackrel{\eqref{equ:quaternionInvolution5}}{=} -\frac{1}{2} \quaternion{e}\quatConj{a}
\end{aligned}
\label{equ:last_layer_error_wrespect_w}
\end{equation}
which concludes the proof
\end{proof}

\subsubsection{Derivative with respect to the bias}

Likewise, we need to derive $\loss$ with respect to the bias $\quaternion{b}_i^{(L)}$.

\begin{theorem}
    The derivative of $\loss$ with respect to the bias $\quaternion{b}_i^{(L)}$ of a \ac{QNN} is
    \begin{equation}
        \frac{\partial \loss(\quaternion{y}(\quaternion{w}, \quaternion{b})}{\partial \quatConj{b}}
        = -\frac{1}{2} \quaternion{e} = - \frac{1}{2}(\quaternion{d}-\quaternion{y}).
    \end{equation}
\label{theorem:bias_final_layer}
\end{theorem}

\begin{proof}
Similarly to the weights, the derivative is defined as follows:

\begin{equation}
    \frac{\partial \loss(\quaternion{y}(\quaternion{w}, \quaternion{b})}{\partial \quatConj{b}} = 
    \quaternionDerivative{\loss}{\quaternion{y}}{\quatConj{b}}.
\end{equation}
The left partial derivatives are already known from Equations \eqref{equ:leftDerivFinalLayerReal} - \eqref{equ:leftDerivFinalLayerK} in \ref{subsubsec:derivativeWeightsFinalLayer}, hence we just have to calculate the right partial derivatives:

\begin{equation}
\begin{aligned}
    \frac{\partial \quaternion{y}}{\partial \quatConj{b}} 
    &= \frac{\partial (\quaternion{w}\quaternion{a} + \quaternion{b})}{\partial \quatConj{b}}
    = \frac{\partial (\quaternion{b})}{\partial \quatConj{b}} \\
    &= \frac{\partial}{\partial \quatConj{b}} ( \quatCompR{b} + \quatCompI{b} \imagI + \quatCompJ{b} \imagJ + \quatCompK{b} \imagK ) \\
    &= \frac{1}{4} \left( 1 + \imagI\imagI + \imagJ\imagJ + \imagK\imagK \right) = - 0.5 \\
\end{aligned}
\label{equ:final_layer_right_biasR}
\end{equation}
For the three involutions $\quatInvI{y}$, $\quatInvJ{y}$ and $\quatInvK{y}$, the detailed calculations can be found in the Appendix \ref{subsec:AppFinalLayerBias} and we just state the results here:

\begin{equation}
    \frac{\partial \quatInvI{y}}{\partial \quatConj{b}} 
    = \frac{\partial \quatInvJ{y}}{\partial \quatConj{b}} 
    = \frac{\partial \quatInvK{y}}{\partial \quatConj{b}} 
    = 0.5
\label{equ:final_layer_right_biasImags}
\end{equation}

Combining the left and right partial derivatives is related to Equation \eqref{equ:last_layer_error_wrespect_w} with the exception that the right derivatives are missing the $a^*$.
Consequently, the final derivative is

\begin{equation}
\begin{aligned}
    \frac{\partial \loss(\quaternion{e}(\quaternion{w}, \quaternion{b})}{\partial \quatConj{b}} = -\frac{1}{2} \quaternion{e} .
\end{aligned}
\end{equation}

\end{proof}

\subsubsection{Derivative with respect to activations}
Finally, we need to derive the loss with respect to the activations / outputs of the previous layer $\quaternion{a}_j^{(L-1)}$. 

\begin{theorem}
    The derivative of $\loss$ with respect to the activations $\quaternion{a}_j^{(L-1)}$ of a \ac{QNN} yields 
    \begin{equation}
        \frac{\partial \loss \left(\quaternion{a}_j^{(L-1)} \right)}{\partial \quaternion{a}_j^{(L-1)}} 
        = \sum_{i \in K} -\frac{1}{2}\quatConj{e}_i \quaternion{w}_{i, j}.
    \end{equation}
    \label{theorem:finalLayerActivations}
\end{theorem}

\begin{proof}

As multiple output neurons $i \in K$ are connected to $\quaternion{a}_j^{(L-1)}$, we have to take the sum of the respective derivatives:

\begin{equation}
\begin{aligned}
    \frac{\partial \loss \! \left(\quaternion{a}_j^{\left(L-1\right)}\right)}{\partial \quaternion{a}_j^{(L-1)}} \!
    &= \! \sum_{i \in K} \!
    \frac{\partial \loss(\quaternion{y}_i^{(l)}(\quaternion{a}_j^{(L-1)}))}{\partial \quaternion{a}_j^{(L-1)}} \\
    &= \! \sum_{i \in K} \! \quaternionDerivativeSmall{\loss}{\quaternion{y}_i}{\quaternion{a}_j} \\
\end{aligned}
\end{equation}
The calculations for the individual parts of the sum are analog to the ones when deriving with respect to the weights, hence we will not list the detailed calculations here but report the results. The detailed derivatives can be found in the Appendix \ref{subsec:AppFinalLayerActivation}.





\begin{equation}
\begin{aligned}
    \frac{\partial \quaternion{y}}{\partial \quaternion{a}} 
    &= \frac{\partial (\quaternion{w}\quaternion{a} + \quaternion{b})}{\partial \quaternion{a}} = \quaternion{w},
    &\frac{\partial \quatInvI{y}}{\partial \quaternion{a}} 
    = \frac{\partial \InvI{(\quaternion{w}\quaternion{a} + \quaternion{b})}}{\partial \quaternion{a}} = 0 \\
    \frac{\partial \quatInvJ{y}}{\partial \quaternion{a}} 
    &= \frac{\partial \InvJ{(\quaternion{w}\quaternion{a} + \quaternion{b})}}{\partial \quaternion{a}} = 0,
    &\frac{\partial \quatInvK{y}}{\partial\quaternion{ }a}
    = \frac{\partial \InvK{(\quaternion{w}\quaternion{a} + \quaternion{b})}}{\partial \quaternion{a}} = 0
\end{aligned}    
\label{equ:right_partial_derivative_a}
\end{equation}
Finally, the derivative formulates as

\begin{equation}
    \frac{\partial \loss(\quaternion{a}_j)}{\partial \quaternion{a}_j} 
    = \sum_{i \in K} -\frac{1}{2}\quatConj{e}_i \quaternion{w}_{i, j} .
\end{equation}
which concludes the proof.
\end{proof}

In the same manner, we can obtain the derivatives
\begin{equation}
\begin{aligned}
    \frac{\partial \loss(\quaternion{a}_j)}{\partial \quatInvI{a}_j} 
    = \sum_{i \in K} -\frac{1}{2} \left( \quatConj{e}_i \quaternion{w}_{i, j} \right)^i \\
    \frac{\partial \loss(\quaternion{a}_j)}{\partial \quatInvJ{a}_j} 
    = \sum_{i \in K} -\frac{1}{2} \left( \quatConj{e}_i \quaternion{w}_{i, j} \right)^j \\
    \frac{\partial \loss(\quaternion{a}_j)}{\partial \quatInvK{a}_j} 
    = \sum_{i \in K} -\frac{1}{2} \left( \quatConj{e}_i \quaternion{w}_{i, j} \right)^k \\
\end{aligned}
\label{equ:right_partial_derivative_a_involutions}
\end{equation}
which we need for applying the chain rule.

\subsubsection{Update rules for the last layer}

Based on the previous calculations, the update rules of weights and biases at timestep $n$ for the last layer are

\begin{equation}
\begin{aligned}
   \quaternion{w}_{i, j}^{(L)}(n + 1) &= \quaternion{w}_{i, j}^{(L)}(n) - \lambda \frac{1}{2} \quaternion{e}_i(n){\quatConj{a}}_j^{(L-1)}(n) \\
   \quaternion{b}_i^{(L)}(n + 1) &= \quaternion{b}_i^{(L)}(n) - \lambda \frac{1}{2} \quaternion{e}_i(n)
\end{aligned}
\end{equation}
whereby $\lambda$ indicates the learning rate.

\subsection{Hidden layer}
\label{subsec:hidden_layer}

For the hidden layers, usually activation functions are used. Thus, we obtain three parts we have to derive for. As we know already from the quaternion chain rule, we cannot simply combine them multiplicatively, especially for the three components. Instead, we first start with deriving with respect to the activation input, where the loss is a function 

\begin{equation}
\loss(\quaternion{a}^{(l)}) = \loss\left(\quaternion{a}^{(l)} \left(\quaternion{z}^{(l)}\right) \right); \quaternion{a}^{(l)}=\sigma \left(\quaternion{z}^{(l)}\right). 
\end{equation}

Then, we can create the involutions of this derivative and continue with deriving with respect to $\quaternion{w}_{i, j}^{(l)}$, $\quaternion{b}_{i}^{(l)}$ and $\quaternion{a}_{i}^{(l-1)}$. For simplicity and better readability, we will avoid the superscript $\square^{(l)}$ indicating the layer when we deal with the involutions to prevent double superscripts.

\subsubsection{Derivative with respect to the activation input}

\begin{theorem}
    The derivative of $\loss$ with respect to the activation input $\quaternion{z}_i^{(l)}$ of a \ac{QNN} is
    \begin{equation}
    \begin{aligned}
        \frac{\partial \loss}{\partial \quaternion{z}_i^{(l)}} 
        &= \quatCompR{p} \sigma^{\prime}(\quatCompR{z}) + \quatCompI{p} \sigma^{\prime}(\quatCompI{z}) \imagI + \quatCompJ{p} \sigma^{\prime}(\quatCompJ{z}) \imagJ + \quatCompK{p} \sigma^{\prime}(\quatCompK{z}) \imagK \\&= \quaternion{p} \circ \sigma^{\prime}(\quaternion{z})     
    \end{aligned}
    \end{equation}
    whereby $\quaternion{p}$ is the derivative of the next layer and $\sigma(\cdot)$ is an elementwise activation function.
\label{theorem:FinalLayerActivationInput}
\end{theorem}

\begin{proof}

For deriving with respect to the activation input $\quaternion{z}_i^{(l)}$ we have to calculate

\begin{equation}
    \frac{\partial \loss\left(\quaternion{a}^{(l)} (\quaternion{z}^{(l)})\right)}{\partial \quaternion{z}^{(l)}}
    = \quaternionDerivative{\loss}{\quaternion{a}}{\quaternion{z}}.
\end{equation}
We can derive the outer equation with respect to both, the regular quaternion or it's conjugate, and here it's more convenient and simultaneously also computationally more efficient to choose the regular quaternion as $\quaternion{z}$ is a direct result of the forward phase while $\quatConj{z}$ is not.

To obtain the derivatives with respect to the involutions $\quatInvI{a},~\quatInvJ{a}~\text{and}~\quatInvK{a}$ we can simply take the known result and flip the signs of the imaginary parts according to Equation \eqref{equ:hr_calculus}.
For the last layer we know $\frac{\partial \loss}{\partial \quaternion{a}}$ already. For an arbitrary hidden layer $(l)$ we don't know it yet. Hence, for better readability and generalization, in the following we will call the result simply $\quaternion{p}^{(l+1)}$. 
As this is coming from the following layer we assign the superscript $\square^{(l+1)}$ Furthermore, this naming is also convenient since the result will change from the last layer to the hidden layer, but we can always refer to it as $\quaternion{p}$.

For the right part of the partial derivatives, we first need to obtain the derivatives of $\quaternion{a}(z)$ and it's involutions $\quatInvI{a}(z),~\quatInvJ{a}(z)\text{and}~\quatInvK{a}(z)$. Using an element-wise operating activation function $\sigma(\cdot)$ to calculate $\quaternion{a}^{(l)} = \sigma(\quaternion{z}) = \sigma(\quatCompR{z}) + \sigma(\quatCompI{z}) \imagI + \sigma(\quatCompJ{z}) \imagJ + \sigma(\quatCompK{z}) \imagK$ these are

\begin{equation}
\begin{aligned}
    \frac{\partial \quaternion{a}}{\partial \quaternion{z}} &= \frac{\partial}{\partial \quaternion{z}}
    \left[ \sigma(\quatCompR{z}) + \sigma(\quatCompI{z}) \imagI + \sigma(\quatCompJ{z}) \imagJ + \sigma(\quatCompK{z}) \imagK \right] \\
    &=\frac{1}{4} \left[\sigma^{\prime}(\quatCompR{z}) + \sigma^{\prime}(\quatCompI{z}) + \sigma^{\prime}(\quatCompJ{z}) + \sigma^{\prime}(\quatCompK{z}) \right]
\end{aligned}
\end{equation}
\begin{equation}
\begin{aligned}
    \frac{\partial \quatInvI{a}}{\partial \quaternion{z}} &= \frac{\partial}{\partial \quaternion{z}}
    \left[ \sigma(\quatCompR{z}) + \sigma(\quatCompI{z}) \imagI - \sigma(\quatCompJ{z}) \imagJ - \sigma(\quatCompK{z}) \imagK \right] \\
    &=\frac{1}{4} \left[\sigma^{\prime}(\quatCompR{z}) + \sigma^{\prime}(\quatCompI{z}) - \sigma^{\prime}(\quatCompJ{z}) - \sigma^{\prime}(\quatCompK{z}) \right]
\end{aligned}
\end{equation}
\begin{equation}
\begin{aligned}
    \frac{\partial \quatInvJ{a}}{\partial \quaternion{z}} &= \frac{\partial}{\partial \quaternion{z}}
    \left[ \sigma(\quatCompR{z}) - \sigma(\quatCompI{z}) \imagI + \sigma(\quatCompJ{z}) \imagJ - \sigma(\quatCompK{z}) \imagK \right] \\
    &=\frac{1}{4} \left[\sigma^{\prime}(\quatCompR{z}) - \sigma^{\prime}(\quatCompI{z}) + \sigma^{\prime}(\quatCompJ{z}) - \sigma^{\prime}(\quatCompK{z}) \right]
\end{aligned}
\end{equation}
\begin{equation}
\begin{aligned}
    \frac{\partial \quatInvK{a}}{\partial \quaternion{z}} &= \frac{\partial}{\partial \quaternion{z}}
    \left[ \sigma(\quatCompR{z}) - \sigma(\quatCompI{z}) \imagI - \sigma(\quatCompJ{z}) \imagJ + \sigma(\quatCompK{z}) \imagK \right] \\
    &=\frac{1}{4} \left[\sigma^{\prime}(\quatCompR{z}) - \sigma^{\prime}(\quatCompI{z}) - \sigma^{\prime}(\quatCompJ{z}) + \sigma^{\prime}(\quatCompK{z}) \right]
\end{aligned}
\end{equation}
Combining the respective partial derivatives yields

\begin{equation}
\begin{aligned}
    \frac{\partial \loss}{\partial z^{(l)}} 
    &= \frac{\partial \loss\left(a^{(l)} (z^{(l)})\right)}{\partial z^{(l)}} \\
    &= \quaternion{p} \frac{1}{4} \left[\sigma^{\prime}(\quatCompR{z}) + \sigma^{\prime}(\quatCompI{z}) + \sigma^{\prime}(\quatCompJ{z}) + \sigma^{\prime}(\quatCompK{z}) \right] \\
    &+ \quatInvI{p} \frac{1}{4} \left[\sigma^{\prime}(\quatCompR{z}) + \sigma^{\prime}(\quatCompI{z}) - \sigma^{\prime}(\quatCompJ{z}) - \sigma^{\prime}(\quatCompK{z}) \right] \\
    &+ \quatInvJ{p} \frac{1}{4} \left[\sigma^{\prime}(\quatCompR{z}) - \sigma^{\prime}(\quatCompI{z}) + \sigma^{\prime}(\quatCompJ{z}) - \sigma^{\prime}(\quatCompK{z}) \right] \\
    &+ \quatInvK{p} \frac{1}{4} \left[\sigma^{\prime}(\quatCompR{z}) - \sigma^{\prime}(\quatCompI{z}) - \sigma^{\prime}(\quatCompJ{z}) + \sigma^{\prime}(\quatCompK{z}) \right] \\ 
    &= \frac{1}{4} \left[\quaternion{p} + \quatInvI{p} + \quatInvJ{p} +\quatInvK{p} \right] \sigma^{\prime}(\quatCompR{z}) \\
    &+ \frac{1}{4} \left[\quaternion{p} + \quatInvI{p} - \quatInvJ{p} -\quatInvK{p} \right] \sigma^{\prime}(\quatCompI{z}) \\
    &+ \frac{1}{4} \left[\quaternion{p} - \quatInvI{p} + \quatInvJ{p} -\quatInvK{p} \right] \sigma^{\prime}(\quatCompJ{z}) \\
    &+ \frac{1}{4} \left[\quaternion{p} - \quatInvI{p} - \quatInvJ{p} +\quatInvK{p} \right] \sigma^{\prime}(\quatCompK{z}) \\
    &= \quatCompR{p} \sigma^{\prime}(\quatCompR{z}) + \quatCompI{p} \sigma^{\prime}(\quatCompI{z}) \imagI + \quatCompJ{p} \sigma^{\prime}(\quatCompJ{z}) \imagJ + \quatCompK{p} \sigma^{\prime}(\quatCompK{z}) \imagK \\
    &= \quaternion{p} \circ \sigma^{\prime}(\quaternion{z})
\end{aligned}
\end{equation}
%
which concludes the proof.
\end{proof}

Note that in the last step we used the alternative representations
\begin{equation}
\begin{aligned}
    \quatCompR{q}        &= \frac{1}{4} \left( \quaternion{q} + \quatInvI{q} + \quatInvJ{q} + \quatInvK{q} \right),
    &\quatCompI{q} \imagI = \frac{1}{4} \left( \quaternion{q} + \quatInvI{q} - \quatInvJ{q} - \quatInvK{q} \right) \\
    \quatCompJ{q} \imagJ &= \frac{1}{4} \left( \quaternion{q} - \quatInvI{q} + \quatInvJ{q} - \quatInvK{q} \right),
    &\quatCompK{q} \imagK = \frac{1}{4} \left( \quaternion{q} - \quatInvI{q} - \quatInvJ{q} + \quatInvK{q} \right) \\
\end{aligned}
\end{equation}
of \eqref{equ:quaternionInvolution2} which can be obtained by multiplying with $\{1, \imagI, \imagJ, \imagK\}$ respectively.
Furthermore, we can safely assume $\frac{\partial \loss}{\partial \quatInvI{a}} = \quatInvI{p}$, $\frac{\partial \loss}{\partial \quatInvJ{a}} = \quatInvJ{p}$ and $\frac{\partial \loss}{\partial \quatInvK{a}} = \quatInvK{p}$ due to Equations \eqref{equ:right_partial_derivative_a_involutions} and following \eqref{equ:hidden_layer_derivative_a_involutions}.
As required for the chain rule, we can further obtain
\begin{equation}
\begin{aligned}
    \frac{\partial \loss}{\partial \quatInvI{z}_i} 
    &= \left(\quaternion{p} \circ \sigma^{\prime}(\quaternion{z}) \right)^i\\
    \frac{\partial \loss}{\partial \quatInvJ{z}_i} 
    &= \left(\quaternion{p} \circ \sigma^{\prime}(\quaternion{z}) \right)^k\\
    \frac{\partial \loss}{\partial \quatInvK{z}_i} 
    &= \left(\quaternion{p} \circ \sigma^{\prime}(\quaternion{z}) \right)^k 
\end{aligned}
\end{equation}
following the same strategy.

\subsubsection{Derivative with respect to the weights}
Now we can continue with deriving the loss function with respect to the weights of a hidden layer. 

\begin{theorem}
    The derivative of $\loss$ with respect to the weights of a hidden layer
    $\quaternion{w}^{(l)}_{i, j}$
    of a \ac{QNN} yields 
    \begin{equation}
        \frac{\partial \loss(w)}{\partial \quaternion{w}^{*^{(l)}}}
        = \quatConj{q} \quatConj{a}
    \end{equation}
    whereby $\quaternion{q}$ is the derivative of $\loss$ with respect to the following activation function.
    \label{theorem:hidden_layer_weights}
\end{theorem}

\begin{proof}

For the derivative with respect to the hidden layers weights $\quaternion{w}^{(l)}_{i, j}$, we need to calculate

\begin{equation}
    \frac{\partial \loss(\quaternion{z}^{(l)}(\quaternion{w}^{(l)}))}{\partial \quaternion{w}^{*^{(l)}}}
    = \quaternionDerivative{\loss}{\quaternion{z}}{\quatConj{w}}.
\end{equation}

Just like in the case above, we assign the name $\frac{\partial \loss}{\partial \quaternion{z}^{(l)}} = \quaternion{q}$ for the result.
Then, we get the following partial derivatives for the involutions:

\begin{equation}
    \frac{\partial \quaternion{y}}{\partial \quatInvI{z}} = \quatInvI{q},~
    \frac{\partial \quaternion{y}}{\partial \quatInvJ{z}} = \quatInvJ{q},~
    \frac{\partial \quaternion{y}}{\partial \quatInvK{z}} = \quatInvK{q} 
\end{equation}
The right partial derivatives with respect to the weights are already known from Equations \eqref{equ:RightDerivFinalLayerReal} and \eqref{equ:RightDerivFinalLayerImag} in Subsection \ref{subsec:final_layer}, namely 
\begin{equation}
\begin{aligned}
    \frac{\partial \quaternion{z}}{\partial \quatConj{w}} = -\frac{1}{2}\quatConj{a},~~
    \frac{\partial \quatInvI{z}}{\partial \quatConj{w}} =  \frac{1}{2}\quatConj{a},~~
    \frac{\partial \quatInvJ{z}}{\partial \quatConj{w}} =  \frac{1}{2}\quatConj{a},~~
    \frac{\partial \quatInvK{z}}{\partial \quatConj{w}} =  \frac{1}{2}\quatConj{a}
\end{aligned}
\end{equation}
whereby $\quaternion{z}=\quaternion{z}_i^{(l)}$, $\quaternion{w}=\quaternion{w}_{i,j}^{(l)}$ and $\quaternion{a}=\quaternion{a}_i^{(l-1)}$. Consequently, the full derivation is 
\begin{equation}
\begin{aligned}
    \frac{\partial \loss(\quaternion{w})}{\partial \quatConj{w}} &= \frac{\partial \loss(\quaternion{z}(\quaternion{w}))}{\partial\quatConj{w}} 
    = \quaternion{q} \frac{-1}{2}\quatConj{a} + \quatInvI{q}\frac{1}{2}\quatConj{a} + \quatInvJ{q}\frac{1}{2}\quatConj{a} + \quatInvK{q}\frac{1}{2}\quatConj{a} \\
    &= \frac{1}{2} \left(-\quaternion{q} + \quatInvI{q} + \quatInvJ{q} + \quatInvK{q} \right) \quatConj{a} 
    \stackrel{\eqref{equ:quaternionInvolution4}}{=} \quatConj{q} \quatConj{a} 
\end{aligned}
\end{equation}
which concludes the proof. 
\end{proof}

\subsubsection{Derivative with respect to the biases}
Likewise, we need the derivative of the loss function with respect to the biases of a hidden layer

\begin{theorem}
    The derivative of $\loss$ with respect to the bias of a hidden layer 
    $\quaternion{b}_i^{(l)}$
    of a \ac{QNN} yields  
    \begin{equation}
        \frac{\partial \loss(\quaternion{b})}{\partial \quaternion{b}^{*^{(l)}}}
        = \quatConj{q}
    \end{equation}
    whereby $\quaternion{q}$ is the derivative of $\loss$ with respect to the following activation function.
\label{theorem:hidden_layer_bias}        
\end{theorem}

\begin{proof}

In the same manner, we can also compute the derivation with respect to the bias $\quaternion{b}_i^{(l)}$

\begin{equation}
    \frac{\partial \loss(\quaternion{z}^{(l)}(\quaternion{b}^{(l)}))}{\partial \quaternion{b}^{*^{(l)}}}
    = \quaternionDerivative{\loss}{\quaternion{z}}{\quatConj{b}}.
\end{equation}
Using the known results
\begin{equation}
\begin{aligned}
    \frac{\partial \quaternion{z}}{\partial \quatConj{b}} = -\frac{1}{2},~~
    \frac{\partial \quatInvI{z}}{\partial \quatConj{b}} =  \frac{1}{2},~~
    \frac{\partial \quatInvJ{z}}{\partial \quatConj{b}} =  \frac{1}{2},~~
    \frac{\partial \quatInvK{z}}{\partial \quatConj{b}} =  \frac{1}{2}
\end{aligned}
\end{equation}
from Equations \eqref{equ:final_layer_right_biasR} and \eqref{equ:final_layer_right_biasImags} the combined derivative yields
\begin{equation}
\begin{aligned}
    \frac{\partial \loss(\quaternion{b})}{\partial \quatConj{b}} &= \frac{\partial \loss(\quaternion{z}(\quaternion{b}))}{\partial \quatConj{b}} 
    = \quaternion{q} \frac{-1}{2} + \quatInvI{q}\frac{1}{2} + \quatInvJ{q}\frac{1}{2} + \quatInvK{q}\frac{1}{2} \\
    &= \frac{1}{2} \left(-\quaternion{q} + \quatInvI{q} + \quatInvJ{q} + \quatInvK{q} \right) 
    \stackrel{\eqref{equ:quaternionInvolution4}}{=} \quatConj{q} .
\end{aligned}
\end{equation}
which concludes the proof.
\end{proof}

\subsubsection{Derivative with respect to the activations of the previous layer}

Finally, we need to derive with respect to the activation of the previous layer $\quaternion{a}_j^{(l-1)}$. Again, the right partial derivatives are already known from Equation \eqref{equ:right_partial_derivative_a}.
As in the regular backpropagation, in this case we need to consider all neurons $K$ where a respective activation $\quaternion{a}_j^{(l-1)}$ is input to.

\begin{theorem}
    The derivative of a hidden layer of a \ac{QNN} with respect to its inputs, the activations $\quaternion{a}_j^{(l-1)}$ of a previous layer, is 
    \begin{equation}
        \frac{\partial \loss \left(\quaternion{a}_j^{\left(l-1\right)}\right)}{\partial \quaternion{a}_j^{(l-1)}} 
        = \sum_{i \in K} \quaternion{q}_i \quaternion{w}_{i, j}
    \end{equation}
    whereby $\quaternion{q}$ is the derivative of $\loss$ with respect to the following activation function
\label{theorem:hidden_layer_activations}
\end{theorem}

\begin{proof}

Just as in Theorem \ref{theorem:finalLayerActivations} we have to take the sum of the respective derivatives:

\begin{equation}
\begin{aligned}
    \frac{\partial \loss \! \left(\quaternion{a}_j^{\left(l-1\right)}\right)}{\partial \quaternion{a}_j^{(l-1)}} 
    &= \! \sum_{i \in K}
    \frac{\partial \loss(\quaternion{z}_i^{(l)}(\quaternion{a}_j^{(l-1)}))}{\partial \quaternion{a}_j^{(l-1)}} \\
    &= \! \sum_{i \in K} \quaternionDerivativeSmall{\loss}{\quaternion{z}_i}{\quaternion{a}_j} \\
\end{aligned}
\end{equation}
By using the known derivatives from Equation \eqref{equ:right_partial_derivative_a} we obtain the final derivation
\begin{equation}
    \frac{\partial \loss \left(\quaternion{a}_j^{\left(l-1\right)}\right)}{\partial \quaternion{a}_j^{(l-1)}}
    = \! \sum_{i \in K} \quaternion{q}_i \quaternion{w}_{i, j}
\end{equation}
which concludes the proof.
\end{proof}

Likewise, we can obtain
\begin{equation}
\begin{aligned}
    \frac{\partial \loss(\quaternion{a}_j)}{\partial \quatInvI{a}_j} 
    &= \sum_{i \in K} \left( \quaternion{a}_i \quaternion{w}_{i, j} \right)^i \\
    \frac{\partial \loss(\quaternion{a}_j)}{\partial \quatInvJ{a}_j} 
    &= \sum_{i \in K} \left( \quaternion{q}_i \quaternion{w}_{i, j} \right)^j \\
    \frac{\partial \loss(\quaternion{a}_j)}{\partial \quatInvK{a}_j} 
    &= \sum_{i \in K} \left( \quaternion{q}_i \quaternion{w}_{i, j} \right)^k .
\end{aligned}
\label{equ:hidden_layer_derivative_a_involutions}
\end{equation}

Note that the superscripts $\square^{\imagI, \imagJ, \imagK}$ indicate the quaternion involutions and the subscripts $\square_{i, j}$ the indices within the neural network. 
This result for a hidden layer $(l)$ is then the starting point when calculating the derivatives for optimizing the previous layer $(l-1)$ (compare Theorem \ref{theorem:FinalLayerActivationInput}).

\subsubsection{Update rules for the hidden layer}

Using the derived results, we can formulate the update rules for the parameters in the hidden layers as

\begin{equation}
\begin{aligned}
    \quaternion{w}_{i,j}^{(l)}(n + 1) &= \quaternion{w}_{i,j}^{(l)}(n) \!-\! \lambda \Bigl(
      {\quatCompR{p}}_i^{(l+1)} \sigma^{\prime}({\quatCompR{z}}_i^{(l)}) 
    \!+\! {\quatCompI{p}}_i^{(l+1)} \sigma^{\prime}({\quatCompI{z}}_i^{(l)}) \imagI \\
    &+ {\quatCompJ{p}}_i^{(l+1)} \sigma^{\prime}({\quatCompJ{z}}_i^{(l)}) \imagJ 
    + {\quatCompK{p}}_i^{(l+1)} \sigma^{\prime}({\quatCompK{z}}_i^{(l)}) \imagK
    \conj{\Bigr)} {\quatConj{a}}_j^{(l-1)} \\
    &= \quaternion{w}_{i,j}^{(l)}(n) - \lambda \left(
    \quaternion{p}_i^{(l)} \circ \sigma^{\prime}(\quaternion{z}_i^{(l)})  
    \right) {\quatConj{a}}_j^{(l-1)}\\
    \quaternion{b}_i^{(l)}(n + 1) &= \quaternion{b}_i^{(l)}(n) \!-\! \lambda \Bigl(
      {\quatCompR{p}}_i^{(l+1)} \sigma^{\prime}({\quatCompR{z}}_i^{(l)}) 
    \!+\! {\quatCompI{p}}_i^{(l+1)} \sigma^{\prime}({\quatCompI{z}}_i^{(l)}) \imagI \\
    &+ {\quatCompJ{p}}_i^{(l+1)} \sigma^{\prime}({\quatCompJ{z}}_i^{(l)}) \imagJ 
    + {\quatCompK{p}}_i^{(l+1)} \sigma^{\prime}({\quatCompK{z}}_i^{(l)}) \imagK
    \conj{\Bigr)} \\
    &= \quaternion{b}_i^{(l)}(n) - \lambda \conj{\left(
    \quaternion{p}_i^{(l)} \circ \sigma^{\prime}(\quaternion{z}_i^{(l)})
    \right)} .
\end{aligned}
\end{equation}
Furthermore, the new $\quaternion{p}_j^{(l)}$ becomes


\begin{equation}
\begin{aligned}
    \quaternion{p}_j^{(l)} 
    &= \sum_{i \in K} \Bigl( 
      {\quatCompR{p}}_i^{(l+1)} \sigma^{\prime}({\quatCompR{z}}_i^{(l)}) 
    + {\quatCompI{p}}_i^{(l+1)} \sigma^{\prime}({\quatCompI{z}}_i^{(l)}) \imagI \\
    &+ {\quatCompJ{p}}_i^{(l+1)} \sigma^{\prime}({\quatCompJ{z}}_i^{(l)}) \imagJ 
    + {\quatCompK{p}}_i^{(l+1)} \sigma^{\prime}({\quatCompK{z}}_i^{(l)}) \imagK
    \Bigr)\quaternion{w}_{i, j}^{(l)} \\
    &= \sum_{i \in K} 
    \left(\quaternion{p}_i^{(l+1)} \circ \sigma^{\prime}(\quaternion{z}_i^{(l)})\right) \quaternion{w}_{i, j}^{(l)} .
\end{aligned}
\end{equation}

Thus, all required derivatives to optimize a \ac{QNN} beginning with the last layer backward to the first layer are derived and the quaternion backpropagation is complete.

%% file: content/04-2_autoGrad.tex
\section{Quaternion Backpropagation and Automatic Differentiation}
\label{sec:autoGrad}

As shown by \cite{valle_understanding_2023}, current deep learning libraries can be used to implement hypercomplex neural networks when using elementwise activation functions, e.g. by utilizing the Kronecker product. These libraries usually make use of automatic differentiation, allowing to train the implemented hypercomplex models without the necessity to define the respective derivatives. Naturally, the question arises how the relation is between automatic differentiation derivatives and the GHR derivatives. Consequently, this subsection investigates if automatic differentiation can be used to implement and train \ac{QNN} following the derived optimization. We specifically target the automatic differentiation in PyTorch \cite{PyTorch}, although similar considerations could be made for other packages. For details on the automatic differentiation in PyTorch, we refer to \cite{paszke2017automatic}.

\subsection{The relation of the GHR chain rule and chain rule in automatic differentiation}

We start by making a general statement about the chain rule in automatic differentiation:

\newcommand{\pd}[2]{
    \frac{\partial #1}{\partial #2}
}

\begin{theorem}
    Assume a function $f(\quaternion{y}(\quaternion{x}))$ and the derivative $\pd{f}{\quaternion{y}}$ to be $\quaternion{s} = \quaternionComponents{s}$. Then the derivative when applying the chain rule on the individual quaternion elements $\quatCompR{y}$, $\quatCompI{y}$, $\quatCompJ{y}$ and $\quatCompK{y}$ is 
    \begin{equation}
        \pd{f}{\quaternion{x}} =
        \left(\quatConj{s} \pd{\quaternion{y}}{\quaternion{x}}
        + \quatConjInvI{s} \pd{\quatInvI{y}}{\quaternion{x}}
        + \quatConjInvJ{s} \pd{\quatInvJ{y}}{\quaternion{x}}
        + \quatConjInvK{s} \pd{\quatInvK{y}}{\quaternion{x}}\right)^* .
    \end{equation}
\label{theorem:ghr-autograd-relation}
\end{theorem}
\newcommand{\test}[5]{
    \pd{\quatCompR{#1}}{#2} #3 
    \pd{\quatCompI{#1}}{#2} \imagI #4 
    \pd{\quatCompJ{#1}}{#2} \imagJ #5 
    \pd{\quatCompK{#1}}{#2} \imagK
}
\begin{proof}
    As all input elements can be involved in the calculation of the respective output elements $\quatCompR{y}$, $\quatCompI{y}$, $\quatCompJ{y}$ and $\quatCompK{y}$, when calculating the derivative with respect to $\quatCompR{x}$, $\quatCompI{x}$, $\quatCompJ{x}$ and $\quatCompK{x}$ all output elements have to be linked with the derivative components $\quatCompR{s}$, $\quatCompI{s}$, $\quatCompJ{s}$ and $\quatCompK{s}$ to apply the chain rule on a component level. Consequently, the derivative formulates as

    \begin{align}
        &\pd{f}{\quaternion{x}} = \pd{f(\quaternion{y}(\quaternion{x}))}{\quaternion{x}} \\
        &= 
        \quatCompR{s}\pd{\quatCompR{y}}{\quatCompR{x}} +
        \quatCompI{s}\pd{\quatCompI{y}}{\quatCompR{x}} +
        \quatCompJ{s}\pd{\quatCompJ{y}}{\quatCompR{x}} + 
        \quatCompK{s}\pd{\quatCompK{y}}{\quatCompR{x}} \\
        &+ \left(
        \quatCompR{s}\pd{\quatCompR{y}}{\quatCompI{x}} +
        \quatCompI{s}\pd{\quatCompI{y}}{\quatCompI{x}} +
        \quatCompJ{s}\pd{\quatCompJ{y}}{\quatCompI{x}} + 
        \quatCompK{s}\pd{\quatCompK{y}}{\quatCompI{x}}
        \right) \imagI \\
        &+ \left(
        \quatCompR{s}\pd{\quatCompR{y}}{\quatCompJ{x}} +
        \quatCompI{s}\pd{\quatCompI{y}}{\quatCompJ{x}} +
        \quatCompJ{s}\pd{\quatCompJ{y}}{\quatCompJ{x}} + 
        \quatCompK{s}\pd{\quatCompK{y}}{\quatCompJ{x}}
        \right) \imagJ \\
        &+ \left(
        \quatCompR{s}\pd{\quatCompR{y}}{\quatCompK{x}} +
        \quatCompI{s}\pd{\quatCompI{y}}{\quatCompK{x}} +
        \quatCompJ{s}\pd{\quatCompJ{y}}{\quatCompK{x}} + 
        \quatCompK{s}\pd{\quatCompK{y}}{\quatCompK{x}}
        \right) \imagK \\
        &\stackrel{\eqref{equ:quaternionInvolution3}}{=}
        \frac{1}{4}\!\left(\quatConj{s} + \quatConjInvI{s} + \quatConjInvJ{s} + \quatConjInvK{s} \right) \!
        \left(
            \pd{\quatCompR{y}}{\quatCompR{x}} +
            \pd{\quatCompR{y}}{\quatCompI{x}} \imagI + 
            \pd{\quatCompR{y}}{\quatCompJ{x}} \imagJ +
            \pd{\quatCompR{y}}{\quatCompK{x}} \imagK
        \right) \\
        &+
        \frac{\imagI}{4}\!\left(\quatConj{s} + \quatConjInvI{s} - \quatConjInvJ{s} - \quatConjInvK{s} \right) \!
        \left(
            \pd{\quatCompI{y}}{\quatCompR{x}} +
            \pd{\quatCompI{y}}{\quatCompI{x}} \imagI + 
            \pd{\quatCompI{y}}{\quatCompJ{x}} \imagJ +
            \pd{\quatCompI{y}}{\quatCompK{x}} \imagK
        \right) \\
        &+
        \frac{\imagJ}{4}\!\left(\quatConj{s} - \quatConjInvI{s} + \quatConjInvJ{s} - \quatConjInvK{s} \right) \!
        \left(
            \pd{\quatCompJ{y}}{\quatCompR{x}} +
            \pd{\quatCompJ{y}}{\quatCompI{x}} \imagI + 
            \pd{\quatCompJ{y}}{\quatCompJ{x}} \imagJ +
            \pd{\quatCompJ{y}}{\quatCompK{x}} \imagK
        \right) \\
        &+
        \frac{\imagK}{4}\!\left(\quatConj{s} - \quatConjInvI{s} - \quatConjInvJ{s} + \quatConjInvK{s} \right) \!
        \left(
            \pd{\quatCompK{y}}{\quatCompR{x}} +
            \pd{\quatCompK{y}}{\quatCompI{x}} \imagI + 
            \pd{\quatCompK{y}}{\quatCompJ{x}} \imagJ +
            \pd{\quatCompK{y}}{\quatCompK{x}} \imagK
        \right) \\
        &= \quatConj{s} \frac{1}{4} \biggl[ 
            \left(
                \pd{\quatCompR{y}}{\quatCompR{x}} +
                \pd{\quatCompR{y}}{\quatCompI{x}} \imagI + 
                \pd{\quatCompR{y}}{\quatCompJ{x}} \imagJ +
                \pd{\quatCompR{y}}{\quatCompK{x}} \imagK
            \right) \\
           &\hphantom{\quatConj{s} \frac{1}{4}} \, + \imagI
            \left(
                \pd{\quatCompI{y}}{\quatCompR{x}} +
                \pd{\quatCompI{y}}{\quatCompI{x}} \imagI + 
                \pd{\quatCompI{y}}{\quatCompJ{x}} \imagJ +
                \pd{\quatCompI{y}}{\quatCompK{x}} \imagK
            \right) \\
            &\hphantom{\quatConj{s} \frac{1}{4}} \,+ \imagJ
            \left(
                \pd{\quatCompJ{y}}{\quatCompR{x}} +
                \pd{\quatCompJ{y}}{\quatCompI{x}} \imagI + 
                \pd{\quatCompJ{y}}{\quatCompJ{x}} \imagJ +
                \pd{\quatCompJ{y}}{\quatCompK{x}} \imagK
            \right) \\
            &\hphantom{\quatConj{s} \frac{1}{4}} \,+ \imagK
            \left(
                \pd{\quatCompK{y}}{\quatCompR{x}} +
                \pd{\quatCompK{y}}{\quatCompI{x}} \imagI + 
                \pd{\quatCompK{y}}{\quatCompJ{x}} \imagJ +
                \pd{\quatCompK{y}}{\quatCompK{x}} \imagK
            \right)
        \biggr] \\
        &+ \quatConjInvI{s} \frac{1}{4} \biggl[ 
            \left(
                \pd{\quatCompR{y}}{\quatCompR{x}} +
                \pd{\quatCompR{y}}{\quatCompI{x}} \imagI + 
                \pd{\quatCompR{y}}{\quatCompJ{x}} \imagJ +
                \pd{\quatCompR{y}}{\quatCompK{x}} \imagK
            \right) \\
            &\hphantom{\quatConjInvI{s} \frac{1}{4}} \,+ \imagI
            \left(
                \pd{\quatCompI{y}}{\quatCompR{x}} +
                \pd{\quatCompI{y}}{\quatCompI{x}} \imagI + 
                \pd{\quatCompI{y}}{\quatCompJ{x}} \imagJ +
                \pd{\quatCompI{y}}{\quatCompK{x}} \imagK
            \right) \\
            &\hphantom{\quatConjInvI{s} \frac{1}{4}} \,- \imagJ
            \left(
                \pd{\quatCompJ{y}}{\quatCompR{x}} +
                \pd{\quatCompJ{y}}{\quatCompI{x}} \imagI + 
                \pd{\quatCompJ{y}}{\quatCompJ{x}} \imagJ +
                \pd{\quatCompJ{y}}{\quatCompK{x}} \imagK
            \right) \\
            &\hphantom{\quatConjInvI{s} \frac{1}{4}} \,- \imagK
            \left(
                \pd{\quatCompK{y}}{\quatCompR{x}} +
                \pd{\quatCompK{y}}{\quatCompI{x}} \imagI + 
                \pd{\quatCompK{y}}{\quatCompJ{x}} \imagJ +
                \pd{\quatCompK{y}}{\quatCompK{x}} \imagK
            \right)
        \biggr] \\
        &+ \quatConjInvJ{s} \frac{1}{4} \biggl[ 
           \left(
                \pd{\quatCompR{y}}{\quatCompR{x}} +
                \pd{\quatCompR{y}}{\quatCompI{x}} \imagI + 
                \pd{\quatCompR{y}}{\quatCompJ{x}} \imagJ +
                \pd{\quatCompR{y}}{\quatCompK{x}} \imagK
            \right) \\
           &\hphantom{\quatConjInvJ{s} \frac{1}{4}} - \imagI
            \left(
                \pd{\quatCompI{y}}{\quatCompR{x}} +
                \pd{\quatCompI{y}}{\quatCompI{x}} \imagI + 
                \pd{\quatCompI{y}}{\quatCompJ{x}} \imagJ +
                \pd{\quatCompI{y}}{\quatCompK{x}} \imagK
            \right) \\
            &\hphantom{\quatConjInvJ{s} \frac{1}{4}} + \imagJ
            \left(
                \pd{\quatCompJ{y}}{\quatCompR{x}} +
                \pd{\quatCompJ{y}}{\quatCompI{x}} \imagI + 
                \pd{\quatCompJ{y}}{\quatCompJ{x}} \imagJ +
                \pd{\quatCompJ{y}}{\quatCompK{x}} \imagK
            \right) \\
            &\hphantom{\quatConjInvJ{s} \frac{1}{4}} - \imagK
            \left(
                \pd{\quatCompK{y}}{\quatCompR{x}} +
                \pd{\quatCompK{y}}{\quatCompI{x}} \imagI + 
                \pd{\quatCompK{y}}{\quatCompJ{x}} \imagJ +
                \pd{\quatCompK{y}}{\quatCompK{x}} \imagK
            \right)
        \biggr] \\
        &+ \quatConjInvK{s} \frac{1}{4} \biggl[ 
            \left(
                \pd{\quatCompR{y}}{\quatCompR{x}} +
                \pd{\quatCompR{y}}{\quatCompI{x}} \imagI + 
                \pd{\quatCompR{y}}{\quatCompJ{x}} \imagJ +
                \pd{\quatCompR{y}}{\quatCompK{x}} \imagK
            \right) \\
           &\hphantom{\quatConjInvK{s} \frac{1}{4}} - \imagI
           \left(
                \pd{\quatCompI{y}}{\quatCompR{x}} +
                \pd{\quatCompI{y}}{\quatCompI{x}} \imagI + 
                \pd{\quatCompI{y}}{\quatCompJ{x}} \imagJ +
                \pd{\quatCompI{y}}{\quatCompK{x}} \imagK
            \right) \\
            &\hphantom{\quatConjInvK{s} \frac{1}{4}} - \imagJ
            \left(
                \pd{\quatCompJ{y}}{\quatCompR{x}} +
                \pd{\quatCompJ{y}}{\quatCompI{x}} \imagI + 
                \pd{\quatCompJ{y}}{\quatCompJ{x}} \imagJ +
                \pd{\quatCompJ{y}}{\quatCompK{x}} \imagK
            \right) \\
            &\hphantom{\quatConjInvK{s} \frac{1}{4}} + \imagK
            \left(
                \pd{\quatCompK{y}}{\quatCompR{x}} +
                \pd{\quatCompK{y}}{\quatCompI{x}} \imagI + 
                \pd{\quatCompK{y}}{\quatCompJ{x}} \imagJ +
                \pd{\quatCompK{y}}{\quatCompK{x}} \imagK
            \right)
        \biggr] \\
        &=\quatConj{s} \frac{1}{4} \left[
            \pd{\quaternion{y}}{\quatCompR{x}} + 
            \pd{\quaternion{y}}{\quatCompI{x}} \imagI + 
            \pd{\quaternion{y}}{\quatCompJ{x}} \imagJ + 
            \pd{\quaternion{y}}{\quatCompK{x}} \imagK
        \right] \\
        &+\quatConjInvI{s} \frac{1}{4} \left[
            \pd{\quatInvI{y}}{\quatCompR{x}} + 
            \pd{\quatInvI{y}}{\quatCompI{x}} \imagI + 
            \pd{\quatInvI{y}}{\quatCompJ{x}} \imagJ + 
            \pd{\quatInvI{y}}{\quatCompK{x}} \imagK
        \right] \\
        &+\quatConjInvJ{s} \frac{1}{4} \left[
            \pd{\quatInvJ{y}}{\quatCompR{x}} + 
            \pd{\quatInvJ{y}}{\quatCompI{x}} \imagI + 
            \pd{\quatInvJ{y}}{\quatCompJ{x}} \imagJ + 
            \pd{\quatInvJ{y}}{\quatCompK{x}} \imagK
        \right] \\
        &+\quatConjInvK{s} \frac{1}{4} \left[
            \pd{\quatInvK{y}}{\quatCompR{x}} + 
            \pd{\quatInvK{y}}{\quatCompI{x}} \imagI + 
            \pd{\quatInvK{y}}{\quatCompJ{x}} \imagJ + 
            \pd{\quatInvK{y}}{\quatCompK{x}} \imagK
        \right] \\
        &= \quatConj{s} \pd{\quaternion{y}}{\quatConj{x}}
        + \quatConjInvI{s} \pd{\quatInvI{y}}{\quatConj{x}}
        + \quatConjInvJ{s} \pd{\quatInvJ{y}}{\quatConj{x}}
        + \quatConjInvK{s} \pd{\quatInvK{y}}{\quatConj{x}} . 
    \end{align}
\end{proof}
In addition to this Theorem, we can further establish the following useful Corollaries:
\begin{corollary}
    For a real valued function $f(\quaternion{y}(\quaternion{x}))$ and $\quaternion{\mu}, \quaternion{\nu} \in \mathbb{H} = 1$ where $\pd{f}{\quaternion{y}} = \quaternion{q}$, the GHR chain rule 
    \begin{equation}
        \pd{f}{\quaternion{x}} = 
        \pd{f}{\quaternion{y}} \pd{\quaternion{y}}{\quaternion{x}}
        + \pd{f}{\quatInvI{y}} \pd{\quatInvI{y}}{\quaternion{x}}
        + \pd{f}{\quatInvJ{y}} \pd{\quatInvJ{y}}{\quaternion{x}}
        + \pd{f}{\quatInvK{y}} \pd{\quatInvK{y}}{\quaternion{x}}
    \end{equation}
    simplifies to 
    \begin{equation}
        \pd{f}{\quaternion{x}} = 
        \quaternion{q} \pd{\quaternion{y}}{\quaternion{x}}
        + \quatInvI{q} \pd{\quatInvI{y}}{\quaternion{x}}
        + \quatInvJ{q} \pd{\quatInvJ{y}}{\quaternion{x}}
        + \quatInvK{q} \pd{\quatInvK{y}}{\quaternion{x}} .
    \end{equation}
\end{corollary}
\begin{proof}
The relation 
\begin{equation}
    \left(\pd{f}{\quaternion{q}} \right)^\eta
    = \pd{f^\eta}{\quaternion{q}^\eta} 
\end{equation}
holds $\forall \eta \in \{\imagI, \imagJ, \imagK\}$ \cite{xu_enabling_2015}. For a real valued $f$, 
\begin{equation}
    f^\imagI = f^\imagJ = f^\imagK = -\imagI f \imagI = -\imagJ f \imagJ = -\imagK f \imagK = f
\end{equation}
and the corollary follows.
\end{proof}
This simplification is used from now on.

\begin{corollary}
    Consider the function $f(\quaternion{y}(\quaternion{x}))$ with $\pd{f}{\quaternion{y}} = \quaternion{q}$, $\pd{f}{\quatInvI{y}} = \quatInvI{q}$, $\pd{f}{\quatInvJ{y}} = \quatInvJ{q}$ and $\pd{f}{\quatInvK{y}} = \quatInvK{q}$ where the derivative $\pd{f}{\quaternion{x}}$ following the chain rule of the GHR calculus is 
    \begin{equation}
        \pd{f}{\quaternion{x}} = 
        \quaternion{q} \pd{\quaternion{y}}{\quaternion{x}}
        + \quatInvI{q} \pd{\quatInvI{y}}{\quaternion{x}}
        + \quatInvJ{q} \pd{\quatInvJ{y}}{\quaternion{x}}
        + \quatInvK{q} \pd{\quatInvK{y}}{\quaternion{x}} .
    \end{equation}
    Then, when applying the chain rule on the quaternion components, the derivative in automatic differentiation (AD) is
    \begin{equation}
        {\pd{f}{\quaternion{x}}}_{AD} = \left( {\pd{f}{\quaternion{x}}}_{GHR} \right)^*
    \end{equation}
    when
    \begin{equation}
        {\pd{f}{\quaternion{y}}}_{AD} = \left({\pd{f}{\quaternion{y}}}_{GHR}\right)^*.
    \end{equation}
\label{cor:conjugate_input-conjugate_output}
\end{corollary}

\begin{proof}
    The proof directly follows from Theorem \ref{theorem:ghr-autograd-relation}
    when choosing $\quaternion{s} = \quatConj{q}$ and applying \eqref{equ:conjugate_rule}.
\end{proof}

To put that Corollary in simple words, one can conclude that the output derivative when applying the chain rule in automatic differentiation is the conjugate of the GHR derivative, as long as the input derivative is also the conjugate.

\begin{corollary}
    Consider the function $f(\quaternion{y}(\quaternion{x}))$ with $\pd{f}{\quaternion{y}} = \quaternion{q}$, $\pd{f}{\quatInvI{y}} = \quatInvI{q}$, $\pd{f}{\quatInvJ{y}} = \quatInvJ{q}$ and $\pd{f}{\quatInvK{y}} = \quatInvK{q}$ where the derivative $\pd{f}{\quatConj{x}}$ following the chain rule of the GHR calculus is 
    \begin{equation}
        \pd{f}{\quatConj{x}} = 
        \quaternion{q} \pd{\quaternion{y}}{\quatConj{x}}
        + \quatInvI{q} \pd{\quatInvI{y}}{\quatConj{x}}
        + \quatInvJ{q} \pd{\quatInvJ{y}}{\quatConj{x}}
        + \quatInvK{q} \pd{\quatInvK{y}}{\quatConj{x}} .
    \end{equation}
    Then, the relation 
    \begin{equation}
        {\pd{f}{\quaternion{x}}}_{AD} = {\pd{f}{\quatConj{x}}}_{GHR}
    \end{equation}
    holds, when 
    \begin{equation}
        {\pd{f}{\quaternion{y}}}_{AD} = \left({\pd{f}{\quaternion{y}}}_{GHR}\right)^*.
    \end{equation}
\label{cor:deriving_for_conjugate}
\end{corollary}

\begin{proof}
    The proof also directly follows from Theorem \ref{theorem:ghr-autograd-relation} when choosing $\quaternion{s} = \quatConj{q}$.
\end{proof}

From this, we can state that the derivatives for GHR and automatic differentiation are the same, when the variable deriving for in automatic differentiation is the conjugate of the one in GHR calculus and the derivative connected using the chain rule is also the conjugate.
This is of particular interest as for the direction of steepest descent in the GHR calculus one needs to derive with respect to $\conj{\quatVec{W}}$ and $\conj{\quatVec{b}}$ \cite{mandic_quaternion_2011, xu_enabling_2015} but when deriving for the individual components, we derive for $\quatCompR{w}$, $\quatCompI{w}$, $\quatCompJ{w}$, $\quatCompK{w}$, $\quatCompR{b}$, $\quatCompI{b}$, $\quatCompJ{b}$ and $\quatCompK{w}$ respectively, and thus not the conjugate.

In the following, we will continue with deriving the loss function as well as the derivatives with respect to $\quaternion{w}_{i,j}$, $\quaternion{b}_{i}$ and $\quaternion{a}_{j}$ for the automatic differentiation. Specifically, Corollary \ref{cor:conjugate_input-conjugate_output} will be used together with the derivative with respect to the loss function for the gradients flowing backward through the model, and  Corollary \ref{cor:deriving_for_conjugate} for the derivatives with respect to the parameters.

\subsection{Forward phase and Loss function}

Following \cite{valle_understanding_2023}, the real valued emulation of a quaternion linear layer with $n$ inputs and $m$ outputs is calculated as 
\begin{equation}
    \begin{bmatrix}
        \mathbf{w}_{1, 1} & \mathbf{w}_{1, 2} & \cdots & \mathbf{w}_{1, n} \\
        \mathbf{w}_{2, 1} & \mathbf{w}_{2, 2} & \cdots & \mathbf{w}_{2, n} \\
        \vdots            & \vdots            & \ddots & \vdots            \\
        \mathbf{w}_{m, 1} & \mathbf{w}_{m, 2} & \cdots & \mathbf{w}_{m, n} \\
    \end{bmatrix}
    \begin{bmatrix}
        \mathbf{a}_1 \\
        \mathbf{a}_2 \\
        \vdots  \\
        \mathbf{a}_n \\
    \end{bmatrix}
    +
    \begin{bmatrix}
        \mathbf{b}_1 \\
        \mathbf{b}_2 \\
        \vdots  \\
        \mathbf{b}_m \\
    \end{bmatrix}
    = 
    \begin{bmatrix}
        \mathbf{z}_1 \\
        \mathbf{z}_2 \\
        \vdots  \\
        \mathbf{z}_m \\
    \end{bmatrix}
\end{equation}

where $\mathbf{w}_{i,j}$ is the real valued matrix representation of the quaternion $\quaternion{w}_{i,j}$ and $\mathbf{a}_{j}$, $\mathbf{b}_{i}$, $\mathbf{z}_{i}$ the respective vector representations of $\quaternion{a}_{i}$, $\quaternion{b}_{i}$ and $\quaternion{z}_{i}$. 

Accordingly, one partial calculation is carried out as follows:
\begin{equation}
    \mathbf{z}_i = \mathbf{w}_{i,j}\mathbf{a}_j + \mathbf{b}_i = 
    \begin{bmatrix}
        \quatCompR{w} & -\quatCompI{w} & -\quatCompJ{w} & -\quatCompK{w} \\
        \quatCompI{w} &  \quatCompR{w} & -\quatCompK{w} &  \quatCompJ{w} \\
        \quatCompJ{w} &  \quatCompK{w} &  \quatCompR{w} & -\quatCompI{w} \\
        \quatCompK{w} & -\quatCompJ{w} &  \quatCompI{w} &  \quatCompR{w} \\
    \end{bmatrix}
    \begin{bmatrix}
        \quatCompR{a} \\
        \quatCompI{a} \\
        \quatCompJ{a} \\
        \quatCompK{a} \\
    \end{bmatrix}
    +
    \begin{bmatrix}
        \quatCompR{b} \\
        \quatCompI{b} \\
        \quatCompJ{b} \\
        \quatCompK{b} \\
    \end{bmatrix}
\end{equation}
The loss calculation as shown in Equation \eqref{equ:loss} remains unchanged. 

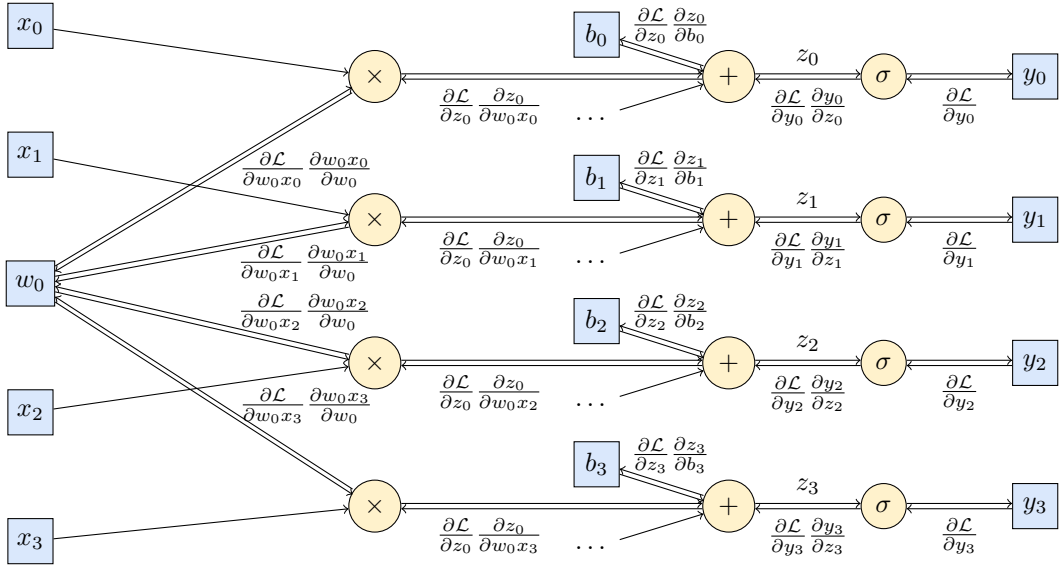
\begin{figure*}[htb]
    \centering
    \input{images/autoDiff.tex}
    \caption{Visualization of the gradient flow}
    \label{fig:gradFlow}
\end{figure*}

Figure \ref{fig:gradFlow} further shows the information and gradient flow for an arbitrary hidden layer $(l)$. It is based on a visualization of a quaternion linear layer implementation in PyTorch\cite{PyTorch} which we generated using PyTorchViz\cite{TorchViz}, shown in Figure \ref{fig:torchViz} in Appendix \ref{sec:layer_visu}. For simplicity, we just show target $\quatCompR{w}$ although the principle stays the same for remaining weights $\quatCompI{w}$, $\quatCompJ{w}$ and $\quatCompK{w}$. For the final layer, $\sigma$ is simply the identity function.

We again start by deriving the loss with respect to $\quaternion{y}_i$ and specifically to its components $\quatCompR{y}, \quatCompI{y}, \quatCompJ{y}$ and $\quatCompK{y}$. However, this time we do not compose them back in a quaternion but keep them as a vector, as the automatic differentiation does treat the outputs as completely independent of each other. Hence, we don't have derivatives as in Equations \eqref{equ:naive_quaternion_derivation}, \eqref{equ:ghr_calculus} or \eqref{equ:ghr_calculus_conjugate}. Instead, we get the simple derivatives 

\begin{equation}
    \begin{bmatrix}
        \frac{\partial \loss}{\partial \quatCompR{y}} \\[4pt]
        \frac{\partial \loss}{\partial \quatCompI{y}} \\[4pt]
        \frac{\partial \loss}{\partial \quatCompJ{y}} \\[4pt]
        \frac{\partial \loss}{\partial \quatCompK{y}} \\[4pt]
    \end{bmatrix}
    = -2
    \begin{bmatrix}
        \left(\quatCompR{d} - \quatCompR{y} \right) \\
        \left(\quatCompI{d} - \quatCompI{y} \right) \\
        \left(\quatCompJ{d} - \quatCompJ{y} \right) \\
        \left(\quatCompK{d} - \quatCompK{y} \right) \\
    \end{bmatrix}
     = -2 
     \begin{bmatrix}
        \quatCompR{e} \\
        \quatCompI{e} \\
        \quatCompJ{e} \\
        \quatCompK{e} \\
     \end{bmatrix} 
\label{equ:deriv_loss_autograd}
\end{equation}

which is the vector representation of $-2\quaternion{e}_i$. We can observe that the derivative with respect to the models output $\quaternion{y}_i$ is the conjugate of the derivative obtained with the GHR calculus and greater by a factor of 4.

Note that in the calculations above we omitted the superscript $\square^{(L)}$ indicating the layer and the subscripts $\square_{i,j}$ indicating the row and column for better readability and to avoid confusion with the subscript  indicating the quaternion component. This is also what we are going to do in the following whenever possible.

\subsection{Derivative with respect to the inputs of the activation functions}

When considering elementwise activation functions $\quaternion{a}^{(l)} = \sigma(\quaternion{z}) = \sigma(\quatCompR{z}) + \sigma(\quatCompI{z}) \imagI + \sigma(\quatCompJ{z}) \imagJ + \sigma(\quatCompK{z}) \imagK$, the derivatives in automatic differentiation are straight forward

\begin{equation}
    \begin{bmatrix}
        \frac{\partial \quatCompR{a}}{\partial \quatCompR{z}} \\[4pt]
        \frac{\partial \quatCompI{a}}{\partial \quatCompI{z}} \\[4pt]
        \frac{\partial \quatCompJ{a}}{\partial \quatCompJ{z}} \\[4pt]
        \frac{\partial \quatCompK{a}}{\partial \quatCompK{z}} \\[4pt]
    \end{bmatrix}
    =
    \begin{bmatrix}
        \frac{\partial}{\partial \quatCompR{z}} \sigma(\quatCompR{z}) \\[4pt]
        \frac{\partial}{\partial \quatCompI{z}} \sigma(\quatCompI{z}) \\[4pt]
        \frac{\partial}{\partial \quatCompJ{z}} \sigma(\quatCompJ{z}) \\[4pt]
        \frac{\partial}{\partial \quatCompK{z}} \sigma(\quatCompK{z}) \\[4pt]
    \end{bmatrix}
    =
    \begin{bmatrix}
        \sigma^{\prime}(\quatCompR{z}) \\[4pt]
        \sigma^{\prime}(\quatCompI{z}) \\[4pt]
        \sigma^{\prime}(\quatCompJ{z}) \\[4pt]
        \sigma^{\prime}(\quatCompK{z}) \\[4pt]
    \end{bmatrix} .
\end{equation}

Thus, as in Theorem \ref{theorem:FinalLayerActivationInput}, the derivatives with respect to the activations are an elementwise multiplication with the derivative coming from the following layer $(l + 1)$ during the backward phase. They don't change anything in terms of conjugation and scaling.

\newcommand{\lossVector}[1]{
    \begin{bmatrix}    
        \frac{\partial \loss}{\partial \quatCompR{#1}} \\[4pt]
        \frac{\partial \loss}{\partial \quatCompI{#1}} \\[4pt]
        \frac{\partial \loss}{\partial \quatCompJ{#1}} \\[4pt]
        \frac{\partial \loss}{\partial \quatCompK{#1}} \\[4pt]
    \end{bmatrix}        
}
\newcommand{\autoGradDeriv}[2]{
    \frac{\partial \loss}{\partial \quatCompR{#1}} \frac{\partial \quatCompR{#1}}{\partial #2} + 
    \frac{\partial \loss}{\partial \quatCompI{#1}} \frac{\partial \quatCompI{#1}}{\partial #2} + 
    \frac{\partial \loss}{\partial \quatCompJ{#1}} \frac{\partial \quatCompJ{#1}}{\partial #2} + 
    \frac{\partial \loss}{\partial \quatCompK{#1}} \frac{\partial \quatCompK{#1}}{\partial #2}}
\newcommand{\autoGradDerivSubscript}[3]{
    \frac{\partial \loss}{\partial \quatCompSubscriptR{#1}{#3}} \frac{\partial \quatCompSubscriptR{#1}{#3}}{\partial #2} + 
    \frac{\partial \loss}{\partial \quatCompSubscriptI{#1}{#3}} \frac{\partial \quatCompSubscriptI{#1}{#3}}{\partial #2} + 
    \frac{\partial \loss}{\partial \quatCompSubscriptJ{#1}{#3}} \frac{\partial \quatCompSubscriptJ{#1}{#3}}{\partial #2} + 
    \frac{\partial \loss}{\partial \quatCompSubscriptK{#1}{#3}} \frac{\partial \quatCompSubscriptK{#1}{#3}}{\partial #2}}

\subsection{Derivative with respect to the activation output of the previous layer}

\begin{proposition}
    Assume the derivative $\frac{\partial \loss}{\partial \quaternion{z}}$ with respect to the output $\quaternion{z}^{(l)}$ of a quaternion linear layer $(l)$ to be $\quatConj{q} = \quaternionConjComponents{q}$. Then, the derivative with respect to its input $\quaternion{a}_j^{(l-1)}$ in automatic differentiation is 
    \begin{equation}
        \frac{\partial \loss \left(\quaternion{a}_j^{\left(l-1\right)}\right)}{\partial \quaternion{a}_j^{(l-1)}} 
        = \sum_{i \in K} \left( \quaternion{q}_i \quaternion{w}_{i, j} \right)^*
    \end{equation}
    and hence the conjugate as for the GHR derivatives with $\frac{\partial \loss}{\partial \quaternion{z}} = \quaternion{q} = \quaternionComponents{q}$.
\label{theo:autograd_activation_previous_layer}
\end{proposition}

\begin{proof}
    The proof follows directly from Theorem \ref{theorem:hidden_layer_activations} and Corollary \ref{cor:conjugate_input-conjugate_output}.
\end{proof}

Based on this, and the fact that an elementwise activation does not change anything in terms of conjugation, we can conclude that the gradient flow backwards through the whole model is the conjugate of the GHR gradients as long as the derivative with respect to the loss-function in automatic differentiation is the conjugate of the GHR derivative. From Equation \eqref{equ:deriv_loss_autograd} we know that this is the case. Furthermore, this result has the consequence that the gradient is also bigger by a factor of four. The detailed calculations are further shown in the Appendix \ref{subsec:autograd_derivative_activation}.

\subsection{Derivative with respect to the weights}

\begin{proposition}
    Assume the derivative $\frac{\partial \loss}{\partial \quaternion{z}}$ with respect to the output $\quaternion{z}^{(l)}_i$ of a quaternion linear layer $(l)$ to be $\quatConj{q} = \quaternionConjComponents{q}$. Then, the derivative with respect to its weights $\quaternion{w}^{(l)}_{i,j}$ in automatic differentiation is 
    \begin{equation}
        \frac{\partial \loss(w)}{\partial \quaternion{w}^{(l)}}
        = \quatConj{q} \quatConj{a}
    \end{equation}
    and hence the same as for the GHR derivatives with $\frac{\partial \loss}{\partial \quaternion{z}} = \quaternion{q} = \quaternionComponents{q}$.
\end{proposition}

\begin{proof}
    The proof directly follows from Theorem \ref{theorem:hidden_layer_weights} and Corollary \ref{cor:deriving_for_conjugate}.
\end{proof}
For the detailed calculations, compare with Appendix \ref{subsec:autograd_derivative_weight}.
If we now insert the results from Equation \eqref{equ:deriv_loss_autograd}, we obtain 
\begin{equation}
    \frac{\partial \loss(w)}{\partial \quaternion{w}^{(L)}}
    = -2\quaternion{e} \quatConj{a} .
\end{equation}
As we can see, the weight updates for the final layer are four times as big as in the GHR Calculus (compare Theorem \ref{theorem:weight_final_layer}). This also holds for arbitrary hidden layers due to Proposition \ref{theo:autograd_activation_previous_layer}.

\subsection{Derivative with respect to the bias}

\begin{proposition}
    Assume the derivative $\frac{\partial \loss}{\partial \quaternion{z}}$ with respect to the output $\quaternion{z}^{(l)}_1$ of a quaternion linear layer $(l)$ to be $\quatConj{q} = \quaternionConjComponents{q}$. Then, the derivative with respect to its bias $\quaternion{b}^{(l)}_i$ in automatic differentiation is 
    \begin{equation}
        \frac{\partial \loss(b)}{\partial \quaternion{b}^{(l)}}
        = \quatConj{q}
    \end{equation}
    and hence the same as for the GHR derivatives with $\frac{\partial \loss}{\partial \quaternion{z}} = \quaternion{q} = \quaternionComponents{q}$.
\end{proposition}

\begin{proof}
    The proof directly follows from Theorem \ref{theorem:hidden_layer_bias} and Corollary \ref{cor:deriving_for_conjugate}.
\end{proof}

Again, the detailed calculations are shown in Appendix \ref{subsec:autograd_derivative_bias}.
Also here, if we insert the results from Equation \eqref{equ:deriv_loss_autograd}, we obtain
\begin{equation}
    \frac{\partial \loss(b)}{\partial \quaternion{b}^{(l)}}
    = -2\quaternion{e}
\end{equation}
and thus four times the derivative of Theorem \ref{theorem:bias_final_layer}. Likewise, gradients four times as big as the GHR gradients holds for the hidden layers due to Proposition \ref{theo:autograd_activation_previous_layer}.

\newcommand{\autoGradDerivThree}[1]{
    \frac{\partial \loss}{\partial \quatCompR{z}} \frac{\partial \quatCompR{z}}{\partial #1} + 
    \frac{\partial \loss}{\partial \quatCompI{z}} \frac{\partial \quatCompI{z}}{\partial #1} + 
    \frac{\partial \loss}{\partial \quatCompJ{z}} \frac{\partial \quatCompJ{z}}{\partial #1} + 
    \frac{\partial \loss}{\partial \quatCompK{z}} \frac{\partial \quatCompK{z}}{\partial #1}}

\subsection{Implications}

As we have shown, emulating \ac{QNN} utilizing matrix-vector calculations in $\mathbb{R}^4$ and elementwise activation functions as well as using automatic differentiation yields proper parameter updates according to the derived quaternion backpropagation.
This can also be unterstood such that one has to apply the chain rule on the quaternion components instead of the whole quaternion when not using the GHR calculus.
The user just has to be aware of two facts: 
the derivative with respect to the loss in automatic differentiation needs to be real valued and the conjugate of the GHR derivative and it can introduce a factor in the gradients, hence a learning rate correction might be desired, and the gradient flowing backwards through the model, the gradient with respect to the activations of the previous layers, is the conjugate of the GHR derivatives. Hence, one can benefit from the advantages of automatic differentiation as long as the limitations are known. 
Due to Theorem \ref{theorem:ghr-autograd-relation}, arbitrary functions can be introduced in the model as long as the gradient flowing backwards through the model is the conjugate of the GHR gradient.
Especially the usage of non elementwise operating activation functions within \ac{QNN} is of particular interest here and something we aim to investigate in future work. 

%% file: images/autoDiff.tex



%
\tikzset{param/.style={draw, rectangle, fill=drawio_blue_fill, minimum size=.6cm}}%
\tikzset{operator/.style={draw, circle, fill=drawio_yellow_fill, minimum size=.6cm}}%
\begin{tikzpicture}
    \node[operator] (p0) at (5.5, 0)  {$+$};
    \node[operator, below=1.2cm of p0] (p1) {$+$};
    \node[operator, below=1.2cm of p1] (p2) {$+$};
    \node[operator, below=1.2cm of p2] (p3) {$+$};
    \node[operator, left=4cm of p0] (m0) {$\times$};
    \node[operator, left=4cm of p1] (m1) {$\times$};
    \node[operator, left=4cm of p2] (m2) {$\times$};
    \node[operator, left=4cm of p3] (m3) {$\times$};
    \node[param, above left = 0.5cm and 4.0cm of m2] (w0) {$w_0$};
    \node[param, above = 1.1cm of w0] (x1) {$x_1$};
    \node[param, above = 1.1cm of x1] (x0) {$x_0$};
    \node[param, below = 1.1cm of w0] (x2) {$x_2$};
    \node[param, below = 1.1cm of x2] (x3) {$x_3$};
    \node[operator, right=1.4cm of p0] (sig0) {$\sigma$};
    \node[operator, right=1.4cm of p1] (sig1) {$\sigma$};
    \node[operator, right=1.4cm of p2] (sig2) {$\sigma$};
    \node[operator, right=1.4cm of p3] (sig3) {$\sigma$};
    \node[param, right=1.4cm of sig0] (y0) {$y_0$};
    \node[param, right=1.4cm of sig1] (y1) {$y_1$};
    \node[param, right=1.4cm of sig2] (y2) {$y_2$};
    \node[param, right=1.4cm of sig3] (y3) {$y_3$};
    \node[param, above left = 0.0cm and 1.2cm of p0] (b0) {$b_0$};  
    \node[param, above left = 0.0cm and 1.2cm of p1] (b1) {$b_1$};  
    \node[param, above left = 0.0cm and 1.2cm of p2] (b2) {$b_2$};  
    \node[param, above left = 0.0cm and 1.2cm of p3] (b3) {$b_3$};  
    \node[param, draw=none, fill=none, below left = 0.0cm and 1.2cm of p0] (add0) {$\dots$};
    \node[param, draw=none, fill=none, below left = 0.0cm and 1.2cm of p1] (add1) {$\dots$};
    \node[param, draw=none, fill=none, below left = 0.0cm and 1.2cm of p2] (add2) {$\dots$};
    \node[param, draw=none, fill=none, below left = 0.0cm and 1.2cm of p3] (add3) {$\dots$};
    \draw[->] ([yshift=+1pt]p0.east) -- ([yshift=+1pt]sig0.west) node [above=0pt, midway] {$z_0$};%
    \draw[->] ([yshift=+1pt]p1.east) -- ([yshift=+1pt]sig1.west) node [above=0pt, midway] {$z_1$};%
    \draw[->] ([yshift=+1pt]p2.east) -- ([yshift=+1pt]sig2.west) node [above=0pt, midway] {$z_2$};%
    \draw[->] ([yshift=+1pt]p3.east) -- ([yshift=+1pt]sig3.west) node [above=0pt, midway] {$z_3$};%
    \draw[->] ([yshift=-1pt]sig0.west) -- ([yshift=-1pt]p0.east) node [below=0pt, midway] {$\frac{\partial \loss}{\partial y_0}\frac{\partial y_0}{\partial z_0}$};%
    \draw[->] ([yshift=-1pt]sig1.west) -- ([yshift=-1pt]p1.east) node [below=0pt, midway] {$\frac{\partial \loss}{\partial y_1}\frac{\partial y_1}{\partial z_1}$};%
    \draw[->] ([yshift=-1pt]sig2.west) -- ([yshift=-1pt]p2.east) node [below=0pt, midway] {$\frac{\partial \loss}{\partial y_2}\frac{\partial y_2}{\partial z_2}$};%
    \draw[->] ([yshift=-1pt]sig3.west) -- ([yshift=-1pt]p3.east) node [below=0pt, midway] {$\frac{\partial \loss}{\partial y_3}\frac{\partial y_3}{\partial z_3}$};%
    \draw[->] ([yshift=+1pt]sig0.east) -- ([yshift=+1pt]y0.west);%
    \draw[->] ([yshift=+1pt]sig1.east) -- ([yshift=+1pt]y1.west);%
    \draw[->] ([yshift=+1pt]sig2.east) -- ([yshift=+1pt]y2.west);%
    \draw[->] ([yshift=+1pt]sig3.east) -- ([yshift=+1pt]y3.west);%
    \draw[->] ([yshift=-1pt]y0.west) -- ([yshift=-1pt]sig0.east) node [below=0pt, midway] {$\frac{\partial \loss}{\partial y_0}$};%
    \draw[->] ([yshift=-1pt]y1.west) -- ([yshift=-1pt]sig1.east) node [below=0pt, midway] {$\frac{\partial \loss}{\partial y_1}$};%
    \draw[->] ([yshift=-1pt]y2.west) -- ([yshift=-1pt]sig2.east) node [below=0pt, midway] {$\frac{\partial \loss}{\partial y_2}$};%
    \draw[->] ([yshift=-1pt]y3.west) -- ([yshift=-1pt]sig3.east) node [below=0pt, midway] {$\frac{\partial \loss}{\partial y_3}$};%
    \draw[->, transform canvas={yshift=+1pt}] (w0) -- (m0);%
    \draw[->, transform canvas={yshift=+1pt}] (w0) -- (m1);%
    \draw[->, transform canvas={yshift=-1pt}] (w0) -- (m2);%
    \draw[->, transform canvas={yshift=-1pt}] (w0) -- (m3);%
    \draw[<-, transform canvas={yshift=-1pt}] (w0) -- (m0)
        node [below right=-15pt and 10pt, midway] {$\frac{\partial \loss}{\partial w_0x_0}\frac{\partial w_0x_0}{\partial w_0}$};%
    \draw[<-, transform canvas={yshift=-1pt}] (w0) -- (m1)
        node [below right=-7pt and 10pt, midway] {$\frac{\partial \loss}{\partial w_0x_1}\frac{\partial w_0x_1}{\partial w_0}$};%
    \draw[<-, transform canvas={yshift=+1pt}] (w0) -- (m2)
        node [above right=-7pt and 10pt, midway] {$\frac{\partial \loss}{\partial w_0x_2}\frac{\partial w_0x_2}{\partial w_0}$};%
    \draw[<-, transform canvas={yshift=+1pt}] (w0) -- (m3)
        node [above right=-15pt and 10pt, midway] {$\frac{\partial \loss}{\partial w_0x_3}\frac{\partial w_0x_3}{\partial w_0}$};%
    \draw[->] (x0) -- (m0);%
    \draw[->] (x1) -- (m1);%
    \draw[->] (x2) -- (m2);%
    \draw[->] (x3) -- (m3);%
    \draw[->, transform canvas={yshift=+1pt}] (m0) -- (p0);%
    \draw[->, transform canvas={yshift=+1pt}] (m1) -- (p1);%
    \draw[->, transform canvas={yshift=+1pt}] (m2) -- (p2);%
    \draw[->, transform canvas={yshift=+1pt}] (m3) -- (p3);%
    \draw[<-, transform canvas={yshift=-1pt}] (m0) -- (p0)
        node [below left=0pt and 0pt, midway] {$\frac{\partial \loss}{\partial z_0}\frac{\partial z_0}{\partial w_0x_0}$};%
    \draw[<-, transform canvas={yshift=-1pt}] (m1) -- (p1)
        node [below left=0pt and 0pt, midway] {$\frac{\partial \loss}{\partial z_0}\frac{\partial z_0}{\partial w_0x_1}$};%
    \draw[<-, transform canvas={yshift=-1pt}] (m2) -- (p2)
        node [below left=0pt and 0pt, midway] {$\frac{\partial \loss}{\partial z_0}\frac{\partial z_0}{\partial w_0x_2}$};%
    \draw[<-, transform canvas={yshift=-1pt}] (m3) -- (p3)
        node [below left=0pt and 0pt, midway] {$\frac{\partial \loss}{\partial z_0}\frac{\partial z_0}{\partial w_0x_3}$};%
    \draw[->, transform canvas={yshift=-1pt}] (b0) -- (p0);%
    \draw[->, transform canvas={yshift=-1pt}] (b1) -- (p1);%
    \draw[->, transform canvas={yshift=-1pt}] (b2) -- (p2);%
    \draw[->, transform canvas={yshift=-1pt}] (b3) -- (p3);%
    \draw[<-, transform canvas={yshift=1pt}] (b0) -- (p0)
        node [above right=0pt and -15pt, midway] {$\frac{\partial \loss}{\partial z_0}\frac{\partial z_0}{\partial b_0}$};%
    \draw[<-, transform canvas={yshift=1pt}] (b1) -- (p1)
        node [above right=0pt and -15pt, midway] {$\frac{\partial \loss}{\partial z_1}\frac{\partial z_1}{\partial b_1}$};%
    \draw[<-, transform canvas={yshift=1pt}] (b2) -- (p2)
        node [above right=0pt and -15pt, midway] {$\frac{\partial \loss}{\partial z_2}\frac{\partial z_2}{\partial b_2}$};%
    \draw[<-, transform canvas={yshift=1pt}] (b3) -- (p3)
        node [above right=0pt and -15pt, midway] {$\frac{\partial \loss}{\partial z_3}\frac{\partial z_3}{\partial b_3}$};%
    \draw[->] (add0) -- (p0);%
    \draw[->] (add1) -- (p1);%
    \draw[->] (add2) -- (p2);%
    \draw[->] (add3) -- (p3);%
\end{tikzpicture}%

%% file: content/methodology.tex
\section{Quaternionic Time-Series Compression}
\label{sec:methodology}

\begin{figure*}[htb]%
    \centering%
    \input{plots/nn_architecture}%
    \caption{Illustration of the quaternionic time-series compression methodology.}
    \label{fig:model_architecture}
\end{figure*}
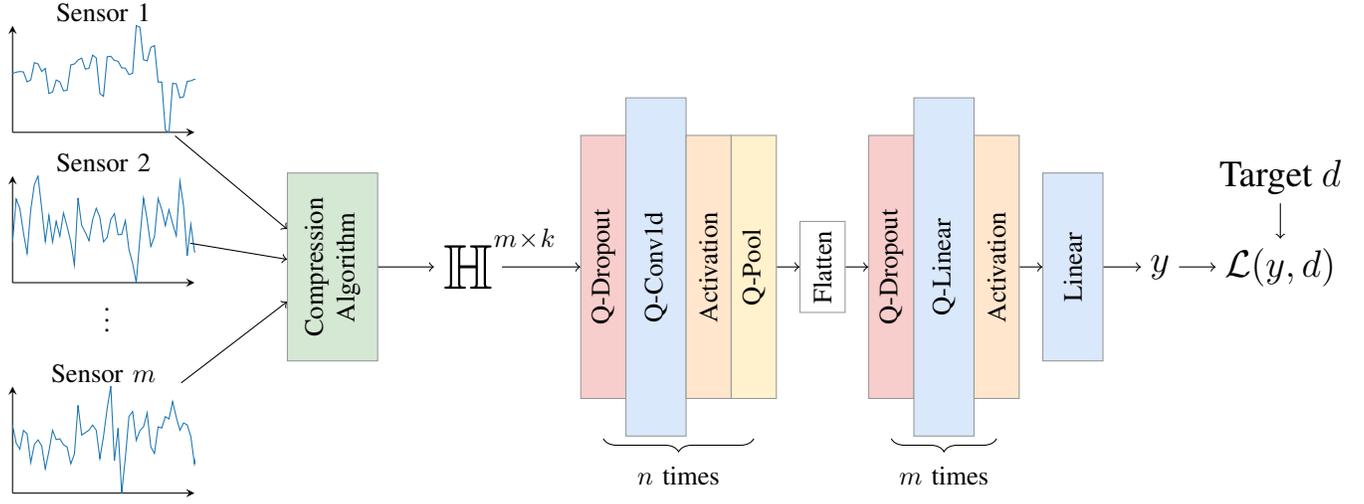

Having introduced the required theoretical fundamentals, namely the quaternion backpropagation required to optimize \ac{QNN} by means of gradient descent, we now proceed to describe the quaternionic time-series compression methodology where an overview is depicted in Figure \ref{fig:model_architecture}.

The sensor readings are processed using the compression algorithm which we describe in more detail in the following subsection. This yields a shorter, quaternion valued time-series which is then fed into the \ac{QNN} architecture

As a feature extractor, we use Dropout - Conv1d - Activation - Pooling blocks, followed by a flatten operation and Dropout - Linear - Activation blocks as a classifier. All of these layers are quaternion valued which will be described in more detail later. The final layer, however, is a conventional linear layer as there is no natural way to express a probability distribution or calculate a cross-entropy in quaternion space. 
Hence, it serves the purpose of a learnable mapping back from $\mathbb{H} \rightarrow \mathbb{R}$.

For comparison with real valued \acp{NN}, we create an equivalent by replacing all quaternion valued layers with real valued ones. As these are lacking the capability to deal with the quaternionic input data, we convert the input data of shape $\mathbb{H}^{m \times k}$ into $\mathbb{R}^{4m \times k}$, i.e. for each quaternion channel of length $k$ we create 4 channels of length $k$, carrying the information about min, max, mean and std. Consequently, that yields the relation of one quaternionic channel to four real valued ones.

\subsection{Compression Algorithm}

As an attempt to shrink time-series data, an intuitive and often used approach is to downsample the data, e.g. by using the mean of the considered data segment. However, while doing so, there is an inherent information loss. For example, the mean does not reflect if there was large deviation in the underlying data or if all values were relatively close to each other. To overcome this issue, multiple characteristics instead of just one can be used in an attempt to lower the information loss and create more representative, compressed data. As these statistics are related to each other, we propose to compose and gather them in a quaternion, specifically in the real and three imaginary parts. Hence, we obtain a mathematical object carrying the compressed information. Naturally, this yields a limit of four statistical values, for extensions one might consider octonion \cite{villa_octonion-valued_2016} or dual-quaternion \cite{poppelbaum_predicting_2022, schwung_rigid_2021} based architectures. We will elaborate more on this in future work.
In this study, we chose the min, max, mean and standard deviation for the descriptive properties as they are very intuitive in terms of representativeness and easy to obtain. Furthermore, this combination has showed good results in \cite{guo_evaluation_2020}.

Assume a multivariate time-series consisting of $m$ channels and $n$ samples. Then, for all channels $m$, we divide the $n$ samples into $k$ parts of equal length $l$. In case $n$ is not divisible by $l$ without remainder, then the last chunk is of a shorter length. Subsequently, for all chunks, we determine the minimum, maximum, mean and standard deviation, which we store in the four components of a quaternion, Hence, we create a mapping from $\mathbb{R}^{m \times n}$ into $\mathbb{H}^{m \times k}$ whereby the compression rate is determined by the ratio $l / 4$. The whole process is further illustrated by Figure \ref{fig:compression_sample}.

\begin{figure}[htb]
    \centering
    \resizebox{\linewidth}{!}{%
        \input{plots/compression_example}
    }
    \caption{Example illustrating the compression algorithm.}
    \label{fig:compression_sample}
\end{figure}

By employing the Hamilton product throughout the \ac{NN} architecture, we ensure that the four features are used in a contiguous manner and emphasize the combined nature as well as their interrelationship within the proposed compression representation

\subsection{Quaternion valued Layer}

In addition to the already described quaternion linear layer, for our model we use the following quaternion valued layers:



\subsubsection{1D Quaternion Convolution Layer}

To extract features from a quaternionic time-series, analog to the real valued equivalent, a one-dimensional convolution in quaternion space is required. Note that these layer can also be seen as a special case of the quaternion linear layer when treating the convolution as a matrix multiplication by a toeplitz matrix.
For a quaternion valued input of size $(B, C_{in}, L_{in})$ where $B$ is the batch size, $C_{in}$ are the input channels and $L_{in}$ is the length of the multivariate input sequence, the $i^{th}$ element of the output channel $j$ using a kernel size $K$ is calculated applying

\begin{equation}
    \quaternion{y}_{j, i} = \quaternion{b}_j + \sum_{c=0}^{C_{in} - 1}\sum_{k=0}^{K - 1} \quaternion{x}_{c, i + k} \otimes \quaternion{w}_{j, c, k}
\end{equation}

whereby the overall shape of the output is $(B, C_{out}, L_{out})$. Here, the weights are of size $\mathbb{H}^{C_{out}, C_{in}, K}$ and the bias is of size $\mathbb{H}^{C_{out}}$.
For ease of notation, we omit the dilation and the stride.

\subsubsection{Activations}
We employ element-wise working activation functions as proposed in 
\cite{zhu_quaternion_2018, parcollet_quaternion_2016, parcollet_speech_2018}. Assuming an already known and tested activation $\psi(\cdot)$ like ReLU or Tanh, the application on a quaternion input is

\begin{equation}
    \psi(\quaternion{q}) = \psi(\quatCompR{q}) + \psi(\quatCompI{q})\imagI + \psi(\quatCompJ{q})\imagJ + \psi(\quatCompK{q})\imagK.
\end{equation}

The design and usage of activation functions specifically designed for the quaternion space provides an interesting direction for future research.

\subsubsection{1D Quaternion Max-Pooling}

Maximum-Pooling in quaternion space directly raises the question "What actually is the maximum of a set of multiple quaternions?" which has no distinct answer. This is due to the fact that it is composed out of four elements which can have their individual maximums, however they still have a combined meaning. In \cite{yin_quaternion_2019}, an approach is proposed based on the amplitude or magnitude of a quaternion: it is used to construct a guidance matrix which selects the pooled quaternions based on the maximum magnitude out of a quaternion valued matrix. Contrary, \cite{zhu_quaternion_2018} opts to consider and pool the real and three imaginary parts separately. In the following, we will describe both approaches in detail, as they tackle the problem from different views: the 1D Component Max-Pooling puts more emphasis on the individual maximums whereas the 1D Magnitude Max-Pooling emphasizes the combined relation of a quaternion.

\paragraph{1D Quaternion Component Max-Pooling}
This approach considers the real part $\quatCompR{q}$ and the imaginary parts $\quatCompI{q}$, $\quatCompJ{q}$ and $\quatCompK{q}$ separately. Specifically, using a kernel size $K$, the output is composed of the maximum of the $K$ individual real parts and the respective $K$ individual imaginary parts $\imagI$, $\imagJ$ and $\imagK$. This is described using
\begin{equation}
\begin{aligned}
    &out(B_i, C_j, k) = \\
    &\max_{m=0, \dots, kernel\_size - 1} \mathfrak{R}\left( in(N_i, C_j, stride \times k + m)\right) \\
    + &\max_{m=0, \dots, kernel\_size - 1} \mathfrak{I}( in(N_i, C_j, stride \times k + m)) \imagI \\
    + &\max_{m=0, \dots, kernel\_size - 1} \mathfrak{J}( in(N_i, C_j, stride \times k + m)) \imagJ\\
    + &\max_{m=0, \dots, kernel\_size - 1} \mathfrak{K}( in(N_i, C_j, stride \times k + m)) \imagK . \\
\end{aligned}
\end{equation}

\paragraph{1D Quaternion Magnitude Max-Pooling}
Contrary to the component-wise approach, here we consider the quaternion as a whole and use its norm $\lVert q \rVert$ as a comparison measurement. Then, out of $K$ considered quaternions, the one with the greatest norm is selected as the pooling output.
We can formulate this process as 
\begin{equation}
\begin{aligned}
    &out(B_i, C_j, k) = in(N_i, C_j, l)~\text{where} \\
    l=&\operatorname*{argmax}_{m=0, \dots, kernel\_size - 1} in(N_i, C_j, \lVert stride \times k + m \rVert).
\end{aligned}
\end{equation}
As the norm operation is not affected by the sign of the quaternion components, using this approach it is more likely that negative values remain after pooling in comparison to the component max-pooling approach.

%% file: plots/nn_architecture.tex
%
%
%
%
%
%
%
%
\definecolor{matplotlib_blue}{HTML}{1f77b4}%
\definecolor{matplotlib_orange}{HTML}{ff7f0e}%
\definecolor{matplotlib_green}{HTML}{2ca02c}%
\definecolor{drawio_yellow_fill}{HTML}{FFF2CC}%
\definecolor{drawio_orange_fill}{HTML}{FFE6CC}%
\definecolor{drawio_red_fill}{HTML}{F8CECC}%
\definecolor{drawio_blue_fill}{HTML}{DAE8FC}%
\definecolor{drawio_green_fill}{HTML}{D5E8D4}%
\definecolor{drawio_yellow_border}{HTML}{D6B656}%
\tikzset{conv/.style={black,draw=gray!80,rectangle,minimum height=0.8cm, minimum width=4.5cm, fill=drawio_blue_fill}}%
\tikzset{lin/.style={black,draw=gray!80,rectangle,minimum height=0.8cm, minimum width=4.5cm, fill=drawio_blue_fill}}%
\tikzset{lin2/.style={black,draw=gray!80,rectangle,minimum height=0.8cm, minimum width=2.5cm, fill=drawio_blue_fill}}%
\tikzset{drop/.style={black,draw=gray!80,rectangle,minimum height=0.6cm, minimum width=3.5cm, fill=drawio_red_fill}}%
\tikzset{act/.style={black,draw=gray!80,rectangle,minimum height=0.6cm, minimum width=3.5cm, fill=drawio_orange_fill}}%
\tikzset{pool/.style={black,draw=gray!80,rectangle,minimum height=0.6cm, minimum width=3.5cm, fill=drawio_yellow_fill}}%
\tikzset{view/.style={black,draw=gray!80,rectangle,minimum height=0.6cm}}%
\tikzset{sensor/.style={black,rectangle,minimum height=1.5cm, minimum width=2.2cm}}
\tikzset{comp/.style={black,draw=gray!80,rectangle,minimum height=1.2cm, minimum width=2.5cm,fill=drawio_green_fill}}%
\newcommand{\sensorPlot}[2]{%
    \begin{axis}[%
            anchor=center,
            xshift=-0.66cm,
            width=4cm,
            height=3cm,
            ticks=none,
            axis x line=bottom,
            axis y line=left,
            title=#1,
            title style={yshift=-5ex,},
            xmax=50,
            no markers,]%
            \addplot[color=matplotlib_blue] table[col sep=comma, y=y, x expr=\coordindex] {#2};%
    \end{axis}%
}%
\begin{tikzpicture}
    \node[sensor] (s1) at (-5.8, 2.5) {
        \sensorPlot{Sensor 1}{plots/ts_data_1.txt}
    };
    \node[sensor] (s2) at (-5.8, 0.5) {    
        \sensorPlot{Sensor 2}{plots/ts_data_2.txt}
    };
    \node              at (-5.8, -0.6) {$\vdots$};%
    \node[sensor] (sn) at (-5.8, -2.3) {
        \sensorPlot{Sensor $m$}{plots/ts_data_3.txt}
    };
    \node[comp, rotate=90,align=center] (comp) at (-2.8, 0) {Compression \\ Algorithm};%
    \node (x) at (-1,0) {\Huge $\mathbb{H}$~};%
    \node at (-0.25, 0.38) {\normalsize $m \! \times \! k$};%
    \node[drop,rotate=90] (drop1) at (0.8,0) {Q-Dropout};%
    \node[conv,rotate=90, below = 0pt of drop1, yshift=0.4pt] (conv1) {Q-Conv1d};%
    \node[act,rotate=90, below = 0pt of conv1, yshift=0.4pt] (act1) {Activation};%
    \node[pool,rotate=90, below = 0pt of act1, yshift=0.4pt] (pool1) {Q-Pool};%

    \node[view,rotate=90, below = 0pt of pool1, yshift=-.3cm] (reshape) {Flatten};%

    \node[drop,rotate=90, below = 0pt of reshape, yshift=-.3cm] (drop2) {Q-Dropout};%
    \node[lin,rotate=90, below = 0pt of drop2, yshift=0.4pt] (fc1) {Q-Linear};%
    \node[act,rotate=90, below= 0pt of fc1, yshift=0.4pt] (act2) {Activation};%

    \node[lin2,rotate=90, below = 0pt of act2, yshift=-.3cm] (fc2) {Linear};%
    
    \node (y) at (8.2, 0) {\Large $y$};%
    \node (loss) at (9.8, 0) {\Large $\loss (y, d)$};%
    \node (d) at (9.8, 1.2) {\Large Target $d$};%

    \draw[->] (s1) -- (comp);%
    \draw[->] (s2) -- (comp);%
    \draw[->] (sn) -- (comp);%

    \draw[->] (comp) -- (x);%

    \draw[->] (x) -- (drop1);%
    \draw[->] (pool1) -- (reshape);%
    \draw[->] (reshape) -- (drop2);%
    \draw[->] (act2) -- (fc2);%
    \draw[->] (fc2) -- (y);%

    \draw[->] (y) -- (loss);%
    \draw[->] (d) -- (loss);%

    \draw[decorate, decoration={brace, amplitude=5pt, raise=15pt, mirror}] (drop1.west) -- (pool1.west) node [below=22pt, midway] {$n$ times};%
    \draw[decorate, decoration={brace, amplitude=5pt, raise=15pt, mirror}] (drop2.west) -- (act2.west) node [below=22pt, midway] {$m$ times};%
\end{tikzpicture}%
%

%% file: plots/compression_example.tex
\definecolor{matplotlib_blue}{HTML}{1f77b4}
\noindent
\begin{NiceTabular}{l c c c c c}[hvlines,rules/color=matplotlib_blue!15]
\CodeBefore
\rowcolor{matplotlib_blue!15}{3,5}
\rowcolor{matplotlib_blue!60}{1}
\columncolor{matplotlib_blue!60}{1}
\Body
\Block{1-2}{\quad} & & \Block[v-center]{1-1}{\textbf{Chunk $\mathbf{1}$}} & \Block[v-center]{1-1}{\textbf{Chunk $\mathbf{2}$}} & \Block[v-center]{1-1}{$\cdots$} & \Block[v-center]{1-1}{\textbf{Chunk $\mathbf{k}$}} \\[8pt]
\Block{2-1}<\rotate>{\textbf{Sensor 1}} 
& \Block[v-center]{1-1}{$\mathbb{R}$} & \Block[v-center]{1-1}{3 5 -1 7 4 9} & \Block[v-center]{1-1}{3 1 -2 -2 0 5} & \Block[v-center]{1-1}{$\cdots$} & \Block[v-center]{1-1}{7 3 2 1 2 6} \\[6pt]
& $\mathbb{H}$ & \Block{}{$-1.00~$\\$+9.00i$\\$+4.50j$\\$+3.45k$} 
               & \Block{}{$-2.00~$\\$+5.00i$\\$+0.83j$\\$+2.79k$}
               & $\cdots$ 
               & \Block{}{$1.00~$\\$+7.00i$\\$+3.67j$\\$+2.42k$} \\
\Block{2-1}<\rotate>{\textbf{Sensor 2}} 
& \Block[v-center]{1-1}{$\mathbb{R}$} & \Block[v-center]{1-1}{0 -1 2 4 3 6} & \Block[v-center]{1-1}{1 0 -4 3 1 -1} & \Block[v-center]{1-1}{$\cdots$} & \Block[v-center]{1-1}{0 -1 -3 -5 -4 -1} \\[6pt]
& $\mathbb{H}$ & \Block{}{$-1.00~$\\$+6.00i$\\$+2.33j$\\$+2.58k$}
               & \Block{}{$-4.00~$\\$+3.00i$\\$+0.00j$\\$+2.37k$}
               & $\cdots$ 
               & \Block{}{$-5.00~$\\$+0.00i$\\$-2.33j$\\$+1.97k$}\\
\end{NiceTabular}

%% file: content/experiments_ts_compression.tex
\section{Experiments}
\label{sec:experiments_ts}

This section provides the experimental evaluation of the proposed methodology. We start with a regular supervised training including a test of the robustness against random parameter initialization. This is intended to prove the effectiveness of the proposed compression and to provide detailed comparison of quaternion valued with real valued architectures. This is followed by using the proposed approach in a self-supervised learning setup to highlight state-of-the-art performance. Finally, a detailed comparison with the literature is provided.

\subsection{Experimental setup}

For our experimental evaluation, we use the \ac{TE} dataset which consists of 21 fault cases and a regular operation case. In total, 52 variables are considered. For the training split, 480 samples are collected for the fault cases and 500 for the regular operation case. For the testing split, 960 samples are collected, whereby the fault was introduced after eight hours or 160 samples. We just consider the last 800 samples after the introduction of the fault in this work.
Furthermore, we apply a sliding window of length 320 on the data. 
For our compression algorithm, we chose 40 chunks of length eight, yielding a compression-rate of factor two. Hence, we obtain data in the shape $(N, 52, 4, 40)$ and $(N, 208, 40)$.

To incorporate different architectural choices, we use a combination of one convolution block with three linear blocks, two convolution blocks with four linear blocks and three convolution blocks with four linear blocks from which we design a narrow variant with a low amount of trainable parameters and a wide variant with a high amount of trainable parameters each. For the quaternion models, we use both presented pooling methods: the component pooling and the magnitude pooling to compare them in different learning setups. For the real valued comparison models, we also opt for two versions: The first is a model with an equal amount of trainable parameters, the second is a version which creates the same number of features as the layer outputs. However, this comes with the effect of having approximately four times the trainable parameters in the non-quaternion variant in comparison to the quaternion version.
Furthermore, we add two baseline models to showcase the effect of the proposed compression: one for the uncompressed input and one using just a simple mean-resampling. Both utilize the same config as the real model with equal parameters, however due to the different input shape, the first convolution layer and consequently the first linear layer after the flatten operation is different. 
Hence, we end up with a total of 36 different model architectures to compare for our experiments.
A detailed overview of the model configurations can be found 
in Table \ref{tab:architectures_te}.

\begin{table*}[htb]
    \centering
    \caption{Used Neural Network Configurations}
    \label{tab:architectures_te}
    \begin{tabular}{lllllllllll}
        \toprule
        ~ & ~ & ~ & params & \multicolumn{3}{l}{conv output channel} & \multicolumn{4}{l}{linear output sizes} \\ 
        \midrule
        \multirow{6}{*}{1c3l} & \multirow{3}{*}{low}  
            & quat                     & 327030  & 32   & ~    & ~ & 128  & 8    & 22 & ~ \\
        ~&~ & real equal params        & 327535  & 25x4 & ~    & ~ & 133  & 30   & 22 & ~ \\ 
        ~&~ & real equal feature space & 1303926 & 32x4 & ~    & ~ & 512  & 32   & 22 & ~ \\ 
        \cmidrule{3-10}
        ~                     & \multirow{3}{*}{high} 
            & quat                     & 1859702 & 96   & ~    & ~ & 256  & 8    & 22 & ~ \\ 
        ~&~ & real equal params        & 1859146 & 85$\times$4 & ~    & ~ & 256  & 32   & 22 & ~ \\ 
        ~&~ & real equal feature space & 7432310 & 96$\times$4 & ~    & ~ & 1024 & 32   & 22 & ~ \\ 
        \midrule
        \multirow{6}{*}{2c4l} & \multirow{3}{*}{low}  
            & quat                     & 229366  & 32   & 32   & ~ & 128  & 128  & 8  & 22 \\ 
        ~&~ & real equal params        & 229342  & 25$\times$4 & 25$\times$4 & ~ & 128  & 96   & 32 & 22 \\ 
        ~&~ & real equal feature space & 911350  & 32$\times$4 & 32$\times$4 & ~ & 512  & 512  & 32 & 22 \\ 
        \cmidrule{3-11}
        ~                     & \multirow{3}{*}{high} 
            & quat                     & 1189366 & 96   & 96   & ~ & 256  & 256  & 8  & 22 \\ 
        ~&~ & real equal params        & 1189749 & 73$\times$4 & 73$\times$4 & ~ & 259  & 256  & 32 & 22 \\ 
        ~&~ & real equal feature space & 4746742 & 96$\times$4 & 96$\times$4 & ~ & 1024 & 1024 & 32 & 22 \\ 
        \midrule
        \multirow{6}{*}{3c4l} & \multirow{3}{*}{low}
            & quat                     & 163958  & 32          & 32          & 32          & 128  & 128  & 8  & 22 \\ 
        ~&~ & real equal params        & 164998  & 23$\times$4 & 22$\times$4 & 22$\times$4 & 96   & 60   & 24 & 22 \\
        ~&~ & real equal feature space & 649334  & 32$\times$4 & 32$\times$4 & 32$\times$4 & 512  & 512  & 32 & 22 \\ 
        \cmidrule{3-11}
        ~                     & \multirow{3}{*}{high} 
            & quat                     & 845686  & 96          & 96          & 96          & 256  & 256  & 8  & 22 \\ 
        ~&~ & real equal params        & 853118  & 66$\times$4 & 66$\times$4 & 66$\times$4 & 124  & 60   & 24 & 22 \\ 
        ~&~ & real equal feature space & 3370870 & 96$\times$4 & 96$\times$4 & 96$\times$4 & 1024 & 1024 & 32 & 22 \\  
        \bottomrule
    \end{tabular}
\end{table*}

For each model, we run a tuning with 50 trials from which 20 are warmup trials using the Python package Optuna \cite{akiba_optuna_2019} to obtain an optimized set of hyperparameters. Specifically, we tune for the following:
\begin{itemize}
    \item Activation: ReLU, Tanh, Tanhshrink
    \item Learning Rate: $1\times 10^{-6}$ - $1\times 10^{-1}$ 
    \item Batch-size: $2^4$ - $2^8$
    \item Dropout ratio: 0, 0.1, 0.2, 0.3, 0.4
\end{itemize}

\subsection{Supervised Learning}

In the fully supervised learning experiments, we trained 36 models in total. All of them were trained using the cross-entropy loss for 50 epochs each. This yielded the results provided in Table \ref{tab:results_te}.

\begin{table}[htb]
\centering
\caption{Accuracies in \% obtained using the supervised learning setup. Maximum values are highlighted bold.}
\label{tab:results_te}
\begin{tabular}{l@{\hskip 6pt}lcccccc}
    \toprule
    ~ & ~ &  \multicolumn{2}{c}{quat} & \multicolumn{2}{c}{real} & \multicolumn{2}{c}{baseline} \\
    \cmidrule(lr){3-4} \cmidrule(lr){5-6} \cmidrule(lr){7-8}
    ~ & ~ & \makecell{comp. \\ pooling} & \makecell{mag.\\pooling} & \makecell{equal \\ params} & \makecell{equal \\ features} & \makecell{uncom- \\ pressed} & mean \\
    \midrule 
    \multirow{2}{*}{\rotatebox[origin=c]{90}{1c3l}} & low  & 68.97 & \textbf{70.82} & 69.85 & 66.57 & 66.71 & 55.46 \\
                          & high & 68.58 & \textbf{71.59} & 66.95 & 68.59 & 70.07 & 52.39 \\
    \multirow{2}{*}{\rotatebox[origin=c]{90}{2c4l}} & low  & \textbf{71.16} & 70.41 & 69.54 & 68.69 & 61.77 & 59.71 \\
                          & high & \textbf{71.41} & 68.90 & 67.08 & 69.72 & 60.94 & 52.66 \\
    \multirow{2}{*}{\rotatebox[origin=c]{90}{3c4l}} & low  & \textbf{69.87} & 69.03 & 67.42 & 66.96 & 57.82 & 56.49 \\
                          & high & 69.70 & \textbf{69.98} & 66.90 & 68.99 & 53.13 & 56.82 \\
    \bottomrule
\end{tabular}
\end{table}

As we can see, in all cases, the quaternion model outperformed its real valued counterparts. This is the case even when being in a disadvantage parameter-wise when comparing with the real eq. features model configuration. Thus, we can conclude that the quaternionic compression in combination with the quaternion valued model architecture outperforms the real valued counterparts being fed the same data in these experiments.  In terms of quaternion pooling, no tendency is observable at this point as both versions achieved the respective highest accuracy three times. With the one exception of the 1c3l high configuration for the uncompressed baseline, in all baseline runs a significant drop in performance is existent. This highlights the advantages of compressing a long time-series for a fault classification task as proposed.

To investigate the effect of random initialization on the respective architectures, for each model we take the best set of hyperparameter, obtained from the experiments above, and train it again for 50 times using this specific set of hyperparameter. This yields the mean accuracies reported in Table \ref{tab:results_random_init_mean} and maximum accuracies reported in Table \ref{tab:results_random_init_max}. Furthermore, the accuracy distribution is shown in Figure \ref{fig:random_init_boxplot}.
\input{plots/random_init_eval_plots}

\begin{table*}[htb]
\centering
\caption{Mean Accuracies in \% including the $95\%$ confidence interval obtained in the random initialization test. Maximum values are highlighted bold.}
\label{tab:results_random_init_mean}
\begin{tabular}{llcccccc} 
    \toprule
    ~ & ~ &  \multicolumn{2}{c}{quat} & \multicolumn{2}{c}{real} & \multicolumn{2}{c}{baseline} \\
    \cmidrule(lr){3-4} \cmidrule(lr){5-6} \cmidrule(lr){7-8}
    ~ & ~ & \makecell{comp. \\ pooling} & \makecell{mag.\\pooling} & \makecell{equal \\ params} & \makecell{equal \\ features} & \makecell{uncom-\\pressed} & mean \\
    \midrule 
    \multirow{2}{*}{\rotatebox[origin=c]{90}{1c3l}} 
        & low  & 66.57 $\pm$ 3.09 & 67.20 $\pm$ 3.13 & \textbf{67.92} $\pm$ 3.15 & 64.16 $\pm$ 3.13 & 65.69 $\pm$ 3.29 & 53.28 $\pm$ 3.23 \\
        & high & 66.16 $\pm$ 3.19 & \textbf{67.91} $\pm$ 2.42 & 65.00 $\pm$ 3.23 & 64.79 $\pm$ 3.43 & 60.31 $\pm$ 5.94 & 46.85 $\pm$ 3.98 \\
    \multirow{2}{*}{\rotatebox[origin=c]{90}{2c4l}} 
        & low  & 66.61 $\pm$ 3.48 & \textbf{67.03} $\pm$ 4.39 & 63.90 $\pm$ 6.15 & 63.27 $\pm$ 5.93 & 56.81 $\pm$ 4.80 & 54.28 $\pm$ 4.28 \\
        & high & \textbf{68.34} $\pm$ 2.35 & 66.88 $\pm$ 4.14 & 65.53 $\pm$ 4.31 & 64.44 $\pm$ 4.18 & 57.97 $\pm$ 6.08 & 45.15 $\pm$ 5.29 \\
    \multirow{2}{*}{\rotatebox[origin=c]{90}{3c4l}} 
        & low  & 66.14 $\pm$ 5.19 & \textbf{66.26} $\pm$ 4.11 & 61.87 $\pm$ 7.26 & 63.51 $\pm$ 5.09 & 51.35 $\pm$ 6.43 & 52.37 $\pm$ 6.50 \\
        & high & 65.98 $\pm$ 3.46 & \textbf{66.72} $\pm$ 4.60 & 63.96 $\pm$ 4.84 & 62.51 $\pm$ 24.25 & 52.45 $\pm$ 5.96 & 47.43 $\pm$ 4.95 \\
    \bottomrule
\end{tabular}
\end{table*}

\begin{table}[htb]
\centering
\caption{Max Accuracies in \% obtained in the random initialization test. Maximum values are highlighted bold.}
\label{tab:results_random_init_max}
\begin{tabular}{l@{\hskip 6pt}lcccccc}
    \toprule
    ~ & ~ &  \multicolumn{2}{c}{quat} & \multicolumn{2}{c}{real} & \multicolumn{2}{c}{baseline} \\
    \cmidrule(lr){3-4} \cmidrule(lr){5-6} \cmidrule(lr){7-8}
    ~ & ~ & \makecell{comp. \\ pooling} & \makecell{mag.\\pooling} & \makecell{equal \\ params} & \makecell{equal \\ features} & \makecell{uncom-\\pressed} & mean \\
    \midrule 
    \multirow{2}{*}{\rotatebox[origin=c]{90}{1c3l}} 
        & low  & 69.93 & \textbf{71.77} &	70.92 &	68.21 &	69.13 &	56.28 \\
        & high & 69.80 & \textbf{70.21} &	68.48 &	68.12 &	67.85 &	51.41 \\
    \multirow{2}{*}{\rotatebox[origin=c]{90}{2c4l}} 
        & low  & 71.18 & \textbf{71.91} &	71.30 &	69.19 &	62.31 &	57.88 \\
        & high & 71.23 & \textbf{72.79} &	70.55 &	68.27 &	63.17 &	49.52 \\
    \multirow{2}{*}{\rotatebox[origin=c]{90}{3c4l}} 
        & low  & \textbf{72.02} & 71.03 &	67.89 &	67.50 &	58.26 &	61.06 \\
        & high & 69.41 & \textbf{70.73} &	69.85 &	68.92 &	58.08 &	52.69 \\
    \bottomrule
\end{tabular}
\end{table}

It can be observed that one of the quaternion variants achieves the highest overall accuracies over all layer and parameter configurations, whereby the results are five to one in favor of the quaternion magnitude pooling. Similar outcomes can be identified in terms of the mean performance as four times the quaternion magnitude pooling performed best, one time the quaternion component pooling and one time the real valued model with equal parameter count. Except for the 1c3l low configuration, both quaternion architectures obtained a higher mean and median accuracy than their real valued counterparts. Additionally, the quaternion variants show the tendency to have smaller variation and confidence intervals. Hence, especially with a limited budget on training runs, they are more likely to produce higher accuracies.

This overall highlights the advantage of using the quaternion valued models in this scenario. Furthermore, the magnitude pooling seems to be favorable in this fully supervised training setup, which we explain with its more regularizing behavior in comparison to the component pooling.

\subsection{Self-Supervised Learning}

To further test state-of-the-art performance, we use the contrastive learning approach SimCLR-TS proposed by \cite{poppelbaum_contrastive_2022} which established a new baseline on the \ac{TE} dataset.
To prove the superiority of quaternion valued architectures in the proposed application, we replicate the contrastive learning setup and transfer it to our use case. For detailed information on SimCLR-TS, we refer to the original work.

We change the models in such a way that we only use the convolutional blocks with a flatten operation at the end for the contrastive learning. The following linear evaluation is done with a single linear layer, similar to SimCLR-TS.
The tuning setup remains the same, however we add a second learning rate as SimCLR-TS is a two-staged methodology where a shared learning rate does not necessary make sense. The contrastive learning part is again trained for 50 epochs, for the following linear evaluation however it is sufficient to use 20 training epochs. Furthermore, in this experiment we use the two model configurations which obtained the highest accuracies in the supervised learning setup, namely 2c4l high and 3c4l low, with which we could obtain the results displayed in Table \ref{tab:results_contrastive}. 

\begin{table}[htb]
    \centering
    \setlength\tabcolsep{5pt} 
    \caption{Accuracies in \% obtained using the self-supervised learning setup. Maximum values are highlighted bold.}
    \label{tab:results_contrastive}
    \begin{tabular}{lcccccc}
        \toprule
        ~ &  \multicolumn{2}{c}{quat} & \multicolumn{2}{c}{real} & \multicolumn{2}{c}{baseline} \\
        \cmidrule(lr){2-3} \cmidrule(lr){4-5} \cmidrule(lr){6-7}
        ~ & \makecell{comp. \\ pooling} & \makecell{mag.\\pooling} & \makecell{equal \\ params} & \makecell{equal \\ features} & \makecell{uncom-\\pressed} & mean \\
        \midrule 
        2c4l high & \textbf{83.90} & 78.26 & 75.69 & 78.79 & 75.92 & 73.47 \\
        3c4l low  & \textbf{78.38} & 76.52 & 76.25 & 77.51 & 72.57 & 65.68 \\
        
        \bottomrule
    \end{tabular}
\end{table}

As we can see, again the quaternion valued models outperform the real valued counterpart. The real-valued models benefit from the higher number of trainable parameters, but still remain inferior. Hence, also in a more advanced, state-of-the-art training setup, the usage of the Hamilton product within the convolutions allows for performance benefits and enables models with fewer trainable parameters. In these experiments, the component pooling outperformed the magnitude pooling.  We attribute that to the fact that most of the training work is done in the self-supervised part, where the more regularizing effect of the magnitude pooling is not as beneficial as in a fully supervised training setup. Instead, the model performance benefits from the higher degree of flexibility offered by the component pooling.

\subsection{Literature comparison}

The \ac{TE} dataset is widely used in machine learning research, however it also bears some problems in terms of comparability: Not always all of the 21 fault cases are used, and also sometimes not all sensor channels are used, hurting the comparability. Further, also varying windowing approaches are used to create the samples for training and testing. Finally, different evaluation metrics exist. Nevertheless, we want to compare our results with other works to put our proposed approach into context. Therefore, we provide a detailed set of results, obtained using a variety of approaches, in Table \ref{tab:acc_comparison}.

The first work using all fault cases we want to compare with is \cite{jing_fault_2014} where Support Vector Machines (SVM) and Principal Component Analysis (PCA) is used for fault classification on the \ac{TE} dataset.
Another relevant work that uses all error cases is \cite{poppelbaum_contrastive_2022} from which we adapted the semi-supervised learning.
All of them, we could outperform with our best accuracy of $83.9\%$ obtained using the 2c4l high configuration. Compared to SimCLR-TS, the biggest performance gains could be achieved on the difficult cases 3, 9 and 15 at the cost of performance losses in cases 0, 13 and 18.

Further, \cite{wang_deep_2020} employed stacked supervised auto-encoder, however they left out the most difficult cases 3, 9 15 which is a practice not untypical on TE.
Since SimCLR-TS also reported a result for that setup, we likewise performed a training on this reduced cases setup. 
This yielded an accuracy of $93.58\%$, which is on par with SSAE. In almost all fault cases improvements over SimCLR-TS could be achieved, the only two exceptions are the regular operation case 0 with a slight drop of about $0.03\%$, the fault case 13 with a drop of $45,51\%$ and case 16 with a drop of $0,61\%$, preventing an even bigger improvement.

\begin{table*}
    \centering
    \caption{Comparison of the achieved accuracies using contrastive learning and the proposed compression method with other approaches}
    \setlength\tabcolsep{4.6pt} 
    \begin{threeparttable}
    \begin{tabular}{l c c c c c c c c c c c c} %
        \toprule 
        & \multicolumn{4}{c}{All cases}  & \multicolumn{3}{c}{Without 3, 9, 15} & \multicolumn{5}{c}{other} \\
        \cmidrule(lr){2-5}  \cmidrule(lr){6-8} \cmidrule(lr){9-13} 
        \textbf{Case}  & 
        \makecell{\textbf{SVM} \\ \textbf{\cite{jing_fault_2014}}} & 
        \makecell{\textbf{PCA} \\ \textbf{\cite{jing_fault_2014}}} & 
        \makecell{\textbf{SimCLR-} \\ \textbf{TS \cite{poppelbaum_contrastive_2022}}} & 
        \makecell{\textbf{2c4l} \\ \textbf{high}}  &
        \makecell{\textbf{SSAE} \\ \textbf{\cite{wang_deep_2020}}} & 
        \makecell{\textbf{SimCLR-} \\ \textbf{TS \cite{poppelbaum_contrastive_2022}}} & 
        \makecell{\textbf{2c4l} \\ \textbf{high}} &
        \makecell{\textbf{MCNN-} \\ \textbf{LSTM \cite{yuan_multiscale_2019}}}  & 
        \makecell{\textbf{Enhanced}\\ \textbf{RF \cite{chai_enhanced_2020}}} & 
        \makecell{\textbf{MC1-} \\ \textbf{DCNN \cite{yu_multichannel_2021}}} & 
        \makecell{\textbf{UN-DBN} \\ \textbf{\cite{yu_whole_2020}} \tnote{\ddag}} & 
        \makecell{\textbf{CNN} \\ \textbf{\cite{singh_chadha_time_2019}} \tnote{\dag}} \\ 
        \midrule
        Case 0  & 19.27 & 19.69 & \textbf{55.12} & 37.42           & 96.48          & \textbf{98.37} & 98.34             & 93.5  & -   &  -    & -     & -      \\
        Case 1  & 88.44 & 88.54 & 96.29 & \textbf{100.0}           & \textbf{100.0} & 99.43          & \textbf{100.0}    & 99.9  & 99  & 100.0 & 1     & 0.9139  \\
        Case 2  & 86.04 & 89.06 & \textbf{100.0} & \textbf{100.0}  & \textbf{100.0} & 99.86          & \textbf{100.0}    & 99.3  & 98  & 100.0 & 0.99  & 0.8796  \\
        Case 3  & 15.73 & 21.25 & 25.57 & \textbf{73.18}           & -              & -              & -                 & -     & 35  & 83.48 & 0.08  & 0.5059  \\
        Case 4  & 58.02 & 81.35 & 96.57 & \textbf{100.0}           & \textbf{100.0} & 87.71          & \textbf{100.0}    & 100   & 97  & 99.22 & 1     & 0.9973  \\
        Case 5  & 64.79 & 87.81 & 83.86 & \textbf{100.0}           & \textbf{100.0} & 79.71          & \textbf{100.0}    & -     & 100 & 90.40 & 1     & 0.9035  \\
        Case 6  & 71.88 & 89.58 & 99.14 & \textbf{100.0}           & \textbf{100.0} & 98.86          & \textbf{100.0}    & -     & 100 & 93.10 & 1     & 0.9150  \\
        Case 7  & 88.13 & 88.75 & \textbf{100.0} & \textbf{100.0}  & \textbf{100.0} & \textbf{100.0} & \textbf{100.0}    & -     & 100 & 100.0 & 1     & 0.9155  \\
        Case 8  & 45.10 & 83.96 & 86.71 & \textbf{96.47}           & 96.32          & 95.29          & \textbf{100.0}    & 98.8  & 76  & 99.27 & 0.98  & 0.8295  \\
        Case 9  & 12.92 & 22.60 & 29.43 & \textbf{53.85}           & -              & -              & -                 & -     & 23  & 74.60 & 0.02  & 0.4953  \\
        Case 10 & 27.08 & 76.98 & 91.71 & \textbf{100.0}           & 62.16          & 97.14          & \textbf{98.54}    & 99.9  & 81  & 93.75 & 0.84  & 0.7005  \\
        Case 11 & 14.90 & 70.21 & 99.29 & \textbf{100.0}           & 95.93          & 98.57          & \textbf{100.0}    & -     & 76  & 95.45 & 0.91  & 0.6016  \\
        Case 12 & 52.40 & 87.08 & 95.14 & \textbf{100.0}           & 99.87          & \textbf{100.0} & \textbf{100.0}    & 97.0  & 89  & 100.0 & 1     & 0.8666  \\
        Case 13 & 35.10 & \textbf{69.58} & 57.86 & 32.02           & \textbf{89.88} & 71.29          & 25.78             & 98.2  & 30  & 99.14 & 0.95  & 0.4692  \\
        Case 14 & 62.19 & 88.23 & 99.57 & \textbf{100.0}           & \textbf{100.0} & \textbf{100.0} & \textbf{100.0}    & 100   & 100 & 100.0 & 1     & 0.8868  \\
        Case 15 & 22.19 & \textbf{26.98} & 02.71 & 19.13           & -              & -              & -                 & -     & 28  & 73.50 & 0.16  & 0.4354  \\
        Case 16 & 16.77 & 73.65 & 98.57 & \textbf{98.96}           & 54.66          & \textbf{99.57} & 98.96             & -     & -   & 100.0 & 0.68  & 0.6684  \\
        Case 17 & 53.65 & 75.94 & \textbf{100.0} & \textbf{100.0}  & \textbf{100.0} & 99.86          & \textbf{100.0}    & 100   & -   & 100.0 & 0.98  & 0.7711  \\
        Case 18 & 30.94 & 73.54 & \textbf{99.00} & 64.24           & \textbf{93.56} & 47.29          & 56.34             & 99.7  & -   & 98.43 & 0.89  & 0.8274  \\
        Case 19 & 51.25 & 85.73 & \textbf{100.0} & \textbf{100.0}  & 79.37          & 99.57          & \textbf{100.0}    & -     & -   & 100.0 & 0.98  & 0.7087  \\
        Case 20 & 44.90 & 79.69 & 99.14 & \textbf{100.0}           & 92.51          & \textbf{100.0} & \textbf{100.0}    & -     & -   & 94.12 & 0.86  & 0.7288  \\
        Case 21 &  8.65 & \textbf{85.63} & 81.71 & 70.48           & 78.06          & 93.29          & \textbf{100.0}    & -     & 6   & 70.59 & 0.50  & 0.3126  \\
        \midrule
        \textbf{Overall} & 44.11\tnote{*}  & 71.17\tnote{*}  & 81.43 & \textbf{83.90} & 91.52 & 93.00 & \textbf{93.58} & 98.8\tnote{*} & 71.46 & 94.08 & 80.10\tnote{*} & 0.7301\tnote{*}\\
        \bottomrule
    \end{tabular}
    \begin{tablenotes}\footnotesize
        \item[*] Self calculated since it is not stated in original work
        \item[\ddag] Here a Detection Rate (DR) is stated
        \item[\dag] Here the F1-score is stated instead of the fault detection accuracy
    \end{tablenotes}
    \end{threeparttable}
    \label{tab:acc_comparison}
\end{table*}

In addition, there are other configurations using just a subset of the 22 cases of \ac{TE}. In \cite{chai_enhanced_2020}, Cases 16 - 21 as well as the regular operation case were omitted when applying a random forest model.
\cite{yuan_multiscale_2019} proposes a multiscale convolutional neural network and long short-term memory (MCNN-LSTM), however they only apply it on cases 0, 1, 2, 4, 8, 10, 12, 13, 14, 17 and 18 of the \ac{TE}-Process.
Similarly, there is \cite{yu_multichannel_2021} which proposes a multichannel one-dimensional convolutional neural network (MC1-DCNN), however they omit the regular operation case 0 and also incorporate features from the frequency domain whereas we only work with compressed data in the time domain.

In contrast to leaving out fault cases, \cite{yu_whole_2020} leaves out sensor channels and only utilizes 33 channels of the dataset when doing process monitoring based on unstable neuron outputs in deep belief networks, and hence lacks comparability.
Finally, \cite{singh_chadha_time_2019} utilizes a CNN-pooling based architecture, however they do binary classification instead of having one model for all cases. Even though they report F1-scores instead of accuracies, we can state that our multi-class model is able to outperform the multiple binary ones.

%% file: plots/random_init_eval_plots.tex
\newcommand{\boxplot}[1]{%
    \begin{tikzpicture}[tight background]
	\pgfplotstableread[col sep=comma]{#1}\csvdata%
	\begin{axis}[%
		boxplot/draw direction = y,%
        ymax = 74,%
        ymin = 38,%
        xmin = 0.3,%
        xmax = 6.7,%
		ymajorgrids,%
		xtick = {1, 2, 3, 4, 5, 6},%
		xticklabel style = {align=center, font=\footnotesize\linespread{0.8}\selectfont},
        xticklabels = {Quat.\\Comp., Quat.\\Mag., Real Eq.\\Params, Real Eq.\\Features, Baseline\\Unc., Baseline\\Mean},%
        tick pos = left,%
        ytick style = {draw=none},
        yticklabel style = {font=\footnotesize\linespread{0.8}\selectfont},
		ylabel = {Accuracy in \%},%
        ylabel near ticks,%
        ylabel style = {anchor=south, inner sep=0pt},
        every boxplot/.style={mark=x,every mark/.append style={fill=white}},%
        boxplot/every median/.style={matplotlib_orange, thick},%
        width=.523\linewidth,%
        height=0.314\linewidth,%
	]%
		\foreach \n in {0,...,5} {%
			\addplot[boxplot, fill=matplotlib_blue, draw=black] table[y index=\n] {\csvdata};%
		}%
	\end{axis}%
    \end{tikzpicture}%
}%
%
%
\noindent
\begin{figure*}[htb]
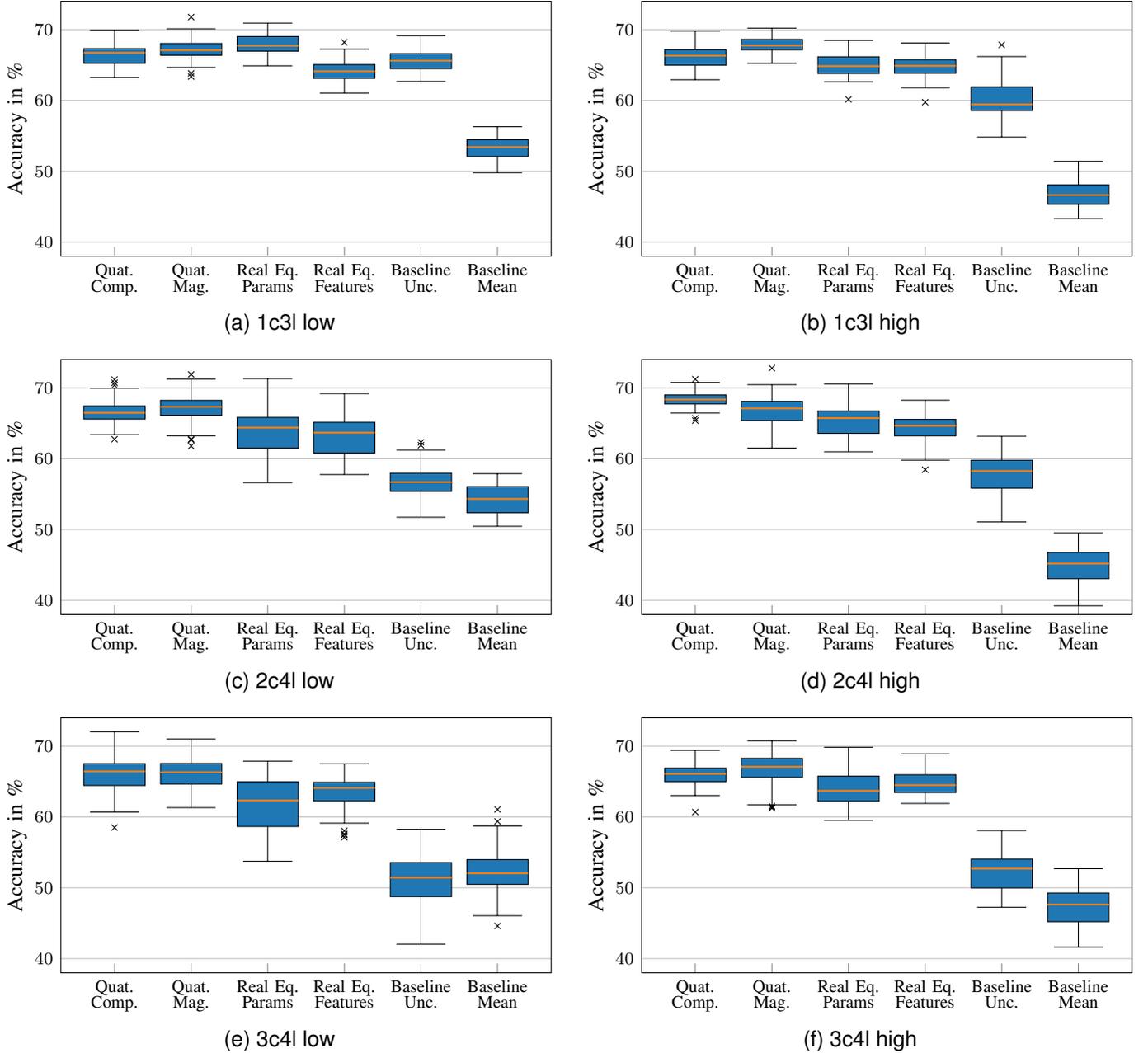

    \centering
    \subfloat[1c3l low]{\boxplot{plots/1c3l_low.csv}}%
    \hfill
    \subfloat[1c3l high]{\boxplot{plots/1c3l_high.csv}}\\
    \subfloat[2c4l low]{\boxplot{plots/2c4l_low.csv}}%
    \hfill
    \subfloat[2c4l high]{\boxplot{plots/2c4l_high.csv}}\\
    \subfloat[3c4l low]{\boxplot{plots/3c4l_low.csv}}%
    \hfill
    \subfloat[3c4l high]{%
        \boxplot{plots/3c4l_high.csv}%
        \label{subfig:3c4l_high}%
    }%
    \caption{Obtained results when training for 50 times with the best determined hyperparameter and different random parameter initialization. Note that in \protect\subref{subfig:3c4l_high} one outlier at $4.55\%$ for the real eq. features boxplot is not displayed.}
    \label{fig:random_init_boxplot}
\end{figure*}%

%% file: content/06_conclusion.tex
\section{Conclusion}
\label{sec:conclusion}

In this paper, we proposed a novel time-series compression approach for time-series data where we encode the minimum, maximum, mean and standard deviation of time-series extracts in a quaternion, yielding a quaternion valued time-series. Through the usage of the Hamilton product as the fundamental multiplication within \acp{QNN}, we were able to link these features in the quaternion convolution layer. 
Further, we developed a novel quaternion backpropagation utilizing the GHR-Calculus. After introducing the required fundamentals and quaternion mathematics, we showed that by using plain partial derivatives with respect to the quaternion components as in other approaches to quaternion backpropagation, the product and more critical, the chain rule, does not hold. By applying the GHR calculus, we end up with derivatives which do, to create our novel quaternion backpropagation algorithm. We further provided insights on the relation of automatic differentiation and quaternion backpropagation, and pointed out a scenario where automatic differentiation can be used to train \ac{QNN}.

In the conducted experiments, quaternion valued models were able to outperform real valued counterparts in a fully supervised learning setup, utilizing the same compressed data, and also a baseline without any compression and a simple mean-resampling baseline. In a random initialization test, we could further show that these results hold for 50 different random model initialization. Furthermore, the quaternion variants showed the tendency to have lower variation and smaller confidence intervals. Also, in a self-supervised learning setup based on SimCLR-TS we could prove the superiority of the quaternion valued architectures, even outperforming the results from SimCLR-TS. 
Simultaneously, we compared two quaternion max-pooling variants, where the magnitude pooling showed an overall performance advantage in the fully supervised learning setup due to its more regularizing behavior, and the component pooling in the self supervised setup due to its higher degrees of flexibility.

Hence, we can conclude that quaternion valued architectures offer a performance advantage in this compression methodology over their real valued counterparts throughout all used model configurations and training setups. 

In future work, we plan to investigate the impact or contribution of the individual properties used to form the quaternion for the final classification result. Based on these outcomes, other features representing the time-series extracts shall be observed to find the ideal combination of four properties. Also, we aim to investigate activation functions specifically tailored for quaternion models instead of using elementwise operating functions, with a particular focus on their derivatives and how the relation with automatic differentiation is in this scenario.

%% file: content/appendix.tex
\section{HR-Calculus}
\label{subsec:HrCalculus}

Similar to the CR-Calculus \cite{kreutz-delgado_complex_2009}, \cite{mandic_quaternion_2011} introduces the HR-Calculus as a method to derive quaternion valued functions.This enables deriving holomorphic quaternionic functions as well as nonholomorphic real functions of quaternion variables. The quaternion derivatives are derived as 
\begin{equation}
\begin{aligned}
    \frac{\partial f}{\partial \quaternion{q}} &= 
    \frac{1}{4}\left( \frac{\partial f}{\partial \quatCompR{q}} - \frac{\partial f}{\partial \quatCompI{q}}\imagI - \frac{\partial f}{\partial \quatCompJ{q}}\imagJ - \frac{\partial f}{\partial \quatCompK{q}}\imagK \right) \\
    \frac{\partial f}{\partial \quatInvI{q}} &= 
    \frac{1}{4}\left( \frac{\partial f}{\partial \quatCompR{q}} - \frac{\partial f}{\partial \quatCompI{q}}\imagI + \frac{\partial f}{\partial \quatCompJ{q}}\imagJ + \frac{\partial f}{\partial \quatCompK{q}}\imagK \right) \\
    \frac{\partial f}{\partial \quatInvJ{q}} &= 
    \frac{1}{4}\left( \frac{\partial f}{\partial \quatCompR{q}} + \frac{\partial f}{\partial \quatCompI{q}}\imagI - \frac{\partial f}{\partial \quatCompJ{q}}\imagJ + \frac{\partial f}{\partial \quatCompK{q}}\imagK \right) \\
    \frac{\partial f}{\partial \quatInvK{q}} &= 
    \frac{1}{4}\left( \frac{\partial f}{\partial \quatCompR{q}} + \frac{\partial f}{\partial \quatCompI{q}}\imagI + \frac{\partial f}{\partial \quatCompJ{q}}\imagJ - \frac{\partial f}{\partial \quatCompK{q}}\imagK \right) .\\
\end{aligned}    
\label{equ:hr_calculus}
\end{equation}

The corresponding conjugate derivatives are defined as

\begin{equation}
\begin{aligned}
    \frac{\partial f}{\partial \quatConj{q}} &= 
    \frac{1}{4} \left(\frac{\partial f}{\partial \quatCompR{q}} + \frac{\partial f}{\partial \quatCompI{q}}\imagI + \frac{\partial f}{\partial \quatCompJ{q}}\imagJ + \frac{\partial f}{\partial \quatCompK{q}}\imagK \right) \\
    \frac{\partial f}{\partial \quatConjInvI{q}} &= 
    \frac{1}{4} \left(\frac{\partial f}{\partial \quatCompR{q}} + \frac{\partial f}{\partial \quatCompI{q}}\imagI - \frac{\partial f}{\partial \quatCompJ{q}}\imagJ - \frac{\partial f}{\partial \quatCompK{q}}\imagK \right) \\
    \frac{\partial f}{\partial \quatConjInvJ{q}} &= 
    \frac{1}{4} \left(\frac{\partial f}{\partial \quatCompR{q}} - \frac{\partial f}{\partial \quatCompI{q}}\imagI + \frac{\partial f}{\partial \quatCompJ{q}}\imagJ - \frac{\partial f}{\partial \quatCompK{q}}\imagK \right) \\
    \frac{\partial f}{\partial \quatConjInvK{q}} &= 
    \frac{1}{4} \left(\frac{\partial f}{\partial \quatCompR{q}} - \frac{\partial f}{\partial \quatCompI{q}}\imagI - \frac{\partial f}{\partial \quatCompJ{q}}\imagJ + \frac{\partial f}{\partial \quatCompK{q}}\imagK \right) .\\
\end{aligned}
\label{equ:hr_calculus_conjugate}
\end{equation}

\begin{proposition}
    When deriving a quaternion valued function $f(q), q \in \mathbb{H}$ using \eqref{equ:hr_calculus} and the known product rule from $\mathbb{R}$, the product rule also does not apply.
\end{proposition}
\begin{proof}
    Again consider $f(\quaternion{q}) = \quaternion{q}\quatConj{q} = \quatCompR{q}^2 + \quatCompI{q}^2 + \quatCompJ{q}^2 + \quatCompK{q}^2$ as the function of choice. Then the direct derivation is 
    \begin{equation}
    \begin{split}
        \frac{\partial f}{\partial \quaternion{q}} 
        &= \frac{1}{4}\left(\frac{\partial f}{\partial \quatCompR{q}} - \frac{\partial f}{\partial \quatCompI{q}}\imagI - \frac{\partial f}{\partial \quatCompJ{q}}\imagJ - \frac{\partial f}{\partial \quatCompK{q}}\imagK \right) \\
        &= \frac{1}{4}\left(2\quatCompR{q} - 2\quatCompI{q}\imagI - 2\quatCompJ{q}\imagJ -2\quatCompK{q}\imagK \right) \\
        &= \frac{1}{2}\quatConj{q}
    \end{split}
    \end{equation}
    
    Using the product rule, we can calculate the same derivation using 
    \begin{equation}
        \frac{\partial}{\partial \quaternion{q}} \quaternion{q}\quatConj{q} = \quaternion{q} \frac{\partial \quatConj{q}}{\partial \quaternion{q}} + \frac{\partial \quaternion{q}}{\partial \quaternion{q}}\quatConj{q}
        \label{equ:product_rule_hr_calculus_derivation}
    \end{equation}
    Calculating the partial results
    \begin{equation}
        \frac{\partial \quatConj{q} }{\partial \quaternion{q}} 
        = \frac{1}{4}(1 +\imagI\imagI + \imagJ\imagJ + \imagK\imagK)
        = -\frac{1}{2}
    \end{equation}
    and 
    \begin{equation}
        \frac{\partial \quaternion{q}}{\partial \quaternion{q}} 
        = \frac{1}{4}(1 - \imagI\imagI - \imagJ\imagJ - \imagK\imagK)
        = 1
    \end{equation}
    and inserting back into \eqref{equ:product_rule_hr_calculus_derivation} yields 
    \begin{equation}
        \begin{aligned}
        \frac{\partial}{\partial \quaternion{q}} \quaternion{q}\quatConj{q} 
        &= \quaternion{q} \frac{\partial \quatConj{q}}{\partial \quaternion{q}} + \frac{\partial \quaternion{q}}{\partial \quaternion{q}}\quatConj{q} \\
        &= \frac{-1}{2} \quaternion{q} + 1 \quatConj{q} \neq \frac{1}{2}\quatConj{q}
        \end{aligned}
    \end{equation}
\end{proof}

\section{GHR-Calculus}
\label{subsec:ExaplesGHR}

\begin{example}
    When using the GHR-Calculus and definitions, product rule can be used as follows: 
    
    \paragraph*{Solution} Again consider $f(\quaternion{q}) = \quaternion{q}\quatConj{q} = \quatCompR{q}^2 + \quatCompI{q}^2 + \quatCompJ{q}^2 + \quatCompK{q}^2$ as the function of choice. Then the direct derivation is 
    \begin{equation}
    \begin{split}
        \frac{\partial f}{\partial \quaternion{q}} 
        &= \frac{1}{4}\left(\frac{\partial f}{\partial \quatCompR{q}} - \frac{\partial f}{\partial \quatCompI{q}}\imagI - \frac{\partial f}{\partial \quatCompJ{q}}\imagJ - \frac{\partial f}{\partial \quatCompK{q}}\imagK \right) \\
        &= \frac{1}{4}\left(2\quatCompR{q} - 2\quatCompI{q}\imagI - 2\quatCompJ{q}\imagJ -2\quatCompK{q}\imagK \right) \\
        &= \frac{1}{2}\quatConj{q}
    \end{split}
    \end{equation}
    
    Using the product rule, we can calculate the same derivation using 
    \begin{equation}
    \frac{\partial(\quaternion{q}\quatConj{q})}{\partial \quaternion{q}^\quaternion{\mu}} 
    = \quaternion{q} \frac{\partial(\quatConj{q})}{\partial \quaternion{q}^\quaternion{\mu}} + \frac{\partial(\quaternion{q})}{\partial \quaternion{q}^{\quatConj{q}\quaternion{\mu}}} \quatConj{q}
    = \quaternion{q} \frac{\partial(\quatConj{q})}{\partial \quaternion{q}} + \frac{\partial(\quaternion{q})}{\partial \quaternion{q}^{\quatConj{q}}} \quatConj{q}
    ~\text{with}~\quaternion{\mu} = 1
    \label{equ:product_rule_ghr_calculus_derivation}
    \end{equation}
    Calculating the partial results
    
    \begin{equation}
    \frac{\partial(\quatConj{q})}{\partial \quaternion{q}^\quaternion{\mu}}
    = \frac{-1}{2}~~ (\text{same as for HR with } \mu = 1)
 \end{equation}
    
    and 
    
    \begin{equation}
    \begin{aligned}
    \frac{\partial(\quaternion{q})}{\partial \quaternion{q}^{\quatConj{q}}} 
     &= \frac{1}{4}\left(\frac{\partial f}{\partial \quatCompR{q}} - \frac{\partial f}{\partial \quatCompI{q}}\imagI^{\quatConj{q}} - \frac{\partial f}{\partial \quatCompJ{q}}\imagJ^{\quatConj{q}} - \frac{\partial f}{\partial \quatCompK{q}}\imagK^{\quatConj{q}} \right) \\
    &= \frac{1}{4} \left( 1 - \imagI\imagI^{\quatConj{q}} - \imagJ\imagJ^{\quatConj{q}} - \imagK\imagK^{\quatConj{q}} \right) \\
    &= \frac{1}{4} \left( \quatConj{q}\quaternion{q}^{*^{-1}} - \imagI \quatConj{q} \imagI \quaternion{q}^{*^{-1}} - \imagJ \quatConj{q} \imagJ \quaternion{q}^{*^{-1}} - \imagK \quatConj{q} \imagK \quaternion{q}^{*^{-1}} \right) \\
    &= \frac{1}{4} \left( \quatConj{q} - \imagI \quatConj{q} \imagI - \imagJ \quatConj{q} \imagJ - \imagK \quatConj{q} \imagK \right) \quaternion{q}^{*^{-1}} \\
    &= \frac{1}{4} \left( \quatConj{q} + \quatConjInvI{q} + \quatConjInvJ{q} + \quatConjInvK{q} \right) \quaternion{q}^{*^{-1}} \\
    &= \quatCompR{q} \quaternion{q}^{*^{-1}}
    \end{aligned}
    \end{equation}
    
    and inserting back into \eqref{equ:product_rule_ghr_calculus_derivation} yields 
    \begin{equation}
    \begin{aligned}
        \frac{\partial(\quaternion{q}\quatConj{q})}{\partial \quaternion{q}^\quaternion{\mu}} 
        &= q \frac{\partial(\quatConj{q})}{\partial \quaternion{q}^\quaternion{\mu}} + \frac{\partial(\quaternion{q})}{\partial \quaternion{q}^{\quaternion{q}\mu}} \quatConj{q} \\
        &= \frac{-1}{2}\quaternion{q} + \quatCompR{q} \quaternion{q}^{*^{-1}} \quatConj{q} \\
        &= \frac{-1}{2}\left(\quatCompR{q} + \quatCompI{q}\imagI + \quatCompJ{q}\imagJ + \quatCompK{q}\imagK\right) + \quatCompR{q} \\
        &= \frac{1}{2}\left(\quatCompR{q} - \quatCompI{q}\imagI - \quatCompJ{q}\imagJ - \quatCompK{q}\imagK\right) \\
        &= \frac{1}{2} \quatConj{q} .
    \end{aligned}
    \end{equation}
    
\end{example}

\section{Detailed Derivative Calculations}

\subsection{Detailed calculations for the derivatives of $\loss$ with respect to the invoutions $\quatInvI{y}$, $\quatInvJ{y}$ and $\quatInvK{y}$}
\label{subsec:AppFinalLayerLeftDerivatives}

\begin{equation}
\begin{aligned}
    \frac{\partial \loss}{\partial \quatInvI{y}} \!
    &= \! \frac{\partial }{\partial \quatInvI{y}} \left[ (\quatCompR{d} \!-\! \quatCompR{y})^2 \!+\! (\quatCompI{d} \!-\! \quatCompI{y})^2 \!+\! (\quatCompJ{d} \!-\! \quatCompJ{y})^2 \!+\! (\quatCompK{d} \!-\! \quatCompK{y})^2 \right]\\ 
    &= \! \frac{1}{4} \! \left[-2(\quatCompR{d} \!-\! \quatCompR{y}) \!+\! 2(\quatCompI{d} \!-\! \quatCompI{y})\imagI \!-\! 2(\quatCompJ{d} \!-\! \quatCompJ{y})\imagJ \!-\! 2(\quatCompK{d} \!-\! \quatCompK{y})\imagK \right] \\
    &= \! -\frac{1}{2}\left[(\quatCompR{d} \!-\! \quatCompR{y}) \!-\! (\quatCompI{d} \!-\! \quatCompI{y})\imagI \!+\! (\quatCompJ{d} \!-\! \quatCompJ{y})\imagJ \!+\! (\quatCompK{d} \!-\! \quatCompK{y})\imagK \right] \\
    &= \! -\frac{1}{2} \quatConjInvI{(d - y)} = -\frac{1}{2} \quatConjInvI{e}
\end{aligned}
\end{equation}

\begin{equation}
\begin{aligned}
    \frac{\partial \loss}{\partial \quatInvJ{y}} \!
    &= \! \frac{\partial }{\partial \quatInvJ{y}} \left[ (\quatCompR{d} \!-\! \quatCompR{y})^2 \!+\! (\quatCompI{d} \!-\! \quatCompI{y})^2 \!+\! (\quatCompJ{d} \!-\! \quatCompJ{y})^2 \!+\! (\quatCompK{d} \!-\! \quatCompK{y})^2 \right] \\ 
    &= \! \frac{1}{4} \! \left[-2(\quatCompR{d} \!-\! \quatCompR{y}) \!-\! 2(\quatCompI{d} \!-\! \quatCompI{y})\imagI \!+\! 2(\quatCompJ{d} \!-\! \quatCompJ{y})\imagJ \!-\! 2(\quatCompK{d} \!-\! \quatCompK{y})\imagK \right] \\
    &= \! -\frac{1}{2}\left[(\quatCompR{d} \!-\! \quatCompR{y}) \!+\! (\quatCompI{d} \!-\! \quatCompI{y})\imagI \!-\! (\quatCompJ{d} \!-\! \quatCompJ{y})\imagJ \!+\! (\quatCompK{d} \!-\! \quatCompK{y})\imagK \right] \\
    &= \! -\frac{1}{2} \quatConjInvJ{(d - y)} = -\frac{1}{2} \quatConjInvJ{e}
\end{aligned}
\end{equation}

\begin{equation}
\begin{aligned}
    \frac{\partial \loss}{\partial \quatInvK{y} } \mkern-3.5mu
    &= \mkern-3.5mu \frac{\partial }{\partial \quatInvK{y}} \left[ (\quatCompR{d} \!-\! \quatCompR{y})^2 \!+\! (\quatCompI{d} \!-\! \quatCompI{y})^2 \!+\! (\quatCompJ{d} \!-\! \quatCompJ{y})^2 \!+\! (\quatCompK{d} \!-\! \quatCompK{y})^2 \right] \\ 
    &= \mkern-3.5mu \frac{1}{4} \! \left[-2(\quatCompR{d} \!-\! \quatCompR{y}) \!-\! 2(\quatCompI{d} \!-\! \quatCompI{y})\imagI \!-\! 2(\quatCompJ{d} \!-\! \quatCompJ{y})\imagJ \!+\! 2(\quatCompK{d} \!-\! \quatCompK{y})\imagK \right] \\
    &= \mkern-3.5mu -\frac{1}{2}\left[(\quatCompR{d} \!-\! \quatCompR{y}) \!+\! (\quatCompI{d} \!-\! \quatCompI{y})\imagI \!+\! (\quatCompJ{d} \!-\! \quatCompJ{y})\imagJ \!-\! (\quatCompK{d} \!-\! \quatCompK{y})\imagK \right] \\
    &= \mkern-3.5mu -\frac{1}{2} \quatConjInvK{(d - y)} = -\frac{1}{2} \quatConjInvK{e}
\end{aligned}
\end{equation}

\subsection{Detailed calculations for the derivatives of the involutions $\quatInvI{y}$, $\quatInvJ{y}$ and $\quatInvK{y}$ with respect to $\quatConj{w}$}
\label{subsec:AppFinalLayerWeights}

\begin{equation}
\begin{aligned}
    \frac{\partial \quatInvI{y}}{\partial \quatConj{w}} 
    &= \frac{\partial \InvI{(\quaternion{w}\quaternion{a} + \quaternion{b})}}{\partial \quatConj{w}}
    = \frac{\partial \InvI{(\quaternion{w}\quaternion{a})}}{\partial \quatConj{w}} \\
    &= \frac{\partial}{\partial \quatConj{w}} [
    (\quatCompR{a} \quatCompR{w} - \quatCompI{a} \quatCompI{w} - \quatCompJ{a} \quatCompJ{w} - \quatCompK{a} \quatCompK{w}) \\
    &+ (\quatCompR{a} \quatCompI{w} + \quatCompI{a} \quatCompR{w} - \quatCompJ{a} \quatCompK{w} + \quatCompK{a} \quatCompJ{w} ) \imagI \\
    &- (\quatCompR{a} \quatCompJ{w} + \quatCompI{a} \quatCompK{w} + \quatCompJ{a} \quatCompR{w} - \quatCompK{a} \quatCompI{w} ) \imagJ \\
    &- (\quatCompR{a} \quatCompK{w} - \quatCompI{a} \quatCompJ{w} + \quatCompJ{a} \quatCompI{w} + \quatCompK{a} \quatCompR{w} ) \imagK ] \\
    &= \frac{1}{4} 
    [\quatCompR{a} \!+\! \quatCompI{a} \imagI \!-\! \quatCompJ{a} \imagJ \!-\! \quatCompK{a} \imagK
    \!+\! (\quatCompR{a}\imagI \!-\! \quatCompI{a} \!-\! \quatCompJ{a}\imagK \!+\! \quatCompK{a}\imagJ) \imagI \\
    &+ (-\quatCompR{a}\imagJ \!+\! \quatCompI{a}\imagK \!-\! \quatCompJ{a} \!+\! \quatCompK{a}\imagI) \imagJ
    \!+\! (-\quatCompR{a}\imagK \!-\! \quatCompI{a}\imagJ \!-\! \quatCompJ{a}\imagI \!-\! \quatCompK{a}) \imagK ] \\
    &= \frac{1}{2} \left[\quaternionConjComponents{a} \right] \\
    &= \frac{1}{2} \quatConj{a}
\end{aligned}    
\end{equation}
\begin{equation}
\begin{aligned}
    \frac{\partial \quatInvJ{y}}{\partial \quatConj{w}} 
    &= \frac{\partial \InvJ{(\quaternion{w}\quaternion{a} + \quaternion{b})}}{\partial \quatConj{w}}
    = \frac{\partial \InvJ{(\quaternion{w}\quaternion{a})}}{\partial \quatConj{w}} \\
    &= \frac{\partial}{\partial \quatConj{w}} [
    (\quatCompR{a} \quatCompR{w} - \quatCompI{a} \quatCompI{w} - \quatCompJ{a} \quatCompJ{w} - \quatCompK{a} \quatCompK{w}) \\
    &- (\quatCompR{a} \quatCompI{w} + \quatCompI{a} \quatCompR{w} - \quatCompJ{a} \quatCompK{w} + \quatCompK{a} \quatCompJ{w} ) \imagI \\
    &+ (\quatCompR{a} \quatCompJ{w} + \quatCompI{a} \quatCompK{w} + \quatCompJ{a} \quatCompR{w} - \quatCompK{a} \quatCompI{w} ) \imagJ \\
    &- (\quatCompR{a} \quatCompK{w} - \quatCompI{a} \quatCompJ{w} + \quatCompJ{a} \quatCompI{w} + \quatCompK{a} \quatCompR{w} ) \imagK ] \\
    &= \frac{1}{4} 
    [\quatCompR{a} \!-\! \quatCompI{a} \imagI \!+\! \quatCompJ{a} \imagJ \!-\! \quatCompK{a} \imagK
    \!+\! (-\quatCompR{a}\imagI \!-\! \quatCompI{a} \!-\! \quatCompJ{a}\imagK \!-\! \quatCompK{a}\imagJ) \imagI \\
    &+ (\quatCompR{a}\imagJ \!+\! \quatCompI{a}\imagK \!-\! \quatCompJ{a} \!-\! \quatCompK{a}\imagI) \imagJ
    \!+\! (-\quatCompR{a}\imagK \!+\! \quatCompI{a}\imagJ \!+\! \quatCompJ{a}\imagI \!-\! \quatCompK{a}) \imagK ] \\
    &= \frac{1}{2} \left[\quaternionConjComponents{a} \right] \\
    &= \frac{1}{2} \quatConj{a}
\end{aligned}    
\end{equation}
\begin{equation}
\begin{aligned}
    \frac{\partial \quatInvK{y}}{\partial \quatConj{w}} 
    &= \frac{\partial \InvK{(\quaternion{w}\quaternion{a} + \quaternion{b})}}{\partial \quatConj{w}}
    = \frac{\partial \InvK{(\quaternion{w}\quaternion{a})}}{\partial \quatConj{w}} \\
    &= \frac{\partial}{\partial \quatConj{w}} [
    (\quatCompR{a} \quatCompR{w} - \quatCompI{a} \quatCompI{w} - \quatCompJ{a} \quatCompJ{w} - \quatCompK{a} \quatCompK{w}) \\
    &- (\quatCompR{a} \quatCompI{w} + \quatCompI{a} \quatCompR{w} - \quatCompJ{a} \quatCompK{w} + \quatCompK{a} \quatCompJ{w} ) \imagI \\
    &- (\quatCompR{a} \quatCompJ{w} + \quatCompI{a} \quatCompK{w} + \quatCompJ{a} \quatCompR{w} - \quatCompK{a} \quatCompI{w} ) \imagJ \\
    &+ (\quatCompR{a} \quatCompK{w} - \quatCompI{a} \quatCompJ{w} + \quatCompJ{a} \quatCompI{w} + \quatCompK{a} \quatCompR{w} ) \imagK ] \\
    &= \frac{1}{4} 
    [\quatCompR{a} \!-\! \quatCompI{a} \imagI \!-\! \quatCompJ{a} \imagJ \!+\! \quatCompK{a} \imagK
    \!+\! (-\quatCompR{a}\imagI \!-\! \quatCompI{a} \!+\! \quatCompJ{a}\imagK \!+\! \quatCompK{a}\imagJ) \imagI \\
    &+ (-\quatCompR{a}\imagJ \!-\! \quatCompI{a}\imagK \!-\! \quatCompJ{a} \!-\! \quatCompK{a}\imagI) \imagJ
    \!+\! (\quatCompR{a}\imagK \!-\! \quatCompI{a}\imagJ \!+\! \quatCompJ{a}\imagI \!-\! \quatCompK{a}) \imagK ] \\
    &= \frac{1}{2} \left[\quaternionConjComponents{a} \right] \\
    &= \frac{1}{2} \quatConj{a}
\end{aligned}    
\end{equation}

\subsection{Detailed calculations for the derivatives of the involutions $\quatInvI{y}$, $\quatInvJ{y}$ and $\quatInvK{y}$ with respect to $\quatConj{b}$}
\label{subsec:AppFinalLayerBias}

\begin{equation}
\begin{aligned}
    \frac{\partial \quatInvI{y}}{\partial \quatConj{b}} 
    &= \frac{\partial \InvI{(\quaternion{w}\quaternion{a} + \quaternion{b})}}{\partial \quatConj{b}}
    =  \frac{\partial \InvI{(\quaternion{b})}}{\partial \quatConj{b}} \\
    &= \frac{\partial}{\partial \quatConj{b}} \left( \quatCompR{b} + \quatCompI{b} \imagI - \quatCompJ{b} \imagJ - \quatCompK{b} \imagK \right)\\
    &= \frac{1}{4} \left( 1 + \imagI\imagI - \imagJ\imagJ - \imagK\imagK \right) \\
    &= \frac{1}{4} \left( 1 -1 +1 +1 \right) \\
    &= 0.5
\end{aligned}    
\end{equation}

\begin{equation}
\begin{aligned}
    \frac{\partial \quatInvJ{y}}{\partial \quatConj{b}} 
    &= \frac{\partial \InvJ{(\quaternion{w}\quaternion{a} + \quaternion{b})}}{\partial \quatConj{b}}
    =  \frac{\partial \InvJ{(\quaternion{b})}}{\partial \quatConj{b}} \\
    &= \frac{\partial}{\partial \quatConj{b}} \left( \quatCompR{b} - \quatCompI{b} \imagI + \quatCompJ{b} \imagJ - \quatCompK{b} \imagK \right)\\
    &= \frac{1}{4} \left( 1 - \imagI\imagI + \imagJ\imagJ - \imagK\imagK \right) \\
    &= \frac{1}{4} \left( 1 +1 -1 +1 \right) \\
    &= 0.5
\end{aligned}    
\end{equation}

\begin{equation}
\begin{aligned}
    \frac{\partial \quatInvK{y}}{\partial \quatConj{b}} 
    &= \frac{\partial \InvK{(\quaternion{w}\quaternion{a} + \quaternion{b})}}{\partial \quatConj{b}}
    =  \frac{\partial \InvK{(\quaternion{b})}}{\partial \quatConj{b}} \\
    &= \frac{\partial}{\partial \quatConj{b}} \left( \quatCompR{b} - \quatCompI{b} \imagI - \quatCompJ{b} \imagJ + \quatCompK{b} \imagK \right)\\
    &= \frac{1}{4} \left( 1 - \imagI\imagI - \imagJ\imagJ + \imagK\imagK \right) \\
    &= \frac{1}{4} \left( 1 +1 +1 -1 \right) \\
    &= 0.5
\end{aligned}    
\end{equation}

\subsection{Detailed calculations for the derivatives of $\quaternion{y}$, $\quatInvI{y}$, $\quatInvJ{y}$ and $\quatInvK{y}$ with respect to $\quaternion{a}$}
\label{subsec:AppFinalLayerActivation}

\begin{equation}
\begin{aligned}
    \frac{\partial \quaternion{y}}{\partial \quaternion{a}} \!
    &=\! \frac{\partial (\quaternion{w}\quaternion{a} + \quaternion{b})}{\partial \quaternion{a}}
    = \frac{\partial (\quaternion{w}\quaternion{a})}{\partial \quaternion{a}} \\
    &= \!\frac{\partial}{\partial \quaternion{a}} [
    (\quatCompR{a} \quatCompR{w} - \quatCompI{a} \quatCompI{w} - \quatCompJ{a} \quatCompJ{w} - \quatCompK{a} \quatCompK{w}) \\
    &+ (\quatCompR{a} \quatCompI{w} + \quatCompI{a} \quatCompR{w} - \quatCompJ{a} \quatCompK{w} + \quatCompK{a} \quatCompJ{w} ) \imagI \\
    &+ (\quatCompR{a} \quatCompJ{w} + \quatCompI{a} \quatCompK{w} + \quatCompJ{a} \quatCompR{w} - \quatCompK{a} \quatCompI{w} ) \imagJ \\
    &+ (\quatCompR{a} \quatCompK{w} - \quatCompI{a} \quatCompJ{w} + \quatCompJ{a} \quatCompI{w} + \quatCompK{a} \quatCompR{w} ) \imagK ] \\
    &=\! \frac{1}{4} [ 
    (\quatCompR{w} \!+\! \quatCompI{w}\imagI \!+\! \quatCompJ{w}\imagJ \!+\! \quatCompK{w}\imagK) \!-\! 
    (\quatCompR{w}\imagI \!-\! \quatCompI{w} \!-\! \quatCompJ{w}\imagK \!+\! \quatCompK{w}\imagJ)\imagI\\ &-\! 
    (\quatCompR{w}\imagJ \!+\! \quatCompI{w}\imagK \!-\! \quatCompJ{w} \!-\! \quatCompK{w}\imagI)\imagJ \!-\!
    (\quatCompR{w}\imagK \!-\! \quatCompI{w}\imagJ \!+\! \quatCompJ{w}\imagI \!-\! \quatCompK{w})\imagK
    ] \\
    &= \quaternionComponents{w} \\
    &= \quaternion{w}
\end{aligned}
\end{equation}

\begin{equation}
\begin{aligned}
    \frac{\partial \quatInvI{y}}{\partial \quaternion{a}} \!
    &= \! \frac{\partial \InvI{(\quaternion{w}\quaternion{a} + \quaternion{b})}}{\partial \quaternion{a}}
    = \frac{\partial \InvI{(\quaternion{w}\quaternion{a})}}{\partial \quaternion{a}} \\
    &= \! \frac{\partial}{\partial \quaternion{a}} [
    (\quatCompR{a} \quatCompR{w} - \quatCompI{a} \quatCompI{w} - \quatCompJ{a} \quatCompJ{w} - \quatCompK{a} \quatCompK{w}) \\
    &+ (\quatCompR{a} \quatCompI{w} + \quatCompI{a} \quatCompR{w} - \quatCompJ{a} \quatCompK{w} + \quatCompK{a} \quatCompJ{w} ) \imagI \\
    &- (\quatCompR{a} \quatCompJ{w} + \quatCompI{a} \quatCompK{w} + \quatCompJ{a} \quatCompR{w} - \quatCompK{a} \quatCompI{w} ) \imagJ \\
    &- (\quatCompR{a} \quatCompK{w} - \quatCompI{a} \quatCompJ{w} + \quatCompJ{a} \quatCompI{w} + \quatCompK{a} \quatCompR{w} ) \imagK ] \\
    &=\! \frac{1}{4} [ 
          ( \quatCompR{w} \!+\! \quatCompI{w}\imagI \!-\! \quatCompJ{w}\imagJ \!-\! \quatCompK{w}\imagK)
    \!-\! ( \quatCompR{w}\imagI \!-\! \quatCompI{w} \!+\! \quatCompJ{w}\imagK \!-\! \quatCompK{w}\imagJ)\imagI\\ 
    &-\!  (\!-\quatCompR{w}\imagJ \!-\! \quatCompI{w}\imagK \!-\! \quatCompJ{w} \!-\! \quatCompK{w}\imagI)\imagJ 
    \!-\! (\!-\quatCompR{w}\imagK \!+\! \quatCompI{w}\imagJ \!+\! \quatCompJ{w}\imagI \!-\! \quatCompK{w})\imagK
    ] \\
    &= 0
\end{aligned}
\end{equation}

\begin{equation}
\begin{aligned}
    \frac{\partial \quatInvJ{y}}{\partial \quaternion{a}} \!
    &= \!\frac{\partial \InvJ{(\quaternion{w}\quaternion{a} + \quaternion{b})}}{\partial \quaternion{a}}
    = \frac{\partial \InvJ{(\quaternion{w}\quaternion{a})}}{\partial \quaternion{a}} \\
    &=\! \frac{\partial}{\partial \quaternion{a}} [
    (\quatCompR{a} \quatCompR{w} - \quatCompI{a} \quatCompI{w} - \quatCompJ{a} \quatCompJ{w} - \quatCompK{a} \quatCompK{w}) \\
    &- (\quatCompR{a} \quatCompI{w} + \quatCompI{a} \quatCompR{w} - \quatCompJ{a} \quatCompK{w} + \quatCompK{a} \quatCompJ{w} ) \imagI \\
    &+ (\quatCompR{a} \quatCompJ{w} + \quatCompI{a} \quatCompK{w} + \quatCompJ{a} \quatCompR{w} - \quatCompK{a} \quatCompI{w} ) \imagJ \\
    &- (\quatCompR{a} \quatCompK{w} - \quatCompI{a} \quatCompJ{w} + \quatCompJ{a} \quatCompI{w} + \quatCompK{a} \quatCompR{w} ) \imagK ] \\
    &=\! \frac{1}{4}[ ( \quatCompR{w}       \!-\! \quatCompI{w}\imagI \!+\! \quatCompJ{w}\imagJ \!-\! \quatCompK{w}\imagK) 
    \!-\!           (-\quatCompR{w}\imagI \!-\! \quatCompI{w} \!+\! \quatCompJ{w}\imagK \!+\! \quatCompK{w}\imagJ)\imagI \\ 
    &-\!            ( \quatCompR{w}\imagJ \!-\! \quatCompI{w}\imagK \!-\! \quatCompJ{w} \!+\! \quatCompK{w}\imagI)\imagJ 
    \!-\!           (-\quatCompR{w}\imagK \!-\! \quatCompI{w}\imagJ \!-\! \quatCompJ{w}\imagI \!-\! \quatCompK{w})\imagK ] \\
    &= 0
\end{aligned}
\end{equation}

\begin{equation}
\begin{aligned}
    \frac{\partial \quatInvK{y}}{\partial \quaternion{a}} \!
    &= \!\frac{\partial \InvK{(\quaternion{w}\quaternion{a} + \quaternion{b})}}{\partial \quaternion{a}}
    = \frac{\partial \InvK{(\quaternion{w}\quaternion{a})}}{\partial \quaternion{a}} \\
    &=\! \frac{\partial}{\partial \quaternion{a}} [
    (\quatCompR{a} \quatCompR{w} - \quatCompI{a} \quatCompI{w} - \quatCompJ{a} \quatCompJ{w} - \quatCompK{a} \quatCompK{w}) \\
    &- (\quatCompR{a} \quatCompI{w} + \quatCompI{a} \quatCompR{w} - \quatCompJ{a} \quatCompK{w} + \quatCompK{a} \quatCompJ{w} ) \imagI \\
    &- (\quatCompR{a} \quatCompJ{w} + \quatCompI{a} \quatCompK{w} + \quatCompJ{a} \quatCompR{w} - \quatCompK{a} \quatCompI{w} ) \imagJ \\
    &+ (\quatCompR{a} \quatCompK{w} - \quatCompI{a} \quatCompJ{w} + \quatCompJ{a} \quatCompI{w} + \quatCompK{a} \quatCompR{w} ) \imagK ] \\
    &=\! \frac{1}{4}[ ( \quatCompR{w}       \!-\! \quatCompI{w}\imagI \!-\! \quatCompJ{w}\imagJ \!+\! \quatCompK{w}\imagK) 
    \!-\!           (\!-\quatCompR{w}\imagI \!-\! \quatCompI{w} \!-\! \quatCompJ{w}\imagK \!-\! \quatCompK{w}\imagJ)\imagI \\ 
    &-\!            (\!-\quatCompR{w}\imagJ \!+\! \quatCompI{w}\imagK \!-\! \quatCompJ{w} \!+\! \quatCompK{w}\imagI)\imagJ 
    \!-\!           (\quatCompR{w}\imagK \!+\! \quatCompI{w}\imagJ \!-\! \quatCompJ{w}\imagI \!-\! \quatCompK{w})\imagK ] \\
    &= 0
\end{aligned}
\end{equation}

\section{Layer visualization}
\label{sec:layer_visu}

\begin{figure}[hb]
    \centering
    \includegraphics[scale=.48, angle=90]{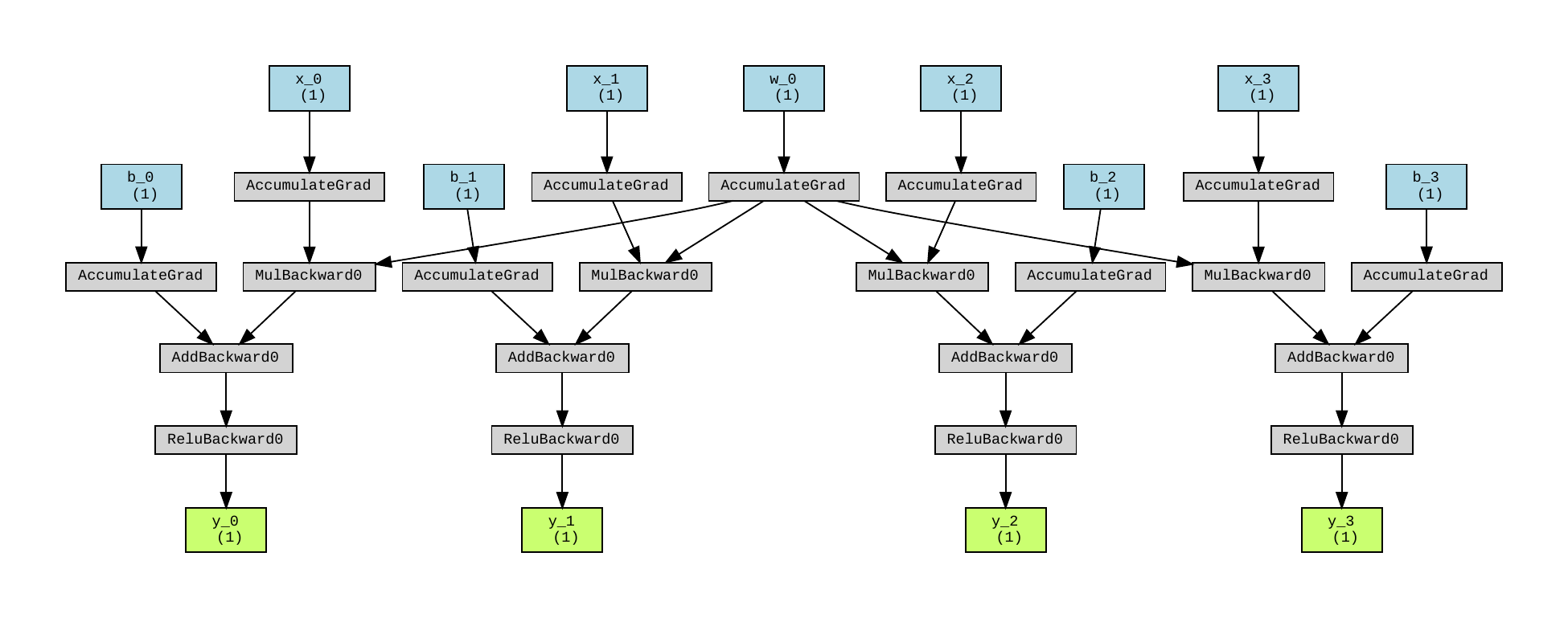}
    \caption{Visiualization of the gradient flow for parameter $\quatCompR{w}$ obtained with TorchViz}
    \label{fig:torchViz}
\end{figure}

\section{Automatic differentiation calculations}

\subsection{Derivative with respect to the activation output of the
previous layer}
\label{subsec:autograd_derivative_activation}

The derivative with respect to the layers input $a^{(l-1)}$ is calculated as
\begin{equation}
\begin{aligned}
    &\lossVector{a}
    =
    \sum_{i \in K}
    \begin{bmatrix}
        \autoGradDerivSubscript{z}{\quatCompR{a}}{i} \\[4pt]
        \autoGradDerivSubscript{z}{\quatCompI{a}}{i} \\[4pt]
        \autoGradDerivSubscript{z}{\quatCompJ{a}}{i} \\[4pt]
        \autoGradDerivSubscript{z}{\quatCompK{a}}{i} \\[4pt]
    \end{bmatrix} \\
    &= \!
    \sum_{i \in K} \!
    \begin{bmatrix}
         \quatCompSubscriptR{q}{i}\quatCompSubscriptR{w}{i,j} - \quatCompSubscriptI{q}{i}\quatCompSubscriptI{w}{i,j} - \quatCompSubscriptJ{q}{i}\quatCompSubscriptJ{w}{i,j} - \quatCompSubscriptK{q}{i}\quatCompSubscriptK{w}{i,j} \\
        -\quatCompSubscriptR{q}{i}\quatCompSubscriptI{w}{i,j} - \quatCompSubscriptI{q}{i}\quatCompSubscriptR{w}{i,j} - 
         \quatCompSubscriptJ{q}{i}\quatCompSubscriptK{w}{i,j} + \quatCompSubscriptK{q}{i}\quatCompSubscriptJ{w}{i,j} \\
        -\quatCompSubscriptR{q}{i}\quatCompSubscriptJ{w}{i,j} + \quatCompSubscriptI{q}{i}\quatCompSubscriptK{w}{i,j} - 
         \quatCompSubscriptJ{q}{i}\quatCompSubscriptR{w}{i,j} - \quatCompSubscriptK{q}{i}\quatCompSubscriptI{w}{i,j} \\
        -\quatCompSubscriptR{q}{i}\quatCompSubscriptK{w}{i,j} - \quatCompSubscriptI{q}{i}\quatCompSubscriptJ{w}{i,j} + 
         \quatCompSubscriptJ{q}{i}\quatCompSubscriptI{w}{i,j} - \quatCompSubscriptK{q}{i}\quatCompSubscriptR{w}{i,j} \\
    \end{bmatrix} \\ 
    &=
    \sum_{i \in K}
    \begin{bmatrix}
         \quatCompSubscriptR{q}{i}\quatCompSubscriptR{w}{i,j} - \quatCompSubscriptI{q}{i}\quatCompSubscriptI{w}{i,j} - \quatCompSubscriptJ{q}{i}\quatCompSubscriptJ{w}{i,j} - \quatCompSubscriptK{q}{i}\quatCompSubscriptK{w}{i,j} \\
        -(\quatCompSubscriptR{q}{i}\quatCompSubscriptI{w}{i,j} + \quatCompSubscriptI{q}{i}\quatCompSubscriptR{w}{i,j} + 
          \quatCompSubscriptJ{q}{i}\quatCompSubscriptK{w}{i,j} - \quatCompSubscriptK{q}{i}\quatCompSubscriptJ{w}{i,j}) \\
        -(\quatCompSubscriptR{q}{i}\quatCompSubscriptJ{w}{i,j} - \quatCompSubscriptI{q}{i}\quatCompSubscriptK{w}{i,j} + 
          \quatCompSubscriptJ{q}{i}\quatCompSubscriptR{w}{i,j} + \quatCompSubscriptK{q}{i}\quatCompSubscriptI{w}{i,j}) \\
        -(\quatCompSubscriptR{q}{i}\quatCompSubscriptK{w}{i,j} + \quatCompSubscriptI{q}{i}\quatCompSubscriptJ{w}{i,j} - 
          \quatCompSubscriptJ{q}{i}\quatCompSubscriptI{w}{i,j} + \quatCompSubscriptK{q}{i}\quatCompSubscriptR{w}{i,j}) \\
    \end{bmatrix} . 
\end{aligned}
\label{equ:autoGrad_hiddenLayer_activation}
\end{equation}

Taking the rows of the output vector for the real part and the imaginary parts respectively, and using Equation \eqref{equ:hamilton_product} we obtain 
\begin{equation}
    \frac{\partial \loss \left(\quaternion{a}_j^{\left(l-1\right)}\right)}{\partial \quaternion{a}_j^{(l-1)}} 
    = \sum_{i \in K} \left( \quaternion{q}_i \quaternion{w}_{i, j} \right)^*
\end{equation}

\subsection{Derivative with respect to the weights}
\label{subsec:autograd_derivative_weight}

The derivative with respect to the layers weights $w^{(l)}$ is calculated following

\begin{equation}
    \begin{aligned}
        &\lossVector{w}
        =
        \begin{bmatrix}
            \autoGradDeriv{z}{\quatCompR{w}} \\[4pt]
            \autoGradDeriv{z}{\quatCompI{w}} \\[4pt]
            \autoGradDeriv{z}{\quatCompJ{w}} \\[4pt]
            \autoGradDeriv{z}{\quatCompK{w}} \\[4pt]
        \end{bmatrix} \\
        &=
        \begin{bmatrix}
             \quatCompR{q}\quatCompR{a} - \quatCompI{q}\quatCompI{a} - \quatCompJ{q}\quatCompJ{a} - \quatCompK{q}\quatCompK{a} \\
            -\quatCompR{q}\quatCompI{a} - \quatCompI{q}\quatCompR{a} + \quatCompJ{q}\quatCompK{a} - \quatCompK{q}\quatCompJ{a} \\
            -\quatCompR{q}\quatCompJ{a} - \quatCompI{q}\quatCompK{a} - \quatCompJ{q}\quatCompR{a} + \quatCompK{q}\quatCompI{a} \\
            -\quatCompR{q}\quatCompK{a} + \quatCompI{q}\quatCompJ{a} - \quatCompJ{q}\quatCompI{a} - \quatCompK{q}\quatCompR{a} \\
        \end{bmatrix} \\
        &=
        \begin{bmatrix}
            \quatCompR{q}\quatCompR{a} - (-\quatCompI{q})(-\quatCompI{a}) - (-\quatCompJ{q})(-\quatCompJ{a}) - (-\quatCompK{q})(-\quatCompK{a}) \\
            \quatCompR{q}(-\quatCompI{a}) + (-\quatCompI{q})\quatCompR{a} + (-\quatCompJ{q})(-\quatCompK{a}) - (-\quatCompK{q})(-\quatCompJ{a}) \\
            \quatCompR{q}(-\quatCompJ{a}) - (-\quatCompI{q})(-\quatCompK{a}) + (-\quatCompJ{q})\quatCompR{a} + (-\quatCompK{q})(-\quatCompI{a}) \\
            \quatCompR{q}(-\quatCompK{a}) + (-\quatCompI{q})(-\quatCompJ{a}) - (-\quatCompJ{q})(-\quatCompI{a}) + (-\quatCompK{q})\quatCompR{a} \\
        \end{bmatrix} .
    \end{aligned}
    \end{equation}
    Composing the rows back to a quaternion and using Equation \eqref{equ:hamilton_product} yields 
    \begin{equation}
        \frac{\partial \loss(w)}{\partial \quaternion{w}^{(l)}}
        = \quatConj{q} \quatConj{a} .
    \end{equation}

\subsection{Derivative with respect to the bias}
\label{subsec:autograd_derivative_bias}

As the respective bias component is not involved in the calculation of all output components, but just $\quatCompR{b}$ for $\quatCompR{z}$, $\quatCompI{b}$ for $\quatCompI{z}$, $\quatCompJ{b}$ for $\quatCompJ{z}$ and $\quatCompK{b}$ for $\quatCompK{z}$, the derivatives simplify to

\begin{equation}
    \begin{aligned}
            \lossVector{b}
            = 
            \begin{bmatrix}
                \frac{\partial \loss}{\partial \quatCompR{z}}\frac{\partial \quatCompR{z}}{\partial \quatCompR{b}} \\[4pt]
                \frac{\partial \loss}{\partial \quatCompI{z}}\frac{\partial \quatCompI{z}}{\partial \quatCompI{b}} \\[4pt]
                \frac{\partial \loss}{\partial \quatCompJ{z}}\frac{\partial \quatCompJ{z}}{\partial \quatCompJ{b}} \\[4pt]
                \frac{\partial \loss}{\partial \quatCompK{z}}\frac{\partial \quatCompK{z}}{\partial \quatCompK{b}} \\[4pt]
            \end{bmatrix}
            =
            \begin{bmatrix}
              \quatCompR{q} \\
             -\quatCompI{q} \\
             -\quatCompJ{q} \\
             -\quatCompK{q} \\
            \end{bmatrix} .
    \end{aligned}
    \end{equation}
    Consequently, the conversion back to a quaternion is
    \begin{equation}
        \frac{\partial \loss(b)}{\partial \quaternion{b}^{(l)}}
        = \quatConj{q} .
    \end{equation}